\documentclass[11pt]{article}

\usepackage{bm}
\usepackage{amsfonts,amsthm,mathrsfs,amssymb,amsmath}
\usepackage{float}
\usepackage[page,title,titletoc,header]{appendix}
\usepackage{color}
\usepackage{authblk}
\usepackage{graphicx}
\usepackage{url}
\usepackage{enumerate,anysize,color}
\usepackage[a4paper,hmargin=1in,vmargin=1in]{geometry}

\newtheorem{thm}{Theorem}[section]
\newtheorem{lemma}[thm]{Lemma}

\newtheorem{corollary}[thm]{Corollary}

\newtheorem{rem}[thm]{Remark}
\newtheorem{Exa}{Example}[section]
\newtheorem{as}{Assumption}
\newtheorem{alg}{Algorithm}

\newcommand{\be}{\begin{equation}}
\newcommand{\ee}{\end{equation}}
\newcommand{\bea}{\begin{eqnarray*}}
	\newcommand{\eea}{\end{eqnarray*}}
\newcommand{\bflalign}{\begin{flalign*}}
\newcommand{\eflalign}{\end{flalign*}}

\newcommand{\mR}{\mathbb{R}}

\newcommand{\mN}{\mathbb{N}}
\newcommand{\mE}{\mathbb{E}}
\newcommand{\mcE}{{\mathcal{E}}}

\newcommand{\mcS}{\mathcal{S}}

\newcommand{\mcN}{\mathcal{N}}

\newcommand{\mcX}{\mathcal{X}}
\newcommand{\mcV}{\mathcal{V}}

\newcommand{\mcK}{\mathcal{K}}
\newcommand{\mcP}{\mathcal{P}}
\newcommand{\bz}{{\bf z}} 
\newcommand{\bx}{{\bf x}} 
\newcommand{\by}{{\bf y}}
\newcommand{\bb}{{\bf b}}
\newcommand{\bK}{{\bf K}}

\newcommand{\ba}{{\bf a}}
\newcommand{\bR}{{\bf R}}

\newcommand{\mcEE}{\tilde{\mathcal{E}}}

\DeclareMathOperator*{\argmin}{arg\,min}

\newcommand{\tr}{\operatorname{tr}}
\newcommand{\la}{\langle}
\newcommand{\ra}{\rangle}
\newcommand{\eref}[1] {(\ref{#1})}

\newcommand{\RegPar}{\lambda}
\newcommand{\PRegPar}{{\lambda}}

\newcommand{\TK}{\mathcal{T}} 
\newcommand{\TKL}{\mathcal{T}_{\PRegPar}} %
\newcommand{\TXL}{\mathcal{T}_{{\bf x}\PRegPar}}

\newcommand{\LK}{\mathcal{L}}

\newcommand{\IK}{\mathcal{S}_{\rho_X}}
\newcommand{\TX}{\mathcal{T}_{\bf x}}
\newcommand{\SX}{\mathcal{S}_{\bf x}}
\newcommand{\SXS}{\mathcal{S}_{\bf \tilde{x}}} 
\newcommand{\TXS}{\mathcal{T}_{\bf \tilde{x}}} 

\newcommand{\HK}{H}
\newcommand{\HR}{H_{\rho_X}}
\newcommand{\LR}{L^2_{\rho_X}}
\newcommand{\HKS}{S}
\newcommand{\st}{P}      %
\newcommand{\proj}{P}  

\newcommand{\FR}{f_{\rho}}
\newcommand{\FH}{f_{\HK}}

\newcommand{\NOutputs}{\bby}

\newcommand{\DZF}{\Delta_1} 
\newcommand{\DZS}{\Delta_2} %
\newcommand{\DZT}{\Delta_3} 
\newcommand{\DZN}{\Delta_4} 
\newcommand{\DZI}{\Delta_5}  
\newcommand{\Bias}{R \lambda^{\zeta - a}}  
\newcommand{\ProjErr}{R \kappa^{2(\zeta - 1)_+} \lambda^{(\zeta - 1)_-}} 
\newcommand{\TailFunB}{\|\fproj p_{t}^{(2)}(\ao) \ao \midFun\|_{\HK}} 

\newcommand{\midFun}{\omega_{\star}^{\lambda}}

\newcommand{\GL}{\mathcal{G}_{\lambda}} 

\newcommand{\ao}{\mathcal{U}} 
\newcommand{\aol}{\ao_{\lambda}}
\newcommand{\aoB}{\mathcal{V}} 

\newcommand{\skt}{{\bf G}} 

\newcommand{\Ss}{\mathcal{Q}}
\newcommand{\tbK}{\tilde{\bK}}

\newcommand{\eig}{\xi_{\bx,i}} 
\newcommand{\eigv}{e_{\bx,i}} 

\newcommand{\fproj}{F_{u}}

\newcommand{\bI}{\mathbf{I}}

\newcommand{\bby}{\bar{\by}}

\newcommand{\cgmt}{\omega_t}
\newcommand{\mP}{\mathbb{P}}

\newcommand{\rvo}[1]{\textcolor{black}{#1}}
\usepackage[page,title,titletoc,header]{appendix}

\begin{document}
  \title{Kernel Conjugate Gradient Methods with Random Projections}

  \author{Junhong Lin\footnote{
  		\scriptsize	Center for Data Science, Zhejiang University, Hangzhou 310027, P.R. China; Email: junhong@zju.edu.cn. } \qquad  Volkan Cevher \footnote{
  		\scriptsize	Laboratory for Information and Inference Systems,  \'{E}cole Polytechnique F\'{e}d\'{e}rale de Lausanne, CH1015-Lausanne, Switzerland; Email: volkan.cevher@epfl.ch. } 
  }

\date{}

  \maketitle \baselineskip 16pt
  \begin{abstract}
  	We propose and study kernel conjugate gradient methods (KCGM) with random projections for least-squares regression over a separable Hilbert space. Considering two types of random projections generated by randomized sketches and Nystr\"{o}m subsampling, we prove optimal statistical results with respect to variants of norms for the algorithms under a suitable stopping rule. Particularly, our results show that if the projection dimension is proportional to the effective dimension of the problem, KCGM with randomized sketches can generalize optimally, while achieving a computational advantage. As a corollary, we derive optimal rates for classic KCGM  in the well-conditioned regimes for the case that the target function may not be in the hypothesis space. 
  \end{abstract}
  \baselineskip 16pt
  
  \section{Introduction}\label{sec:introduction}
  Let the input space  be a separable Hilbert space $\HK$ with inner product
 $\la \cdot, \cdot \ra_{\HK}$, and the output space $\mR$. 
 Let $\rho$ be an unknown probability measure on $\HK \times \mR$. 
 We study the following expected risk minimization,
 \be\label{expectedRiskA} 
 \inf_{\omega \in \HK} \mcEE(\omega), \quad \mcEE(\omega) = \int_{\HK\times \mR} ( \la \omega, x \ra_{\HK} - y)^2 d\rho(x,y),
 \ee
 where the measure  $\rho$ is  known only through
 a sample $\bz =\{z_i=(x_i, y_i)\}_{i=1}^n$ of size $n\in\mN$, independently and identically distributed (i.i.d.) according to $\rho$. As noted in \cite{lin2017optimal,lin2018optimal}, this setting covers nonparametric regression with kernel methods \cite{cucker2007learning,steinwart2008support}.
 
%
 
  In the large-scale learning scenarios, the search of an approximated estimator for the above
  problem via some specific algorithms could be limited to a smaller subspace $\HKS$, in order to achieve some computational advantages \cite{williams2000using,smola2000sparse,drineas2012fast,yang2015randomized,rudi2015less}.  Typically, with a subsample/sketch dimension $m<n$, $\HKS = \overline{span\{ \tilde{x}_j : 1\leq j\leq m\}}$ where $\tilde{x}_j $ is chosen (randomly) from the input set $\bx = \{x_1,\cdots,x_n\}$ \cite{gittens2013revisiting,alaoui2015fast}, or 
  $\HKS = \overline{span\{ \sum_{j=1}^n G_{ij} x_j : 1\leq i\leq m\}}$ where $\skt = [G_{ij}]_{1\leq i\leq m, 1\leq j\leq n}$ \cite{yang2015randomized} is a general random matrix whose rows are drawn according to a distribution.  The former is called Nystr\"{o}m subsampling while the latter is called randomized sketches.\footnote{The Nystr\"{o}m subsampling scheme corresponds to a randomized sketches with the rows of the sketch matrix $\skt$ randomly chosen from the rows of an identity matrix.  In this paper, by  abuse of terminology,  we sometimes use ``randomized sketches"  to mean a sketched scheme generated by Subgaussian sketches or randomized bounded orthogonal system sketches those will be introduced in Section \ref{sec:main}.}
  Limiting the solution within the subspace $\HKS$, replacing expected risk by empirical risk over $\bz$, and combining with a  (explicit) regularized technique based on spectral-filtering of  the empirical covariance operator (which is referred as linear regularization in \cite{blanchard2010optimal}),  this leads to the projected-regularized algorithms.
  \rvo{We refer} to the previous papers \cite{alaoui2015fast,yang2015randomized,rudi2015less,kriukova2017nystrom,lin2018optimal_ske}  and references therein for the statistical results  and computational advantages of this kind of algorithms.

   In this paper, we take a different step and apply the random-projection techniques \cite{alaoui2015fast,yang2015randomized} to another efficient  iterative algorithms: kernel conjugate gradient type algorithms \cite{blanchard2010optimal}.  As noted in \cite{lin2020convergences},  a solution of the empirical risk minimization over the subspace $\HKS$ can be given by solving a projected normalized linear equation.  We apply the kernel conjugate gradient methods (KCGM)  \cite{engl1996regularization,hanke2017conjugate,blanchard2010optimal} for ``solving'' this normalized linear equation (without any explicit regularization term), and at $t$th-iteration, we get an estimator that fits the linear equation best over the $t$th-order Krylov subspace.  
   The regularization to ensure its best performance is realized by early-stopping the iterative procedure.
   
  
  Considering either randomized sketches or Nystr\"{o}m subsampling, we provide statistical results in terms of different norms with optimal rates. 
  Particularly, our results indicate that for KCGM with randomized sketches, 
  the algorithm can generalize optimally after some number of iterations, provided that the sketch dimension is proportional to the effective dimension \cite{zhang2006learning} of the problem.
  
  We point out that the computational complexities for the algorithm are $O(m^3)$ in time and $O(m^2)$ in space, which are lower than $O(n^2 t)$ in time and $O(n^2)$ in space of classic KCGM.  Thus, 
  our results suggest that KCGM with randomized sketches can generalize optimally with less computational complexities, e.g.,  $O(n^{3/2})$ in time and $O(n)$ in space without considering the benign assumptions of the problem in the attainable case (i.e. the expected risk minimization has at least one solution in $\HK$).

  Finally, as a corollary, we derive result with optimal capacity-dependent rates for classical KCGM  in the well-conditioned regimes for the non-attainable case, without requiring additional unlabeled data as in \cite{blanchard2010optimal}.
  
  

\paragraph{Related Works} The conjugated gradient  method (CGM) \cite{hestenes1952methods} is an algorithm for the numerical solution of particular systems of linear equations in optimization. 
It is also a popular and efficient tool in the inverse problem literature, in a fixed design setting (i.e., the input is deterministic while the output treated randomly), see \cite{engl1996regularization} for a comprehensive overview. The  statistical learning setting is different, as both the input and the error are stochastic. 
In the statistical learning literature, CGM has appeared under the name of partial least squares (PLS). In the latter setting, a kernelized PLS and  its variant, kernelized CGM have been developed in \cite{rosipal2001kernel} and \cite{blanchard2010optimal}, respectively.
It is now considered part of the standard toolbox of kernel methods, see \cite[Section 6.7.2]{shawe2004kernel}.
The algorithm we try to scale up in this paper is from \cite{blanchard2010optimal} in the kernel setting. Following \cite{blanchard2010optimal}, we  call our algorithm kernel CGM with random projections. In the fixed design setting, convergence results have been derived, in \cite{bach2013sharp,alaoui2015fast} for ridge regression with  Nystr\"{o}m subsampling and in  \cite{yang2015randomized} for  ridge regression with randomized sketches. 
Due to \cite{rudi2015less},  ridge regression with  Nystr\"{o}m subsampling or randomized sketches could be formulated as ridge regression with random projections. Convergence results in the statistical learning setting for linear regularization methods with random projections have been investigated in \cite{rudi2015less,kriukova2017nystrom,myleiko2019regularized,lin2018optimal_ske,lu2019analysis,lin2020convergences}.

\paragraph{Organization} The rest of this paper is organized as follows. 
We first introduce some preliminary notations and the studied algorithms in Section \ref{sec:learning}.  We then introduce some basic assumptions and state our main results with some simple discussions in Section \ref{sec:main}, \rvo{followed with some numerical illustrations in Section \ref{sec:num}.}
All the missed proofs are given in Section \ref{sec:proof} and Appendix.



%

 \section{Learning with Kernel Conjugate Gradient Methods and Random Projection}\label{sec:learning}
  In this section, we first introduce some necessary notations. We then present KCGM with projection (abbreviated as projected-KCGM), and discuss their numerical realizations considering two types of projection generated by randomized sketches and Nystr\"{o}m sketches/subsampling .
 \subsection{Notations and Auxiliary Operators}

 Let  $Z = \HK \times \mR$, $\rho_X(\cdot)$  the induced marginal measure on $\HK$ of $\rho$, and $\rho(\cdot | x)$ the conditional probability measure on $\mR$ with respect to $x \in \HK$ and $\rho$. 
 Define the hypothesis space $$\HR = \{f: \HK \to  \mR| \exists \omega \in \HK \mbox{ with } f(x) = \la \omega, x \ra_{\HK}, \rho_X \mbox{-almost surely}\}.$$  Denote $\LR$ the Hilbert space of \rvo{square-integrable} functions from $\HK$ to $\mR$ with respect to $\rho_X$, with its norm given by $\|f\|_{\rho} = \left(\int_{\HK} |f(x)|^2 d \rho_X\right)^{1\over 2}.$   Throughout this paper, we assume that 
 there exists a constant $\kappa \in [1,\infty[$, such that
 \be\label{boundedKernel} \la x,x' \ra_{\HK} \leq \kappa^2, \quad \forall x,x'\in \HK,   \rho_X\mbox{-almost every}.
 \ee
 
 For a given bounded operator $L$ mapping from a separable Hilbert space $H_1$ to another separable Hilbert space $H_2,$
  $\|L\|$ denotes the operator norm of $L$, i.e., $\|L\| = \sup_{f\in H_1, \|f\|_{H_1}=1} \|Lf\|_{H_2}$.  Let $r \in \mN_+,$ the set $\{1,\cdots,r\}$ is denoted by $[r].$
 For any real number $a$, $a_+ = \max(a,0)$, $a_- = \min(0,a)$.
 
 Let $\IK: \HK \to \LR$ be the linear map $\omega \to \la \omega, \cdot \ra_{\HK}$, which is bounded by $\kappa$ under Assumption \eref{boundedKernel}. Furthermore, we consider the adjoint operator $\IK^*: \LR \to \HK$, the covariance operator $\TK: \HK \to \HK$ given by $\TK = \IK^* \IK$, and the integral operator $\LK : \LR \to \LR$ given by $\IK \IK^*.$ It can be easily proved that $$ \IK^*g = \int_{\HK} x g(x) d\rho_X(x), $$  $$\LK f = \IK \IK^* f = \int_{\HK} f(x) \la x, \cdot \ra_{\HK} d \rho_X(x), \quad \mbox{and}$$
 $$\TK  = \IK^* \IK= \int_{\HK} \la \cdot , x \ra_{\HK} x d \rho_X(x).$$  
 Under Assumption \eqref{boundedKernel},
 the operators $\TK$ and $\LK$ can be proved to be positive trace class operators (and hence compact):
 \be\label{eq:TKBound}
 \begin{split}
 	\|\LK\| = \|\TK\| \leq \tr(\TK) = \int_{\HK} \tr(x \otimes x)d\rho_X(x) =  \int_{\HK} \|x\|_{\HK}^2 d\rho_{X}(x) \leq \kappa^2.
 \end{split}
 \ee
 For any $\omega \in \HK$,
 one can prove the following isometry property \rvo{ \cite{steinwart2008support}},
 \be\label{isometry}
 \|\IK \omega \|_{\rho} = \|\sqrt{\TK} \omega\|_{\HK},
 \ee
 Moreover, according to the singular value decomposition of a compact operator, one can prove
 \be\label{eq:rho2hk}
 \|\LK^{-{1\over2 }}\IK \omega\|_{\rho} \leq \|\omega\|_{\HK}.
 \ee
 Similarly,  for all $f \in \LR, $ there holds,
 \be\label{isometryB}
 \|\IK^* f \|_{\HK} = \|\LK^{1\over 2} f \|_{\rho},\quad \mbox{and}
 \ee
  \be\label{eq:hk2rho}
 \|\TK^{-{1\over 2}}\IK^* f \|_{\HK} \leq  \| f \|_{\rho}.
 \ee
 
 We define the (normalized) sampling operator $\SX: \HK \to \mR^n$ by $$(\SX \omega)_i = {1 \over \sqrt{n}}\la \omega, x_i \ra_{\HK},\quad  i \in [n].$$ 
 Its adjoint operator $\SX^*: \mR^n \to \HK,$ defined by $\la \SX^*{\bf y}, \omega \ra_{\HK} = \la {\bf y}, \SX \omega\ra_{2}$ for ${\bf y} \in \mR^n$ is thus given by $$\SX^*{\bf y} = {1 \over \sqrt{n}} \sum_{i=1}^n y_i x_i.$$ \rvo{ Here, the norm $\|\cdot\|_{2}$ in $\mR^n$ is the usual Euclidean norm.}
 For notational simplicity, we also denote $\bby = {1\over \sqrt{n}} \by.$
  Moreover, we can define the empirical covariance operator $\TX: \HK \to \HK$ such that $\TX = \SX^* \SX$. Obviously,
 $$
 \TX = \SX^* \SX = {1 \over n} \sum_{i=1}^n \la \cdot, x_i \ra_{\HK} x_i.
 $$
 By Assumption \eqref{boundedKernel}, similar to \eref{eq:TKBound}, we have
 \be\label{eq:TXbound}
 \|\TX\| \leq \tr(\TX) \leq \kappa^2.
 \ee
 Denote  $\bK_{\bx \tilde{\bx}} $ the $|\bx| \times |\tilde{\bx}|$ matrix with its $(i,j)$-th entry given by
 $ {1 \over \sqrt{|\bx| |\tilde{\bx}|} }\la x_i, \tilde{x}_j \ra_{\HK}$ for any two input sets $\bx$ and  $\tilde{\bx}.$  Obviously, 
 \be\label{eq:kmatrix}
 \bK_{\bx \tilde{\bx}} =  \SX \SXS^* = {1 \over \sqrt{|\bx| |\tilde{\bx}|} } \left[ \la x_i, \tilde{x}_j \ra_{\HK}  \right]_{i\in [|\bx| ] ,j \in [\tilde{\bx}]}.
 \ee
 
 We \rvo{ rewrite Problem \eqref{expectedRiskA} as}
 \be\label{expectedRisk}
 \inf_{f \in \HR} \mcE(f), \quad \mcE(f)= \int_{\HK \times \mR} ( f(x) - y)^2 d\rho(x,y)\rvo{.}
 \ee
 The function that minimizes the expected risk over all measurable functions is the
 regression function \cite{cucker2007learning,steinwart2008support}, defined as,
 \be\label{regressionfunc}
 f_{\rho}(x) = \int_{\mR} y d \rho(y | x),\qquad x \in \HK,  \rho_X\mbox{-almost every} .
 \ee
 A simple calculation shows that the following  well-known fact holds \cite{cucker2007learning,steinwart2008support}, for all  $f \in \LR,$
 $$
 \mcE(f) - \mcE(\FR) = \|f - \FR\|_{\rho}^2.
 $$
 Under Assumption \eqref{boundedKernel},  $\HR$ is a subspace of $\LR.$
 Thus a  solution $\FH$ for the problem   \eqref{expectedRisk} is the projection of the regression function $f_{\rho}$ onto the closure of $\HR$ in $\LR$, and for all $f\in \HR$ (e.g. \cite{lin2017optimal}),
 \be\label{frFH}
 \IK^* f_{\rho} = \IK^* \FH, \quad \mbox{and}
 \ee
 \be\label{eq:exceRisk}
 \mcE(f) -  \mcE(\FH) = \|f - \FH\|_{\rho}^2.
 \ee

 \subsection{Kernel Conjugate Gradient Methods with Projection}\label{subsec:kcgm}
In this subsection, we introduce KCGM with solutions restricted to the subspace $\HKS$, a closed subspace of $\HK$. 

Let $\proj$ be the \rvo{orthogonal} projection operator with its range $\HKS$. 
As noted in \cite{lin2020convergences}, \rvo{a solution  for the (unpenalized) empirical risk minimization over $\HKS$ is given by $\hat{\omega} \in\HKS$  such that}
\be\label{eq:NormEq}
\ao \hat{\omega}= \proj \SX^*\bby, 
\ee
where for notational simplicity, we denote
\be \label{eq:ao}
\rvo{\ao = \proj \TX \proj.}
\ee

As $\TX = \SX^* \SX$,
$\proj \TX \proj  = \proj \SX^* \SX \proj = (\SX\proj)^* \SX \proj$.
Thus, \eqref{eq:NormEq} could be viewed as a normalized equation of 
$\SX \proj \omega = \bby.$
\rvo{ Motivated by \cite{engl1996regularization,hanke2017conjugate,blanchard2010optimal},} in this paper, we study the conjugate gradient type algorithm for ``solving" this normalized equation, \rvo{in combination with a suitable stopping rule.
The conjugate gradient type algorithm is a computationally efficient scheme to approximately solve linear equation such as \eqref{eq:NormEq}.  The principle of  conjugate gradient type algorithm is to restrict the problem to a nest  set of subspace, the so-called Krylov subspace, defined as
$$\mcK_t(\ao, \proj \SX^* \NOutputs) = \mbox{span}\{\proj\SX^*\NOutputs, \ao \proj\SX^*\NOutputs, \cdots, \ao^{t-1} \proj\SX^*\NOutputs\} = \{p(\ao) \proj \SX^* \NOutputs: p \in\mcP_{t-1}\} ,$$
where $\mcP_{t-1}$ denotes the set of real polynomials of degree at most $t-1$. The  algorithm  we study in this paper is detailed as follows.}

\begin{alg}[Projected-KCGM]\label{alg:1}
	For any $t = 1,\cdots, T,$
\be\label{eq:alg}
\omega_t = \argmin_{\omega \in \mcK_t(\ao, \proj\SX^*\NOutputs) } \|\ao \omega - \proj\SX^* \bby\|_{\HK}.
\ee
\end{alg}

 \rvo{
 At $t$-th iteration, the algorithm finds an approximated solution of \eqref{eq:NormEq} via a residual least squares optimization restricted to the order-$t$ Krylov subspace generated by the operator $\ao$ and the vector $\proj\SX^*\NOutputs$.  Note that Algorithm \ref{alg:1} appears to be a problem over an infinite-dimensional Hilbert space, which could not be directly realized on the computer. In what follows, we show that Algorithm \ref{alg:1}  is equivalent to a  CGM  applied to a finite-dimensional linear equation, thanks to the so-called representation theorem similar to those for kernel ridge regression, considering three different special schemes.}
 

Different choices on the subspace $\HKS$ correspond to different algorithms. When $\proj = I$, the
algorithm reduces to classical KCGM and its statistical result has been investigated in \cite{blanchard2010optimal}.  In this case,
we have the following representation theorem for Algorithm \ref{alg:1}.

\begin{Exa}(Non-sketches)\label{exa:nonSke}
	For the ordinary non-sketching regimes, $\HKS = \HK$.
	Let $\bK = \SX \SX^*.$ Then Algorithm \ref{alg:1} 
	is equivalent to $\omega_t = {1 \over \sqrt{n}} \sum_{i=1}^n (\ba_t)_i x_i,$ with ${\ba}_t$ given by
	$$
	\ba_t = \argmin_{\ba \in \mcK_t(\bK, \NOutputs) } \|\bK \ba - \NOutputs\|_{\bK}.
	$$
	\rvo{This algorithm has been investigated in \cite{blanchard2010optimal}.  }
\end{Exa}
\rvo{
The coefficients $\{\ba_t\}$ in the above example  can be computed using a simple iterative algorithm, using forward multiplication of vectors by the matrix $\bK$ for each iteration. We refer to \cite{blanchard2010optimal} for further details.}

\rvo{To reduce the computational complexity of Example \ref{exa:nonSke}, we apply the so-called random projections technique, that have been used to scale up the classical linear regularization methods
 \cite{gittens2013revisiting,alaoui2015fast,yang2015randomized,rudi2015less,lin2020convergences}, to classical KCGM. } The basic idea is to restrict the solution to a smaller subspace $\HKS$, either with 
$$\HKS = \overline{span\{ \sum_{j=1}^n G_{ij} x_j : 1\leq i\leq m\}}$$ where $\skt = [G_{ij}]_{1\leq i\leq m, 1\leq j\leq n}$ is a  random matrix  \rvo{\cite{yang2015randomized}}, or 
 $$\HKS = \overline{span\{ \tilde{x}_j : 1\leq j\leq m\}}$$ with $\tilde{x}_j$ chosen (randomly) from $\bx$ \rvo{\cite{gittens2013revisiting,alaoui2015fast}. }
The following examples provide numerical realizations of Algorithm \ref{alg:1} for these two regimes. 

\begin{Exa}[Randomized sketches]\label{exa:ranSke}
	Let $\HKS = \overline{span\{ \sum_{j=1}^n G_{ij} x_j : 1\leq i\leq m\}}$, and $\skt = [G_{ij}]$ be a  matrix in $\mR^{m \times n}$ \rvo{\cite{yang2015randomized}.  }
	\rvo{Let  $\bK_{\bx \bx}$ be as in \eqref{eq:kmatrix},
	$\bR \in \mR^{m \times r}$ be a full column-rank matrix such that $\bR \bR^{\top} = (\skt \bK_{\bx \bx} \skt^\top)^{\dag}$. }  Let
	$\tbK =  \bR^\top \skt \bK_{ \bx \bx}^2 \skt^\top \bR$ and
	$\bb = \bR^\top \skt \bK_{\bx\bx} \bby.$
	In this case, Algorithm \ref{alg:1} is equivalent to $\omega_t =   {1 \over \sqrt{n}}\sum_{i=1}^n  (\skt^\top \bR {\ba}_t)_i x_i$ with ${\ba}_t$ given by 
\be \label{eq:psealg}
	{\ba}_t = \argmin_{\ba \in \mcK_t(\tbK, \bb)} \| \tbK \ba  -  \bb \|_2.
	\ee
	We call this type of algorithm sketched-KCGM.
\end{Exa}

\begin{Exa}[Subsampling sketches]\label{exa:nysSke}
	In Nystr\"{o}m-subsampling sketches,  $\tilde{\bx} = \{\tilde{x}_1,\cdots, \tilde{x}_m\}$ with each $\tilde{x}_j$
	drawn (randomly following a distribution) from $\bx$ \rvo{\cite{gittens2013revisiting,alaoui2015fast}.  Let $\bK_{\tilde{\bx} \bx}$ and $\bK_{ \tilde{\bx} \tilde{\bx}}$ be as in \eqref{eq:kmatrix}.} Let
	$\bR \in \mR^{m \times r}$ be a full column-rank matrix such that $\bR \bR^{\top} = \bK_{\tilde{\bx}\tilde{\bx}}^{\dag}$.  Let
	$\tbK =  \bR^\top \bK_{\tilde{\bx} \bx} \bK_{\tilde{\bx} \bx} ^{\top} \bR$ and $\bb = \bR^\top \bK_{\tilde{\bx} \bx} \bby.$
	In this case, Algorithm \ref{alg:1} is equivalent to $\omega_t =  {1 \over \sqrt{m}}\sum_{i=1}^m  (\bR {\ba}_t)_i \tilde{x}_i$ with ${\ba}_t$ given by 
	$$
	{\ba}_t = \argmin_{\ba \in \mcK_t(\tbK, \bb)} \| \tbK \ba  -  \bb \|_2.
	$$
	We call this  algorithm Nystr\"{o}m-KCGM.
\end{Exa}

The proofs for the above examples are postponed in Section \ref{sec:proof}.
In all the above examples, in order to execute the algorithms, one only needs to know
how to compute $\la x, x'\ra_{\HK}$ for any two points $x,x' \in \HK$, which is met by many cases such as learning with kernel methods.

The optimization criterion for sketched/Nystr\"{o}m KCGM in the last two examples can be computed by a simple iterative algorithm, using forward multiplication of vectors by $\tbK$ for each iteration, and it is not hard to show that \be \rvo{ \tbK \ba_r = \bb}, \label{eq:zero}
\ee using this optimization formulation, see e.g., \cite[Chapter 2]{hanke2017conjugate}. Note also that $r$ is always less than $m$.

In general, as that the computation of the matrix $\skt \bK_{ \bx \bx}  = {1 \over \sqrt{n}} [\skt \bK_{\bx x_1}, \skt \bK_{\bx x_2},\cdots, \bK_{\bx x_n}]$ (or $\bK_{ \bx \tilde{\bx}} \bR$) can be parallelized \cite{yang2015randomized}, the computational costs are
$O(m^3 + m^2 T)$ in time and $O(m^2)$ in space
for sketched/Nystr\"{o}m KCGM after $T$-iterations, while they are
$O(n^2 T)$ in time and $O(n^2)$ in space for non-sketched KCGM. As shown both in theory and our numerical results, the total number of iterations $T$ for the algorithms to achieve best performance is typically less than $r (\leq m)$ for sketched/Nystr\"{o}m KCGM.
 
A classical \cite{rosipal2001kernel} or sketched \cite{avron2017faster,rudi2017falkon} kernel conjugate gradient type algorithm was proposed for solving the penalized empirical risk minimization. In contrast, Algorithm \ref{alg:1} is for ``solving"  the (unpenalized) empirical risk minimization and it does not involve any explicit penalty.  In this case, we do not need to tune the penalty parameter. The best generalization ability of  Algorithm \ref{alg:1} is ensured by early-stopping the procedure, considering a suitable stopping rule.
 
The proofs for the three examples will be given in Subsection \ref{subsect:proof}.

\section{Main Results}\label{sec:main}
In this section, we first introduce some assumptions from statistical learning theory then present our statistical results for sketched/Nystr\"{o}m-KCGM and classical KCGM.

\subsection{Assumptions}
\rvo{We recall the considered statistical learning model, where each sample $(x_i,y_i)$ is independently and identically drawn from the data distribution $\rho$ over $\HK \times \mR$. With the definition of $\FR$, we can write $y_i = \FR(x_i) + \epsilon_i$, where $\epsilon_i$ is the so-called noise. Note that our setting covers non-parametric regression problems over a reproducing kernel Hilbert space, as noted in \cite{lin2017optimal}. See Appendix \ref{sec:learningKernel} for further details.} 

 The first assumption relates to a Bernstein-type moment
condition on the output value $y$.
\begin{as}\label{as:noiseExp}
	There exist positive constants $Q$ and $M$ such that for all $l \geq 2$ with $l \in \mN,$
	\be\label{noiseExp}
	\int_{\mR} |y| ^{l} d\rho(y|x) \leq {1 \over 2} l! M^{l-2} Q^2, 
	\ee
	$\rho_{ X}$-almost surely. 
	Furthermore, for some $B>0$, $\FH$ satisfies  \be\label{eq:FHFR}
	\int_{\HK} (\FH(x) - \FR(x))^2 x \otimes x d \rho_X(x) \preceq B^2 \TK,
	\ee 
\end{as}
Obviously,  \eqref{noiseExp} is satisfied if 
$y$ is bounded almost surely or $y = \la \omega_{*},x\ra_{\HK} + \epsilon$ for some Gaussian noise $\epsilon.$ 
It implies that the regression function $\FR$
is bounded almost surely, as
\be\label{eq:bounRegFunc}
|\FR(x)| \leq \int_{\mR} |y| d\rho(y|x) \leq \left(\int_{\mR} |y|^2 d\rho(y|x)\right)^{1\over 2} \leq Q. 
\ee
\eqref{eq:FHFR} is satisfied if $\FH -\FR$ is bounded almost surely. \rvo{ Moreover, when making a consistency assumption, i.e., $\inf_{\HR} \mcE = \mcE(\FR)$, as that in \cite{smale2007learning,caponnetto2007optimal,steinwart2009optimal} for kernel-based methods, it is satisfied with $B=0$.}

\rvo{Recall that $\FH$ is the projection of the regression function $\FR$ onto the closure of $\HR$ in $\LR$. It is easy to see
that the search for a solution of Problem \eqref{expectedRisk} is equivalent to the search of a linear function in $\HR$
to approximate $\FH$. From this point of view, bounds on the excess risk of a learning algorithm
on $\HR$ or $\HK$, naturally depend on the following assumption, which quantifies how well, the target
function $\FH$ can be approximated by $\HR$.}
  \begin{as}\label{as:regularity}
	$\FH$ satisfies   the following H\"{o}lder source condition
	\be\label{eq:socCon}
	\FH = \LK^{\zeta} g_0, \quad \mbox{with}\quad  \|g_0\|_{\rho} \leq R.
	\ee
	Here, $R$ and $\zeta$ are non-negative numbers.
\end{as}
\rvo{Assumption \ref{as:regularity} relates to the regularity/smoothness of $\FH$ in non-parametric regression \cite{cucker2007learning}. } The bigger the $\zeta$ is, the stronger the assumption is, the smoother $\FH$ is, as $$\LK^{\zeta_1}(\LR) \subseteq \LK^{\zeta_2}(\LR)\quad \mbox{when } \zeta_1 \geq \zeta_2.$$
Particularly, when $\zeta \geq 1/2$, there exists some $\omega_{\HK} \in \HK$ such that $\IK \omega_{\HK} = \FH$ almost surely \cite[Page 151]{steinwart2008support},
 while for $\zeta = 0,$ the assumption holds trivially. 
 \rvo{When $\zeta \geq 1/2$, we call it the attainable case, while for $\zeta<1/2$, we call it the non-attainable case.}

  \begin{as}\label{as:eigenvalues}	
	For some $\gamma \in [0,1]$ and $c_{\gamma}>0$, $\TK$ satisfies
	\be\label{eigenvalue_decay}
\mcN(\lambda):=	\tr(\TK(\TK+\lambda I)^{-1})\leq c_{\gamma} \lambda^{-\gamma}, \quad \mbox{for all } \lambda>0.
	\ee
\end{as}
Assumption \ref{as:eigenvalues} characters the capacity of  $\HK.$
The left-hand side of \eqref{eigenvalue_decay} is called the effective dimension \cite{zhang2006learning}.  As $\TK$ is a trace-class operator,
Condition \eqref{eigenvalue_decay}  is trivially satisfied with $\gamma = 1$ (which is called the capacity-independent case). 
\rvo{Assumption
	\eqref{eigenvalue_decay} with $ \gamma \in [0, 1]$ allows to derive better error rates.
 It is satisfied
with a  general $\gamma \in(0,1]$ if the eigenvalues $\{\lambda_i\}$ of $\TK$ satisfy $\lambda_i \sim i^{-{1 \over \gamma}},$ or $\gamma = 0$ when $\TK$ is finite rank.}

We refer to \cite{smale2007learning,caponnetto2007optimal,cucker2007learning,steinwart2008support,lin2018optimal} for more comments on the above assumptions.
\subsection{General Results for Kernel Conjugate Gradient Method with Projection}
The following results provide convergence results for general projected-KCGM with a data-dependent stopping rule. 
\begin{thm}\label{thm:gen}
	Under Assumptions \ref{as:noiseExp},  \ref{as:regularity} and \ref{as:eigenvalues}, let $a \in [0, {\zeta} \wedge {1\over 2}]$. 
	Denote 
	\be\label{eq:bnGam} 
	b_{n,\zeta, \gamma} =
	(1 \vee \log n^{\gamma})^{{\bf 1}_{\{2\zeta+\gamma \leq 1\}}} .
	\ee
	Assume that for some $C_1'\geq 1$, 
	and for any $\delta\in (0,1)$, 
	\be\label{eq:projCond}
	\mathbb{P} \left( \| (I - \proj) \TK^{1\over 2}\|^2   > C_1'\lambda^{1\vee\zeta - a \over 1 - a}  \log {2 \over \delta} \right) \leq \delta, \quad \lambda = n^{-{1\over (2\zeta+\gamma) \vee 1}} b_{n,\zeta, \gamma} .
	\ee	Then the following results hold with probability at least $1-\delta$.
	There exist positive constants  $\tilde{C}_1$ and $\tilde{C}_2$ (which depend only on 
	$\zeta,\gamma,c_{\gamma}, \|\TK\|,\kappa^2, M, Q,B, R, C_1'$) such that \rvo{ if $\hat{t}$ is the first iteration satisfying the following stopping rule,}
	\be \label{eq:stopr}
	\| \ao \omega_{t} - \proj\SX^* \NOutputs\|_{\HK} \leq  \tilde{C}_1\log^{3 \over  2} {2 \over  \delta} n^{-{\zeta+1/2 \over 1\vee (2\zeta +\gamma)}}  b_{n,\zeta,\gamma}^{\zeta+ 1/2},
	\ee
	then
	$$
	\|\LK^{-a}(\IK \omega_{\hat{t}} - \FH) \|_{\rho} \leq \tilde{C}_2  \log^{2 - a} {2 \over  \delta} n^{-{\zeta - a \over 1\vee(2\zeta +\gamma)}} b_{n,\zeta,\gamma}^{\zeta -a}  .
	$$
	Furthermore, if $\zeta\geq {1/2},$ $\FH = \IK \omega_{\HK}$ for some $\omega_{\HK} \in \HK$ and 
	\be\label{eq:resHK}
	\|\TK^{{1\over 2} - a} (\omega_{\hat{t}} - \omega_{\HK}) \|_{\HK} \leq \tilde{C}_2  \log^{2 - a} {2 \over  \delta} n^{-{\zeta - a \over 1\vee (2\zeta +\gamma)}}.
	\ee

\end{thm}

The convergence rate from the above is optimal as it matches the minimax lower rate $O(n^{-{\zeta - a \over 2\zeta +\gamma}})$  derived for $2\zeta +\gamma >1$ in \cite{caponnetto2007optimal,steinwart2009optimal,blanchard2016optimal}.  \rvo{ Note that 
	$$
	b_{n,\zeta, \gamma} \leq  \begin{cases}
	1, &\mbox{if} \  \gamma = 0 \  \mbox{or} \ 2\zeta+\gamma >1, \\
	\log n, &\mbox{otherwise}.
	\end{cases}
	$$
}

 \rvo{Theorem \ref{thm:gen} provides convergence results with different error measures for the studied algorithms. When $a = 0, $ $\|\LK^{-a}(\IK \omega_{\hat{t}} - \FH) \|_{\rho} = \|\IK \omega_{\hat{t}} - \FH\|_{\rho}^2 = \tilde{\mcE}(\omega_{\hat{t}}) - \inf_{\HK}\tilde{\mcE} $, which is the so-called prediction error in statistical learning theory. When $a=1/2,$ (assuming that $\zeta \geq 1/2$), the error measure $\|\LK^{-a}(\IK \omega_{\hat{t}} - \FH) \|_{\rho}  = \| \omega_{\hat{t}} - \omega_{\HK} \|_{\HK} $ is in particular interesting to the inverse problems community. For $a \in (0,1/2)$, the error measures is between the prediction error and $\HK$-norm error. A bigger $a$ leads to a ``stronger" convergence result but with a slower convergence rate.}

Theorem \ref{thm:gen} asserts that projected-KCGM converges optimally if the projection error is small enough. The condition \eqref{eq:projCond} is satisfied with random projections induced by randomized sketches or Nystr\"{o}m subsampling if the sketching dimension is large enough, as shown in Section \ref{sec:proof}.  Thus we have the following corollaries for sketched or \rvo{Nystr\"{o}m} KCGM.

\subsection{Results for Kernel Conjugate Gradient Methods with Randomized Sketches}
In this subsection, we state optimal convergence results with respect to different norms for KCGM with randomized sketches from Example \ref{exa:ranSke}. 

We assume that the sketching matrix $\skt$ satisfies the following concentration property:
 For any finite subset $E$ in  $\mR^n$ and for any $t>0,$ 
\be \label{eq:isoPro}
\mP (|\|\skt\ba \|_2^2 - \|\ba\|_2^2 | \geq t \|\ba\|_2^2 : \exists \ba \in E) \leq 2  |E|\mathrm{e}^{-t^2m \over  c_0'\log^{\beta} n}.
\ee
Here, $c_0'$ and $\beta$ are universal non-negative  constants.
\begin{Exa} Many matrices satisfy the concentration property.  \\
	1) {\bf Subgaussian sketches.}
 Matrices with i.i.d. subgaussian (such as Gaussian or Bernoulli) entries satisfy \eqref{eq:isoPro} with some universal constant $c_0'$ and $\beta = 0$.   More general, if the rows of $\skt$ are
independent (scaled) copies of an isotropic $\psi_2$ vector, then $\skt$ also satisfies \eqref{eq:isoPro} \cite{mendelson2008uniform}.  Recall that a random vector $\ba\in\mR^n$ is $\psi_2$ isotropic if for all $\mathbf{v} \in\mR^n$
$$
\mE[\la \ba,  \mathbf{v} \ra_2^2] = \|\mathbf{v}\|_2^2,  \quad \mbox{and}  \quad \inf\{t: \mE[\exp(\la \ba,  \mathbf{v} \ra_2^2/t^2)] \leq2 \} \leq \alpha \|\mathbf{v}\|_2,
$$
for some constant $\alpha$. \\
2) {\bf Randomized orthogonal system (ROS) sketches.}  As noted in \cite{krahmer2011new}, 
 matrix that satisfies restricted isometric property  from compressed sensing \cite{candes2005decoding} with randomized column signs satisfies \eqref{eq:isoPro}. Particularly,  random partial Fourier matrix, or random partial Hadamard matrix with randomized column signs (after scaling) satisfies \eqref{eq:isoPro} with $\beta = 4$ for some universal constant $c_0'$.
\end{Exa}

%
%

\begin{corollary}\label{cor:RanSke}
	Under Assumptions \ref{as:noiseExp},  \ref{as:regularity} and \ref{as:eigenvalues}, let 
	$\HKS =  \overline{range\{\SX^* \skt^\top\}},$ where  $\skt \in \mR^{m \times n}$ is a random matrix satisfying \eqref{eq:isoPro}. 
	Let  $\delta\in (0,1)$, $a \in [0, {\zeta} \wedge {1\over 2}]$ and
	\be\label{eq:mnum}
	m \geq \tilde{C}_3 \log^3{3\over \delta} \log^{\beta} n  \begin{cases}
		n^{\gamma}[1\vee \log n^{\gamma}]^{-\gamma},  & \mbox{if } 2\zeta+\gamma \leq 1,\\
		n^{\gamma(\zeta-a) \over (1-a)(2\zeta+\gamma)} ,  & \mbox{if } \zeta \geq 1, \\
		n^{\gamma \over 2\zeta+\gamma}    & \mbox{otherwise},
	\end{cases}
	\ee
	 for some $\tilde{C}_3>0$ (which depends only on
	$\zeta,\gamma,c_{\gamma}, \|\TK\|,\kappa^2, M, Q,B, R, c_0').$
	Then  the conclusions in Theorem \ref{thm:gen} hold.
\end{corollary}

\rvo{When $1 - \gamma < 2\zeta \leq 2$, the optimal choice of the explicit regularization parameter $\lambda_\star$ for a linear regularized algorithm (e.g.  \cite{caponnetto2007optimal,lins2018optimal}) is $\lambda_\star =n^{-1 \over 2\zeta+\gamma} $, and the effective dimension for the studied problem is $\mcN(\lambda_\star) \lesssim O(n^{\gamma \over 2\zeta+\gamma})$ by Assumption \ref{as:eigenvalues}.}
Thus,  the  minimal sketching dimension in the case of $1 - \gamma < 2\zeta \leq 2$ is proportional to the effective dimension up to a logarithmic factor. 

According to Corollary \ref{cor:RanSke}, sketched-KCGM can generalize optimally if the sketching dimension is large enough.

\subsection{Results for Kernel Conjugate Gradient Methods with Nystr\"{o}m Sketches}
In this subsection, we provide optimal rates with respect to different norms for KCGM with Nystr\"{o}m sketches from  Example \ref{exa:nysSke}. The first result provides convergence of the algorithm for plain Nystr\"{o}m sketches.

\begin{corollary}\label{thm:NySub}
	Under Assumptions \ref{as:noiseExp},  \ref{as:regularity} and \ref{as:eigenvalues}, let $\HKS = \overline{span\{x_1,\cdots,x_m\}}$,
	 $2\zeta+ \gamma>1$, $\delta\in (0,1)$, $a \in [0, {\zeta} \wedge {1\over 2}]$ and
	$$
	m \geq \tilde{C}_4 n^{1 \vee \zeta  - a \over (1 - a)(2\zeta+\gamma)}  [1 \vee \log n^{\gamma}],
	$$
	for some $\tilde{C}_4>0$ (which depends only on
	$\zeta,\gamma,c_{\gamma}, \|\TK\|,\kappa^2, M, Q,B, R).$ Then the conclusions in Theorem \ref{thm:gen} are true.
\end{corollary}

The requirement on the \rvo{projection dimension} $m$ of (plain) Nystr\"{o}m-KCGM
does not depend on the probability constant $\delta$. It is stronger than that of sketched-KCGM if $\gamma<1$ (ignoring  the factor $\delta$),  which is consistent with the suboptimality example on Nystr\"{o}m ridge regression in \cite{yang2015randomized}.

\begin{rem}\label{rem:alsNy}
	In the above, we only consider the plain Nystr\"{o}m subsampling. Using the approximated leveraging score (ALS) Nystr\"{o}m subsampling \cite{drineas2012fast,alaoui2015fast}, we can further improve the projection dimension condition to \eqref{eq:mnum}, see Section \ref{sec:proof} for details. However, in this case, we need to compute the ALS with an appropriate pseudo regularization parameter $\lambda$. 
\end{rem}

\begin{rem}
	a) The stopping rule  \eqref{eq:stopr} is a so-called discrepancy stopping rule\rvo{, assuming we know exactly the constant} . It is similar to \cite{blanchard2010optimal} for classical KCGM. The realization of sketched/Nystr\"{o}m KCGM is given by Example \ref{exa:ranSke}/\ref{exa:nysSke}. According to the coming equation, \eqref{eq:interm4}, 
	the left-hand side of \eqref{eq:stopr} in this case is $\|\tbK \ba_t - \bb\|_2$. Thus $\hat{t}$ is always less than $r(\leq m)$ under this discrepancy stopping rule, since we have \eqref{eq:zero}. \\
	b) The  discrepancy stopping rule \rvo{requires \rvo{a priori knowledge of the constant in \eqref{eq:stopr}}. In practice, one may use a cross-validation(CV) approach to tune this constant} or to tune the number of iterations \rvo{$\hat{t}$}, and it is possible to develop related theoretical results using a similar argument from \rvo{\cite{caponnetto2010cross,lins2018optimal}.   
	For subsampling sketches,
 if there are $V$ samples in the validation dataset,
	then the additional computational cost for tuning the number of iterations is at most $O(m \hat{t}_{cv} V) (\leq O(m^2 V))$.}
	\rvo{We leave this for a future study.} \\
	\\
	\end{rem}

\subsection{Optimal Rates for Classical Kernel Conjugate Gradient Methods}
As a direct corollary, we derive optimal rates  for classical KCGM as follows, covering the non-attainable cases.

\begin{corollary}\label{thm:CGM}
	Under Assumptions \ref{as:noiseExp},  \ref{as:regularity} and \ref{as:eigenvalues},  let $\proj = I,$ $\delta\in(0,1)$ and $a \in [0,\zeta \wedge {1\over 2}]$ .  Then the conclusions in Theorem \ref{thm:gen} are true.
\end{corollary}

Convergence results with optimal rates for KCGM have been  derived in \cite{blanchard2010optimal} for both the attainable and non-attainable cases. But the results in \cite{blanchard2010optimal}
for the non-attainable cases require extra unlabeled data. To the best of our knowledge, our results provide the first optimal capacity-dependent rate for KCGM in the well-conditioned regimes for the non-attainable case (i.e. \rvo{$\zeta < 1/2$}), without requiring unlabeled data.

Convergence results for kernel partial least squares
under different stopping rules have been derived in \cite{lins2018optimal,singer2017kernel}, but the derived optimal rates are only for the attainable cases.  Our analysis could be  extended to this different type of algorithm with similar stopping rules.

All the results stated in this section will be proved in Section \ref{sec:proof}.

\section{Numerical Simulations} \label{sec:num}

\begin{figure}[H]
	\centering
	N	\includegraphics[width=\textwidth]{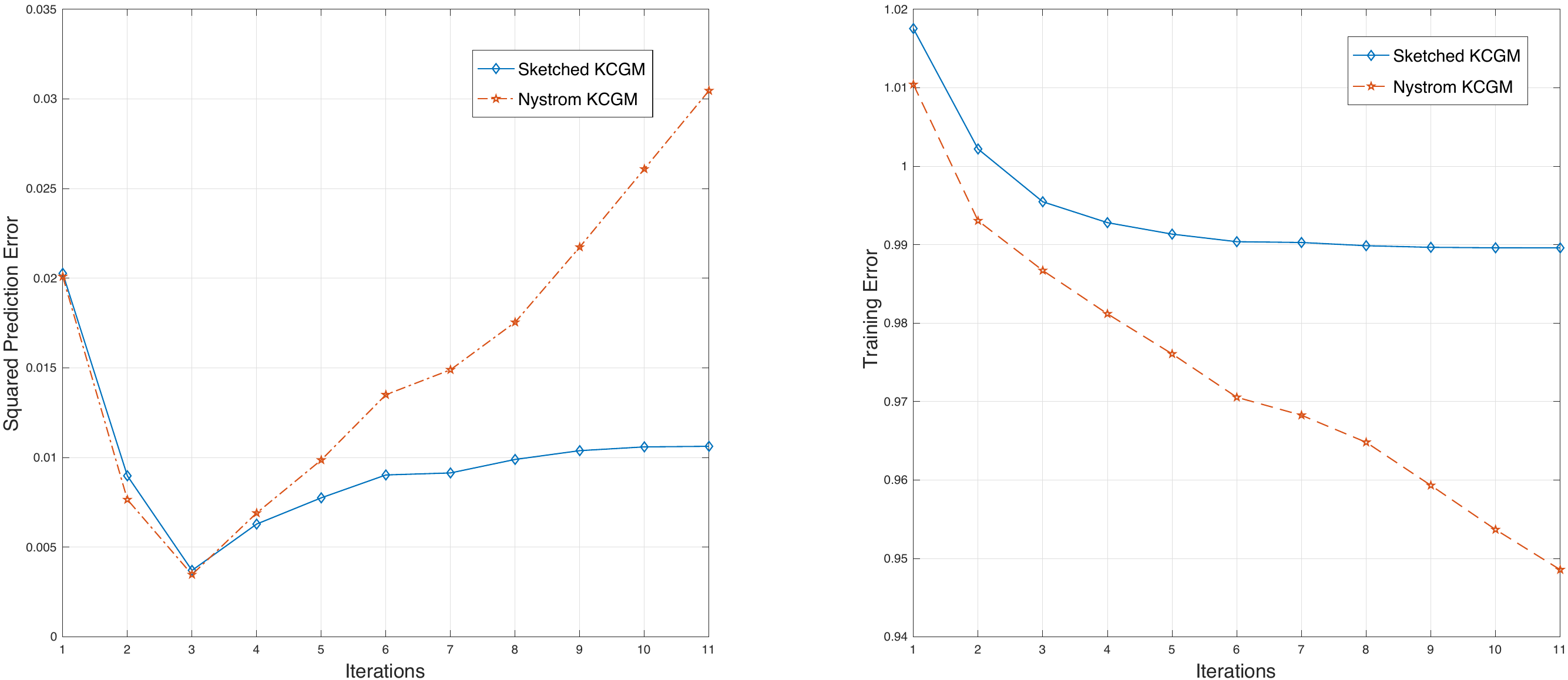}
	\caption{Squared prediction errors and training errors for sketched KCGM with $m = \lceil n^{1/3}\rceil$ and plain \rvo{Nystr\"{o}m} KCGM with $m = \lceil n^{2/3} \rceil$ and $n = 1024.$ }\label{fig:1}
\end{figure}
\begin{figure}
	\centering
	\includegraphics[width=\textwidth]{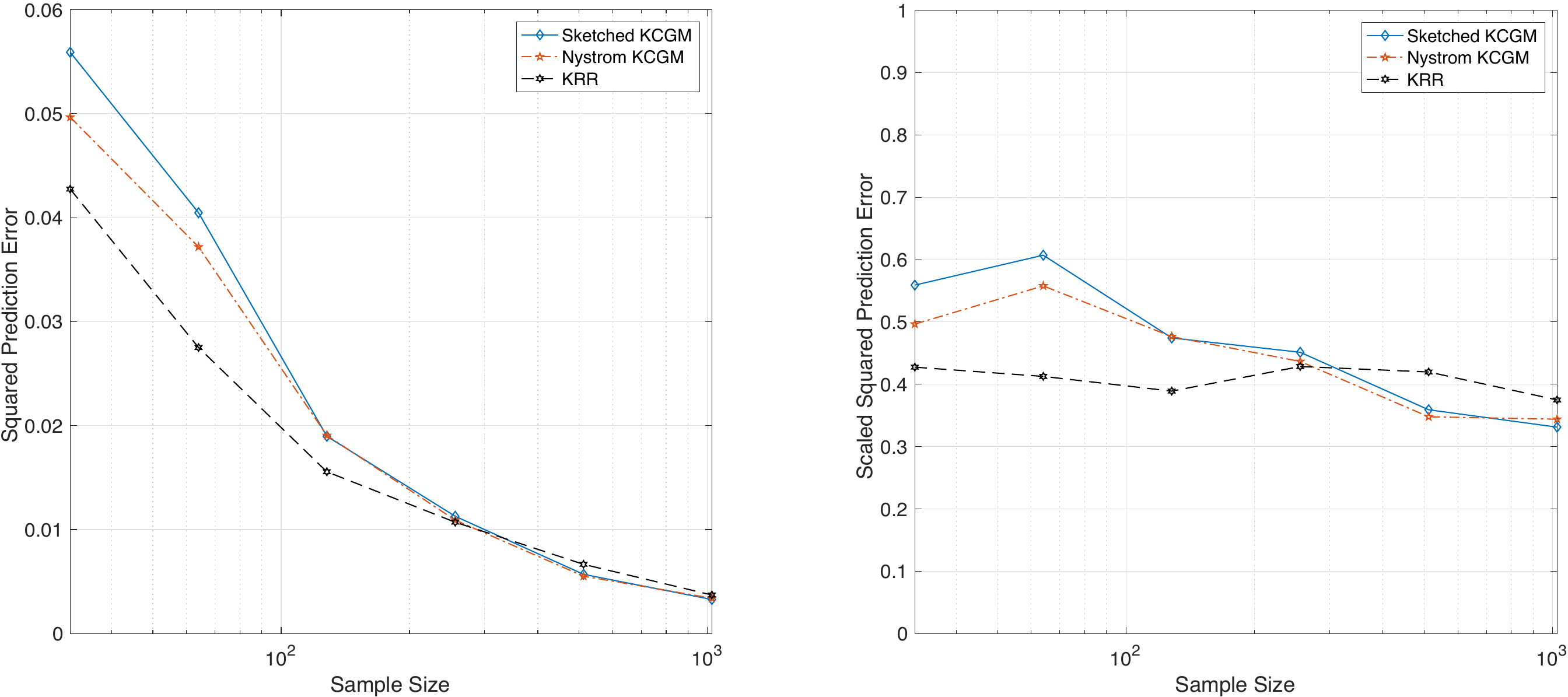}
	\caption{Prediction errors ($ \|\IK \hat{\omega} - \FR\|_{\rho}^2$)
		and scaled prediction errors ($ n^{2/3} \|\IK \hat{\omega} - \FR\|_{\rho}^2$) versus sample sizes for KRR,  sketched KCGM with $m = \lceil n^{1/3}\rceil$,  plain Nystrom KCGM with $m = \lceil n^{2/3} \rceil$. }\label{fig:2}
\end{figure}

We present some numerical results to illustrate our derived results in the setting of learning with kernel methods.   
In all the simulations, we constructed  training datas $\{(x_i,y_i)\}_{i=1}^{n} \subseteq \mR \times \mR$ from the regression model $y = \FR(x) + \xi$, where  the regression function $\FR(x) = |x - 1/2| - 1/2$, the input $x$ is uniformly drawn from $[0, 1]$, and $\xi$
is a Gaussian noise with zero mean and standard
deviation $1$. By construction,  the function $\FR$
belongs to the first-order Sobolev space with $\|\FR\|_{\HK} = 1$.
In all the simulations,  the RKHS is associated with a Sobolev kernel $K(x,x') = 1 + \min(x,x')$. 
As noted in \cite[Example 3]{yang2015randomized} for Sobolev kernel, according to \cite{gu2013smoothing},   Assumption \ref{as:eigenvalues} is satisfied with $\gamma = {1\over 2}.$
As suggested by our theory, we set the projection dimension $m = \lceil n^
{1/3}\rceil,$ for KCGM with ROS sketches based on the fast
Hadamard transform while $m = \lceil n^{2/3}\rceil$ for KCGM with plain Nystr\"{o}m sketches. We performed simulations for $n$ in the set $\{32, 64, 128, 256, 512, 1024\}$ so as to study scaling with the sample size.
For each $n$, we performed 100 trials and both squared prediction errors and training errors averaged over these 100 trials were computed.
The errors for $n=1024$ versus the iterations were reported in Figure \ref{fig:1}. 
For each $n,$ the minimal squared prediction error over the first $m$ iterations is computed and these errors versus the sample size
were reported in Figure \ref{fig:2} in order to compare with state-of-the-art algorithm, kernel ridge regression (KRR).  From Figure \ref{fig:1}, we see that the squared prediction errors decrease at the first $3$ iterations and then they increase for both sketched and plain Nystr\"{o}m KCGM. 
This indicates that  the number of iterations has a regularization effect, and we can use a cross-validation approach to choose the best number of iterations in practice.  We also see that after some number of iterations,  the prediction error of sketched KCGM increases faster, comparing with  Nystr\"{o}m KCGM. We believe that the reason for this is that Nystr\"{o}m KCGM has a larger projection dimension in this example. 
Our theory predicts that the
squared prediction loss should tend to zero at the same rate $n^{-2/3}$
 as that of KRR.  Figure \ref{fig:2} confirms this theoretical prediction.

\section{Proof}\label{sec:proof}
In this section and the appendix, we provide all the proofs.   
\rvo{In Subsection \ref{subsect:proof}, we prove the representation theorem in 
Examples \ref{exa:nonSke}-\ref{exa:nysSke}.  Subsections \ref{subsect:proof}-\ref{subsec:der} are devoted to the proof of the main theorem and corollaries stated in Section \ref{sec:main}. The proof of the main results borrows ideas from the literature, e.g., \cite{smale2007learning,caponnetto2007optimal,blanchard2010optimal,rudi2015less,lin2020convergences}. }

\rvo{
 We here provide a proof sketch of the main theorem. As in \cite{blanchard2010optimal,lin2017optimal,lin2018optimal}, we first introduce a reference function $\midFun$ in \eqref{eq:midFun} and we have 
 \bea
\|\LK^{-a}(\IK \cgmt - \FH) \|_{\rho}  \leq  \|\LK^{-a} \IK (\cgmt - \midFun)\|_{\rho} +   \|\LK^{-a} (\IK\midFun - \FH)\|_{\rho} . 
 \eea
As the solution of our algorithm belongs to the subspace $\HKS$, following \cite{rudi2015less,lin2020convergences},
  we decompose $\midFun = (I -\proj)\midFun  + \proj\midFun$ and get 
   \bea
\|\LK^{-a}(\IK \cgmt - \FH) \|_{\rho}  \leq   \|\LK^{-a} \IK (\cgmt - \proj \midFun)\|_{\rho} + \|\LK^{-a} \IK (I - \proj) \midFun)\|_{\rho}  +    \|\LK^{-a} (\IK\midFun - \FH)\|_{\rho} . 
  \eea
   The estimations for the last two terms (which are related to the projection error and the bias, respectively) are standard and can be found in, e.g., \cite{blanchard2010optimal,lin2018optimal,lin2020convergences}. 
 We thus reach \eqref{eq:interm}.  We then  estimate the remaining term, $ \|\LK^{-a} \IK (\cgmt - \proj \midFun)\|_{\rho} $.
}\rvo{
In this step, we can not follow the strategies in \cite{lin2020convergences} for the linear regularized methods as  we consider a non-linear regularized scheme, i.e., $\cgmt = p_t(\ao) \proj \SX^* \NOutputs $   where $p_t$ a polynomial of degree $\leq t-1$  depending on $\NOutputs$ in a nonlinear fashion. 
Rather, we follow a similar argument as in \cite{hanke2017conjugate,blanchard2010optimal}. 
However, the considered normalized equation in this paper is different with \cite{blanchard2010optimal} and thus we need 
to introduce a different system of orthogonal polynomials as well as some basic estimations from \cite{lin2020convergences}. 
We thus  derive the estimates \eqref{eq:step1}/\eqref{eq:step1b}. Observe that there are several random quantities $\Delta_{\{1-5\}}$ and $|p'_t(0)|$  in \eqref{eq:step1}/\eqref{eq:step1b}. The estimations of the random quantities $\Delta_{\{1-5\}}$ are standard, which could be found in \cite{caponnetto2007optimal,smale2007learning,blanchard2010optimal,rudi2015less,lin2018optimalconve,lin2020convergences}.
The novelty of our proof compared with \cite{blanchard2010optimal} for classic KCGM (i.e., $\proj=I$) is that we use Lemma \ref{lem:operDifRes} to estimate the random quantity $\Delta_1$, which results from the concentration inequalities for sums of random self-adjoint  operators (rather than the  Hilbert-Schmidt operators). This allows us to get optimal
capacity-dependent rates for classic KCGM for the non-attainable case in the well-specific regime  without requiring additional unlabeled data.
The rest proofs are similar as that in \cite{blanchard2010optimal}, with simple modifications using basic estimations from \cite{lin2020convergences}.
}

\subsection{Proof for Subsection \ref{subsec:kcgm}}\label{subsect:proof}
Let $\Ss$ be a compact operator from the Euclidean space $(\mR^m, \|\cdot\|_{2})$ to $\HK$ such that   $\HKS = \overline{range(\Ss)}$. 
It is easy to see that $\Ss^* \Ss \in \mR^{m \times m}$. Let $t = \mbox{rank} ({\bf R})$ and
$\bR \in \mR^{m \times t}$ be the matrix such that $\bR \bR^{*} = (\Ss^* \Ss)^{\dag }$.  Here, $\mathbf{M}^\dag$ denotes the pseudo inverse of a matrix $\mathbf{M}.$ 
 As $\proj$ is the projection operator onto $\HKS,$ then
\be\label{eq:proj}
\proj = \Ss(\Ss^* \Ss)^{\dag} \Ss^* = \Ss \bR \bR^* \Ss^* .
\ee
For any polynomial function $q,$ we have that
\begin{align*}
q(\ao) \proj \SX^*\bby= 
q(\proj \TX \proj) \proj \SX^*\bby = q(\proj \SX^* \SX \proj) \proj \SX^*\bby.
\end{align*}
Noting that $\SX \proj = (\proj \SX^*)^*$, and using Lemma \ref{lem:sss} from {the coming subsection},  
\begin{align*}
q(\ao) \proj \SX^*\bby  = &  \proj \SX^*  q( \SX \proj  \proj \SX^*) \bby =   \proj \SX^*  q(\SX  \proj \SX^*) \bby.
\end{align*}
Introducing with \eqref{eq:proj}, 
\begin{align}\label{eq:interm2}	
q(\ao) \proj \SX^*\bby =& \Ss \bR \bR^* \Ss^* \SX^* q\left(\SX \Ss \bR \bR^* \Ss^*\SX^* \right) \bby.
\end{align}
Noting that $\bR^* \Ss^* \SX^* =  (\SX \Ss \bR)^*$, and applying Lemma \ref{lem:sss},
\begin{align}\label{eq:interm3}
q(\ao) \proj \SX^*\bby  = &  \Ss \bR q( \bR^* \Ss^* \SX^*\SX \Ss \bR ) \bR^* \Ss^*\SX^* \bby  = \Ss \bR  q(\tbK) \bb,
\end{align}
where we denote $$\bb = \bR^* \Ss^*\SX^* \bby,\quad \mbox{and} \quad \tbK =  \bR^* \Ss^* \SX^*\SX \Ss \bR.$$
Using $\bR \bR^{*} = (\Ss^* \Ss)^{\dag} $, which implies 
$\bR \bR^{*} (\Ss^* \Ss)  \bR \bR^{*} = \bR \bR^{*}$
and for any $g \in \HK, $
$$
\|\Ss \bR \bR^* \Ss^* g\|^2_{\HK} =   \la \Ss \bR \bR^* \Ss^* \Ss \bR \bR^* \Ss^* g, g
\ra_{\HK} =   \la \Ss \bR \bR^* \Ss^* g, g
\ra_{\HK} = \| \bR^* \Ss^* g \|_{2}^2,
$$
we get from \eqref{eq:interm2}  that
\begin{align}	\label{eq:interm4}
\| q(\ao) \proj \SX^*\bby\|_{\HK}  = \| \bR^* \Ss^* \SX^* q\left(\SX \Ss \bR \bR^* \Ss^*\SX^* \right) \bby \|_{\HK} = \|  q(\tbK) \bb\|_{2},
\end{align}
where we used Lemma \ref{lem:sss} for the last equality.

Note that  the solution of \eqref{eq:alg} is given by $\cgmt = p_t(\ao) \proj \SX^* \NOutputs$, with 
$$
p_t = \argmin_{p \in \mcP_{t-1}} \| (\ao p(\ao)  -I) \proj \SX^* \NOutputs \|_\HK.
$$
Using \eqref{eq:interm3} and \eqref{eq:interm4}, we know that 
$\cgmt = \Ss \bR p_t(\tbK) \bb$, with 
$$
p_t = \argmin_{p \in \mcP_{t-1}} \| (\tbK p(\tbK)  -\bI) \bb \|_2,
$$
which is equivalent to 
$\cgmt = \Ss \bR \ba_t  $, with 
$$
\ba_t = \argmin_{\ba \in \mcK_t(\tbK, \bb)} \| \tbK \ba  -  \bb \|_2.
$$
\begin{proof}[Proof for Example \ref{exa:nonSke}]
	For the ordinary non-sketching regimes, $\HKS = \HK$ and $\proj = I.$
	Denote $\bK = \SX\SX^*.$ Then
	$$
	\cgmt = \argmin_{\omega \in \mcK_t(\TX, \SX^*\NOutputs) } \|\TX \omega - \SX^* \NOutputs\|_\HK,
	$$
	is equivalent to $\cgmt = p_t(\TX) \SX^* \bby = p_t(\SX^* \SX) \SX^* \bby = \SX^* p_t(\bK) \bby = \SX^*\hat{\ba}_t,$ with $\hat{\ba}_t$ given by
	$$
	\hat{\ba}_t = \argmin_{\ba \in \mcK_t(\bK, \NOutputs) } \|\bK \ba - \NOutputs\|_{\bK}.
	$$
	Indeed,  $$\|\TX \omega - \SX^* \NOutputs\|^2_\HK = \|\SX^*(\SX \omega - \NOutputs)\|^2_\HK =  \|\SX \omega - \NOutputs\|_{\bK}^2,$$
	and for any polynomial function $p$, 
	$\SX p(\TX)\SX^* \NOutputs = \SX p(\SX^* \SX)\SX^* \NOutputs = \bK p(\bK)\NOutputs.$
\end{proof}

\begin{proof} [Proof for Example \ref{exa:ranSke}]
	For general randomized sketches, $\Ss = \SX^*\skt^*$. In this case,  $\Ss^* \Ss = \skt \SX \SX^* \skt^* = \skt \bK_{\bx \bx} \skt^*,$
	$$\tbK =  \bR^* \skt \SX \SX^*  \SX \SX^* \skt^* \bR = \bR^* \skt \bK_{ \bx \bx}^2 \skt^* \bR, $$
	$\bb = \bR^* \skt \SX \SX^* \bby = \bR^* \skt \bK_{\bx\bx} \bby,$
	and $\cgmt = \SX^*\skt^* \bR\hat{\ba}_t$.
\end{proof}

\begin{proof} [Proof for Example \ref{exa:nysSke}]
	In Nystr\"{o}m subsampling,  $\tilde{\bx}$ is a subset of size $m<n$
	drawn randomly following a distribution from $\bx$,  $\Ss = \SXS^*$,  and
	$\Ss^* \Ss = \bK_{\tilde{\bx}\tilde{\bx}}.$ In this case,
	$\tbK =  \bR^* \bK_{\tilde{\bx} \bx}\bK_{ \bx \tilde{\bx}} \bR,$ $\bb = \bR^* \bK_{\tilde{\bx} \bx} \bby, $
	and	  $\cgmt = \SXS^* \bR \ba_t$.
\end{proof}

In the \rvo{rest of the subsections}, we present the proofs for Section \ref{sec:main}.

\subsection{Operator Inequalities}\label{subsec:oper}
We first introduce some well-known operator inequalities.
 \begin{lemma}(\cite[Cordes inequality]{fujii1993norm})
	\label{lem:operProd}
	Let $A$ and $B$ be two positive bounded linear operators on a separable Hilbert space. Then
	\bea
	\|A^s B^s\| \leq \|AB\|^s, \quad\mbox{when } 0\leq s\leq 1.
	\eea
\end{lemma}

\begin{lemma}\label{lem:sss}
	Let $H_1,H_2$ be two separable Hilbert spaces and  $\mcS: H_1 \to H_2$ a compact operator. Then for any well-defined piecewise continuous function $f$ over $[0,\|\mcS\|]$,
	$$
	f(\mcS\mcS^*)\mcS = \mcS f(\mcS^*\mcS).
	$$
\end{lemma}
\begin{proof}
	This well known result can be proved using the singular value decomposition of a compact operator, see \cite[(2.43)]{engl1996regularization}.
\end{proof}

\begin{lemma}\label{lem:operDiff}
	Let $A$ and $B$ be two non-negative bounded linear operators on a separable Hilbert space with $\max(\|A\|,\|B\|) \leq \kappa^2$ for some non-negative $\kappa^2.$
	Then for any $\zeta>0,$
	\be
	\|A^\zeta - B^\zeta\| \leq C_{\zeta,\kappa} \|A -B\|^{\zeta \wedge 1},
	\ee
	where
	\be
	C_{\zeta,\kappa} = \begin{cases}
		1   & \mbox{when } \zeta \leq 1,\\
		2	\zeta \kappa^{2\zeta - 2} &\mbox{when } \zeta >1.
	\end{cases}
	\ee
\end{lemma}
\begin{proof}
	This is a well known result and its proof is based on the fact that $u^{\zeta}$ is operator monotone  if $0<\zeta\leq 1$. For $\zeta \geq 1$, we refer to \cite{dicker2017kernel}, or \cite{blanchard2016optimal} for the proof. {\color{red} }
\end{proof}

\begin{lemma}\label{lem:hansen_ineq}
	Let $X$ and $A$ be bounded linear operators on a separable Hilbert space $\HK$. Suppose that
	$A \succeq  0$ and $\|X\| \leq 1$. Then for any $s\in[0,1]$  and any $\lambda \geq 0,$
	\be \label{eq:3prod}
	X^* (A+\lambda I )^s X \preceq (X^* A X+\lambda X^* X)^s   \preceq (X^* A X + \lambda I )^s.
	\ee
	As a result, for any $\lambda \geq 0$ and any $\omega \in \HK$,
	\be \label{eq:3prodRes}
	\|(A+ \lambda I )^{s \over 2} X \omega\|_{\HK} \leq 	\|(X^* A X +  \lambda X^* X)^{s \over 2} \omega\|_\HK \leq  \| (X^*A X + \lambda I )^{s\over 2} \omega\|_\HK,
	\ee
	and for any bounded linear operator $F$ on $\HK,$
	\be\label{eq:3prodResB}
	\| F X^* (A + \lambda I)^{s \over 2} \| \leq \| F (X^* A X + \lambda I )^{s \over 2} \|.
	\ee
\end{lemma}
See Appendix \ref{appen:hansen_ineq} for the proof of the above lemma.

\begin{lemma}[\cite{lin2018optimal_ske}]\label{lem:proje2te}
	Let $\proj$ be a projection operator in a Hilbert space $\HK$, and $A$, $B$ be two semidefinite positive  operators on $\HK.$
	For any $0 \leq s,t \leq {1\over 2}$, we have 
	$$
	\|A^{s} (I - \proj) A^{t}\| \leq \|A - B\|^{s+t} + \|B^{1\over 2}(I - \proj)B^{1\over 2}\|^{s+t}.
	$$
\end{lemma}

\subsection{Orthogonal Polynomials and Some Notations}\label{subsec:ortho}
In this subsection, we review some basic properties of orthogonal polynomials, see, e.g. \cite{hanke2017conjugate,blanchard2010optimal}.

We denote by $(\eig , \eigv)_i$ an eigenvalue-eigenvector orthogonal basis for the operator $\ao$.
It is easy to see that $\eig \in [0 , \kappa^2 ]$, as $\ao$ is semi-definite and 
$\| \ao \| \leq \|\TX\| \leq \kappa^2$ by \eqref{eq:TXbound}.
For any $u\geq 0$, we denote  $F_{u}$ the orthogonal projection in $\HK$ onto the subspace $\{ \eigv: \eig < u \}$   and let $F_{u}^{\bot} = I - F_{u}.$

Denote $\mN_0 = \mN \cup \{0\}.$
For any $ t \in \mN_0$,
denote with  $\mcP_t$ the set of polynomials of degree at most $t$
and $\mcP_t^0 $ the set of polynomials in $\mcP_t$ having constant term equal to $1$.
For any $ t \in \mN_0$ and functions $\psi, \phi:  \mR \to \mR,$ define $$[\psi,\phi]_{(r)} = \la \psi(\ao)\proj \SX^*\bby, \ao^r \phi(\ao) \proj \SX^* \bby  \ra_{\HK}.$$ 
Denote $p_t^{(r)}$ the minimizer for 
$$
\argmin_{p  \in \mcP_t^0 } [p, p]_{(r-1)},
$$
and let $q_{t}^{(r)} \in \mcP_{t-1}$ be such that
 $ p_t^{(r)}(u) = 1 - u q_t^{(r)}(u).$
 We write $p_t$ and $q_t$ to mean $p_t^{(1)}$ and $q_t^{(1)}$, respectively.
According to the definition from Algorithm \ref{alg:1}, we know that $\omega_i = q_{i}(\ao) \proj \SX^* \bby$, $p_i(u) = 1 - u q_{i}(u)$. In the case $i=0,$ we set $q_{0} = 0$ and $p_0 = 1.$ 

Let $r\in \mN_0$. Observe that for any function $\phi,$ 
$$
[\phi,\phi]_{(r)} =  \sum_{i} \phi(\eig)^2 \eig^r \la \proj \SX^* \bby, \eigv \ra_{\HK}^2.
$$
Define $m_0$ the number of distinct positive eigenvalues of  $\ao$ such that $\proj\SX^* \bby$ has nonzero projection on the corresponding eigenspace.  Using that $\ao \eigv = 0$ implies 
$\SX  \proj \eigv  = 0$ as $\ao  = ( \SX \proj)^* \SX \proj$, we can prove that the measure defining $[\cdot, \cdot]_{(r)}^{1\over 2} $ has finite support of cardinality $m_0.$
Using the fact that a polynomial of degree  $t$  has at most  $t$  roots except $t=0$, it is easy to show that
$[\cdot, \cdot]_{(r)}$ with $r \in \mN_0$ is an inner product on the space $\mcP_{m_0-1}$. Furthermore,
there exists some $ p_{m_0} \in \mcP_{m_0}^{(0)}$ such that
$[p_{m_0}, p_{m_0}]_{(r-1)} = 0,$ and $p_{m_0}$ has $m_0$ distinct roots belonging to $(0,\kappa^2]$.

Based on \cite[Proposition 2.1]{hanke2017conjugate} , or using a similar argument based on the projection theorem as that in \cite{blanchard2010optimal}, $\{p_i^{(r)}\}_{i=1}^{m_0}$ are orthogonal with respect to $[\cdot, \cdot]_{(r)}$. Thus the polynomial $p_t^{(r)}$ with $t < m_0$ has exactly $t$ distinct roots belonging to $(0,\kappa^2]$, denoted by 
$(x_{k,t}^{(r)})_{1 \leq k \leq t}$ in increasing order.
For notational simplicity, we write $x_{k,t}$ to mean $x_{k,t}^{(1)}.$

The following lemma summarizes some basic facts about the orthogonal polynomials, see \cite{hanke2017conjugate}.
\begin{lemma}\label{lem:poly}
	Let $r \in \mN$ and
	$t$ be any integer satisfying $1\leq  t < m_0$.  Then  the following results hold.\\
	1) $
	  x_{1,t}^{(r)} <     x_{1,t}^{(r+1)}
	$
	 \\
	 2) For $u\in [0,x_{1,t}^{(r)}],$ $0\leq p_t^{(r)}(u) \leq 1$, $0 \leq q_t^{(r)}(u) u \leq 1$ 
	  and $q_t^{(r)}(u) \leq |(p_t^{(r)})'(0)|$. \\
	 3) 	$|(p_{t}^{(r)})'(0)|^{-1} \leq x_{1,t}^{(r)}.$ \\
	 4)  $
	 |p'_t(0)| \leq |p'_{t-1}(0)| + {[p_{t-1}, p_{t-1}]_{(0)} \over [p_{t-1}^{(2)}, p_{t-1}^{(2)}]_{(1)}}.
	 $
	\end{lemma}
\begin{proof}
 1) See \cite[Corollary 2.7]{hanke2017conjugate}.  \\
2) As $p_t^{(r)} \in \mcP_{t}^0$, $p_t^{(r)}(0) = 1$. Thus,  $p_t^{(r)}$ is convex and decreasing on  $[0,x_{1,t}^{(r)}]$. Therefore, $0\leq p_t^{(r)}(u) \leq 1$. Moreover, 
$0 \leq q_t^{(r)}(u) u = 1 - p_t^{(r)}(u) \leq 1$ and
$$q_t^{(r)}(u) = {1 -  p_t^{(r)}(u) \over u} = {p_t^{(r)}(0) -  p_t^{(r)}(u) \over -(0 - u)} \leq - (p_t^{(r)})'(0) =  | (p_t^{(r)})'(0)|.$$
3)  Rewriting  $p_t^{(r)}(u)$ as $\prod_{j=1}^{t}(1 - u/x_{j,t}^{(r)})$, and taking the derivative on $0$, we get $$|(p_t^{(r)})'(0) | =\Big| -\sum_{j=1}^t(x_{j,t}^{(r)})^{-1}\Big| \geq   (x_{1,t}^{(r)})^{-1},$$
which leads to the desired result. \\
4) Following from  \cite[Corollary 2.6]{hanke2017conjugate}, $(p_t^{(r)})'(0) \leq 0$ in the proof for Part 2), and that $[p_{t-1}^{(1)}, p_{t-1}^{(1)}]_{(0)} \geq [p_{t}^{(1)}, p_{t}^{(1)}]_{(0)}$ since $p_{t}^{(1)}$ is the minimizer of $[\cdot,\cdot]_{(0)}$ over $\mcP_{t}^0,$
 one can get the result.
	\end{proof}





\subsection{Deterministic Analysis}\label{subsec:determ}
In the proof, we introduce an intermediate function $\midFun \in \HK$, defined as follows, 
\be\label{eq:midFun}
\midFun = 
	\GL(\TK) \IK^* \FH,
\ee
where 
$$\GL(u) = \begin{cases}
u^{-1}, & \mbox{ if } u \geq \lambda, \\ 
 0,    &  \mbox{ if } u < \lambda.
\end{cases}
$$
For notational simplicity, we use the following notations: for $\lambda>0,$
\bea
\TXL= \TX + \lambda I,\quad \TKL = \TK + \lambda I, \quad \mbox{and} \quad  \rvo{\aol =  \ao + \lambda I}.
\eea
\begin{lemma}\label{lemma:midFun}
	Under Assumption \ref{as:regularity}, let $\midFun$ be given by \eqref{eq:midFun} for some $\lambda >0$. Then we have \\
	1) For any $a \leq \zeta,$ 	\be\label{eq:trueBias}
	\|\LK^{-a}(\IK \midFun - \FH)\|_{\rho} \leq
		R  \lambda^{\zeta-a}.
	\ee
	2) 
	\be\label{eq:popSeqNorm}
	\|\TK^{a-1/2} \midFun \|_{\HK} \leq R\cdot  \begin{cases}
		\RegPar^{\zeta+a -1}, & \text{if } -\zeta \leq a\leq 1-\zeta, \\
		\kappa^{2(\zeta +a-1)},& \text{if } \ a\geq 1 -\zeta.
	\end{cases}
	\ee
\end{lemma}
The proof for the above lemma  can be found in \cite{lin2018optimalconve}.

We next introduce some useful notations.
$$
\DZF   : = 1\vee \|\TXL^{-{1\over 2}}\TKL^{1\over 2}\|^2 \vee \|\TXL^{1\over 2} \TKL^{-{1\over 2}}\|^2 ,
$$
$$
\DZS : = \|\TKL^{-{1\over 2}}(\TX\midFun - \SX^*\NOutputs) \|_{\HK},
$$
$$
\DZT : = \|\TX - \TK\|_{HS} ,
$$
$$
\Delta_4 : = \| \TKL^{-{1\over 2}} (\TK - \TX)\|,
$$
$$
\DZI :=  \|\TK^{1\over 2}(I - \proj) \|^2 =  \|\TK^{1\over 2}(I - \proj) \TK^{1\over 2}\|,
$$
Here, $\|\cdot\|_{HS}$ is the Hilbert-Schmidt norm of a operator.
We also need the following preliminary lemmas. 

\begin{lemma}\label{lem:samProjErr}
	Under Assumption \ref{as:regularity} and the notations of Lemma \ref{lemma:midFun}, we have
	\be \label{eq:samProjErr}
	\| \TXL^{-{1\over 2}} ( \SX^* \bby  - \TX \proj \midFun)\|_{\HK}  \leq  
		\DZF^{1\over 2}  \DZS + \DZF^{1\over 2} R \times
		\begin{cases}
			(\DZI + \lambda) \lambda^{\zeta-1}, & \mbox{ if } \zeta \leq 1, \\ 
			 \left( \kappa\DZN  + \DZI   \right) \kappa^{2(\zeta-1)}, & \mbox{ if } \zeta >1.
		\end{cases}
	\ee
\end{lemma}
The proof for the above lemma can be found in \cite{lin2020convergences}. We provide a proof in Appendix \ref{subapp:48} for completeness.

\begin{lemma}\label{lem:APmidfun}
	Let $A: \HK \to \HK$ be a bounded operator.  Under Assumption \ref{as:regularity}, 
\be
\| A \proj  \midFun \|_{\HK} 
\leq  \begin{cases} R \| A \aol^{ {1\over 2}} \|  \DZF^{ {1 \over 2}}  \lambda^{\zeta - 1}, & \mbox{ if } \zeta \leq 1, \\
 R(  \| A \| C_{\zeta-{1\over 2},\kappa}  \DZT^{(\zeta - {1\over 2}) \wedge 1}  + \|A \ao^{1\over 2}\|  C_{\zeta-1,\kappa}(\DZT + \DZI)^{(\zeta-1) \wedge 1} +  \| A  \ao^{\zeta - {1\over 2}} \|),   & \mbox{ if } \zeta > 1.
\end{cases}
\ee	
\end{lemma}
\begin{proof}
	Part of this proof can be found in \cite{lin2020convergences}.
	
	If $0< \zeta \leq 1,$ by a simple calculation, and applying  Part 2) of Lemma \ref{lemma:midFun},
	\begin{align}
	\| A \proj  \midFun \|_{\HK} 
	\leq & \| A \proj \TXL^{ {1\over 2}} \|  \| \TXL^{ - {1\over 2} } \TKL^{{1\over 2}} \| \| \TK^{ - {1 \over 2} } \midFun \|_{\HK} \nonumber \\
	\leq  &  \| A \proj \TXL^{ {1\over 2}} \|  \DZF^{1\over 2}  \| \TK^{ - {1 \over 2} } \midFun \|_{\HK} \nonumber \\
	\leq  &  \| A \proj \TXL^{ {1\over 2}} \|  \DZF^{ {1 \over 2}}  R \lambda^{\zeta - 1}  \nonumber.
	\end{align} 
	Using \eqref{eq:3prodResB} from Lemma \ref{lem:hansen_ineq}, we get 
		\begin{align*}
	\| A \proj  \midFun \|_{\HK} 
	\leq  &  \| A (\proj \TX \proj + \lambda I )^{ {1\over 2}} \|  \DZF^{ {1 \over 2}}  R \lambda^{\zeta - 1} ,
	\end{align*} 
	which leads to the desired result. 
	
	If {\it  $\zeta \geq 1,$} applying Part 2) of Lemma \ref{lemma:midFun},
	\begin{align*}
	\| A \proj   \midFun \|_{\HK} 
	\leq   \| A \proj  \TK^{\zeta - {1\over 2}} \| \|\TK^{{1\over 2} - \zeta} \midFun \| \leq \| A \proj  \TK^{\zeta - {1\over 2}} \| R.
	\end{align*} 
	Adding and subtracting with the same term and using the triangle inequality,
	\begin{align*}
	\| A \proj   \midFun \|_{\HK} 
	\leq & R (\| A \proj  (\TK^{\zeta - {1\over 2}} - \TX^{\zeta-{1\over 2}})\|  + \| A \proj   \TX^{\zeta-{1\over 2}} \|) \\
	\leq & R (\| A \proj \|  \| \TK^{\zeta - {1\over 2}} - \TX^{\zeta-{1\over 2}}\|  + \| A \proj   \TX^{\zeta-{1\over 2}} \|).
	\end{align*} 
	Applying Lemma \ref{lem:operDiff} with \eqref{eq:TKBound} and \eqref{eq:TXbound}, we get
	\begin{align}
	\| A \proj   \midFun \|_{\HK} 
	\leq & R (\| A \proj \|   C_{\zeta-{1\over 2},\kappa} \DZT^{(\zeta - {1\over 2}) \wedge 1}  + \| A \proj   \TX^{\zeta-{1\over 2}} \|) \nonumber \\
	\leq & R (  \| A \proj \| C_{\zeta-{1\over 2},\kappa}  \DZT^{(\zeta - {1\over 2}) \wedge 1}  + \| A \proj   \TX^{\zeta-{1\over 2}} \|) . \label{eq:int4}
	\end{align} 
	With $$\aoB = \TX^{1\over 2} \proj \TX^{1\over 2} = (\proj \TX^{1\over 2})^* \proj \TX^{1\over 2} 
	$$ and Lemma \ref{lem:sss}, we can rewrite $\proj   \TX^{\zeta-{1\over 2}}$ as 
	$$
	\proj \TX^{1\over 2}  (\TX^{\zeta-1} - \aoB^{\zeta - 1} ) + \proj \TX^{1\over 2} \aoB^{\zeta - 1} =  \proj \TX^{1\over 2}  (\TX^{\zeta-1} - \aoB^{\zeta - 1} ) + \ao^{\zeta - 1} \proj \TX^{1\over 2} .
	$$
	Thus, combining with the triangle inequality, we get 
	\begin{align*}
	\| A \proj   \TX^{\zeta-{1\over 2}} \|  \leq &
	\| A \proj \TX^{1\over 2}  (\TX^{\zeta-1} - \aoB^{\zeta - 1} ) \| +  \| A \ao^{\zeta - 1} \proj \TX^{1\over 2} \| \\
	\leq &
	\| A \proj \TX^{1\over 2} \|  \|  \TX^{\zeta-1} - \aoB^{\zeta - 1}  \| +  \|A \ao^{\zeta - 1} \proj \TX^{1\over 2} \|.
	\end{align*} 
	Applying Lemma \ref{lem:operDiff} with $\| \aoB \| \leq \|\TX\| \leq \kappa^2,$ 
	\begin{align*}
	\| A \proj   \TX^{\zeta-{1\over 2}} \| 
	\leq 
	\| A \proj \TX^{1\over 2}\|  C_{\zeta-1,\kappa} \|\TX - \aoB\|^{(\zeta-1) \wedge 1} +  \| A \ao^{\zeta - 1} \proj \TX^{1\over 2} \|.
	\end{align*} 
	Using Lemma \ref{lem:proje2te}, $(I -\st)^2 = I -\st$ and $\|A^* A\|=\|A\|^2$, we have 
	$$
	\|\TX - \aoB\| = \|\TX^{1\over 2}(I - \st) \TX^{1\over2}\| \leq\|\TX - \TK\| + \|\TK^{1\over 2}(I - \st) \TK^{{1\over 2}}\| \leq  \DZT + \DZI,
	$$
	and we thus get
	\begin{align}\label{eq:int3}
	\| A \proj   \TX^{\zeta-{1\over 2}} \| 
	\leq 
	\| A \proj \TX^{1\over 2}\|  C_{\zeta-1,\kappa}(\DZT + \DZI)^{(\zeta-1) \wedge 1} +  \| A  \ao^{\zeta - 1} \proj \TX^{1\over 2} \|.
	\end{align} 
	Applying \eqref{eq:3prodResB} of Lemma \ref{lem:hansen_ineq}, we get 
	$\| A \proj \TX^{1\over 2}\| \leq \| A (\proj \TX \proj)^{1\over 2}\| = \|A \ao^{1\over 2}\| $ and $\| A  \ao^{\zeta - 1} \proj \TX^{1\over 2} \| \leq  \| A  \ao^{\zeta - {1\over 2}} \|.$ Thus, 
		\begin{align*}
	\| A \proj   \TX^{\zeta-{1\over 2}} \| 
	\leq 
	\|A \ao^{1\over 2}\|  C_{\zeta-1,\kappa}(\DZT + \DZI)^{(\zeta-1) \wedge 1} +  \| A  \ao^{\zeta - {1\over 2}} \| .
	\end{align*} 
	Introducing the above into \eqref{eq:int4}, one can get
	\begin{align*}
\| A \proj   \midFun \|_{\HK} 
\leq  R (  \| A \proj \| C_{\zeta-{1\over 2},\kappa}  \DZT^{(\zeta - {1\over 2}) \wedge 1}  + \|A \ao^{1\over 2}\|  C_{\zeta-1,\kappa}(\DZT + \DZI)^{(\zeta-1) \wedge 1} +  \| A  \ao^{\zeta - {1\over 2}} \|),
\end{align*} 
which leads to the desired result by noting that $\| A \proj \| \leq \|A\|.$
	\end{proof}

With the above lemmas, we can prove the following result for estimating $\|\LK^{-a}(\IK \cgmt - \FH) \|_{\rho} .$

\begin{lemma}\label{lem:step1}
	Under Assumption \ref{as:regularity},
	let $u \in (0,x_{1,t}]$ and $0 \leq a \leq (\zeta \wedge {1\over 2}).$ Then the following statements hold. \\
1) If $\zeta \leq 1,$
\begin{align}
\|\LK^{-a}(\IK \cgmt - \FH) \|_{\rho} 
\leq  
\DZF^{{1} - a} \left(|p_t' (0)|^a + \lambda^{1-a} |p_t' (0)| +  {\left( u + \lambda\right)^{1-a} \over u} \right)  (\DZS +  (\DZI/\lambda + 1) R \lambda^{\zeta}) \nonumber \\
 +   \DZF^{{1\over 2} - a} \left(  {\left( u + \lambda\right)^{{1\over 2} - a} \over u}  {\| \ao \cgmt - \proj \SX^* \bby\|_{\HK}}  +  R  \DZF^{{1 \over 2}}  ( u + \lambda)^{1 - a} \lambda^{\zeta - 1} \right) +  \left( (\DZI/\lambda)^{1-a} + 1 \right) R \lambda^{\zeta - a}.  \label{eq:step1}
\end{align}	 
2) If $\zeta \geq 1,$
	\begin{align}
\|\LK^{-a}(\IK \cgmt - \FH) \|_{\rho} 
\leq &
\DZF^{1 - a} \big(|p_t' (0)|^a + \lambda^{1-a} |p_t' (0)| +  {\left( u + \lambda\right)^{1-a} \over u} \big)  \left(\DZS +  R \kappa^{2(\zeta-1)} \left( \kappa\DZN  + \DZI   \right) \right) \nonumber \\
& +   \DZF^{{1\over 2} - a}  R C_{\zeta-{1\over 2},\kappa} \left(     \DZT^{(\zeta - {1\over 2}) \wedge 1}  +  (\DZT+\DZI)^{(\zeta-1) \wedge 1} u^{ {1\over 2}}  +  u^{\zeta - {1\over 2}} \right)  (u + \lambda)^{{1\over 2} - a}  \nonumber \\
&  +   \DZF^{{1\over 2} - a}    {\| \ao \cgmt - \proj \SX^* \bby\|_{\HK}}   {(u + \lambda)^{{1\over 2} -a } \over u} + R \left( \kappa^{2(\zeta-1)} \DZI^{1-a} +  \lambda^{\zeta - a}\right). \label{eq:step1b}
\end{align}
\end{lemma}
\begin{proof}
	The proof \rvo{borrows} ideas from \cite{blanchard2010optimal} and \cite{lin2020convergences}.
	
	Adding and subtracting with the same term, and	then using the triangle inequality, 
	\begin{align*}
		\|\LK^{-a}(\IK \cgmt - \FH) \|_{\rho} 
		\leq& \|\LK^{-a} \IK (\cgmt - \midFun)\|_{\rho} +   \|\LK^{-a} (\IK\midFun - \FH)\|_{\rho} \leq 
		\|\LK^{-a} \IK (\cgmt - \midFun)\|_{\rho} +   R \lambda^{\zeta - a},
	\end{align*}
	where we used Part 1) of Lemma \ref{lemma:midFun} for the last inequality.
	Using 
	$$
	\LK^{-a} \IK = \LK^{-{1\over 2}} ( \IK \IK^*)^{{1\over 2} -a} \IK = \LK^{-{1\over 2}} \IK ( \IK^* \IK)^{{1\over 2} -a} =   \LK^{-{1\over 2}} \IK \TK^{{1\over 2} -a}
	$$
	and \eqref{eq:rho2hk},
	\begin{align}\label{eq:rhoHK}
		\|\LK^{-a}(\IK \cgmt - \FH) \|_{\rho} 
		\leq \| \LK^{-{1\over 2}} \IK \TK^{{1\over 2}-a} (\cgmt - \midFun)\|_{\rho} + R \lambda^{\zeta - a}	\leq   \|\TK^{{1\over 2}-a} (\cgmt - \midFun)\|_{\HK} + R \lambda^{\zeta - a}.
	\end{align}
	Subtracting and adding with the same term, then using the triangle inequality, 
	\begin{align*}
		\|\LK^{-a}(\IK \cgmt - \FH) \|_{\rho} 
		\leq& \|\TK^{{1\over 2}-a} (\cgmt - \proj \midFun)\|_{\HK} + \|\TK^{{1\over 2}-a} (I - \proj) \midFun \|_{\HK} + 
		R \lambda^{\zeta - a}.
	\end{align*}
	Since $\proj$ is a projection operator, $(I - \proj)^s = I - \proj$ for any $s>0$, and we thus can get
	\begin{align*}
		& \|\LK^{-a}(\IK \cgmt - \FH) \|_{\rho} \\
		\leq& \|\TK^{{1\over 2}-a} (\cgmt - \proj \midFun)\|_{\HK} + \|\TK^{{1\over 2}-a} (I - \proj)^{1 - 2a} \|  \|(I - \proj) \TK^{1\over 2}\| \| \TK^{-{1\over 2}}\midFun\|_{\HK} +  \Bias.
	\end{align*}
	Using Lemma \ref{lem:operProd} and Part 2) of Lemma \ref{lemma:midFun},  we  get  \cite{lin2018optimal_ske},
	\begin{align}\label{eq:interm}
		\|\LK^{-a}(\IK \cgmt - \FH) \|_{\rho} 
		\leq& \|\TK^{{1\over 2}-a} (\cgmt - \proj \midFun)\|_{\HK} + \DZI^{1 - a} \ProjErr + 
		\Bias.
	\end{align}
In what follows, we estimate $\|\TK^{{1\over 2}-a} (\cgmt - \proj \midFun)\|_{\HK} .$
\\
{\bf Estimating $ \|\TK^{{1\over 2}-a} (\cgmt - \proj \midFun)\|_{\HK} .$}	
	We first have
	$$
	\|\TK^{{1\over 2}-a} (\cgmt - \proj \midFun)\|_{\HK} \leq \| \TK^{{1\over 2}-a} \TKL^{a - {1\over 2}}\|
	\|\TKL^{{1\over 2}-a} \TXL^{a - {1\over 2}}\| \| \TXL^{{1\over 2} -  a}  (\cgmt - \proj \midFun)\|_{\HK}.
	$$
	Obviously, $\| \TK^{{1\over 2}-a} \TKL^{a - {1\over 2}}\| \leq 1$ and by Lemma \ref{lem:operProd},  $\|\TKL^{{1\over 2}-a} \TXL^{a - {1\over 2}}\| \leq \|\TKL^{{1\over 2}} \TXL^{ - {1\over 2}}\|^{1 - 2a}  \leq \DZF^{{1\over 2} - a}.$ Thus, 
	$$
	\|\TK^{{1\over 2}-a} (\cgmt - \proj \midFun)\|_{\HK} \leq  \DZF^{{1\over 2} - a} \| \TXL^{{1\over 2} -  a}  (\cgmt - \proj \midFun)\|_{\HK} =   \DZF^{{1\over 2} - a} \| \TXL^{{1\over 2} -  a} \proj (\cgmt - \proj \midFun)\|_{\HK},
	$$
	where the last equality follows from the facts that $\cgmt \in  \HKS$ and that $\proj$ is the projection operator with range $\HKS$ which implies $\proj^2 = \proj$ and $\cgmt = \proj \cgmt$.
Noting that $\|\proj\| \leq 1$, using \eqref{eq:3prodRes}, we get
	$$
	\|\TK^{{1\over 2}-a} (\cgmt - \proj \midFun)\|_{\HK} \leq  \DZF^{{1\over 2} - a} \|  \aol^{{1\over 2} -  a}  (\cgmt - \proj \midFun)\|_{\HK} .
	$$
	Adding and subtracting with the same term, using the triangle inequality,  and noting that $\cgmt = \proj \cgmt,$
	\begin{align*}
		\|\TK^{{1\over 2}-a} (\cgmt - \proj \midFun)\|_{\HK} \leq &
		\DZF^{{1\over 2} - a} \left(\| \fproj \aol^{{1\over 2} -  a} (\cgmt - \proj \midFun)\|_{\HK} +  \| \fproj^{\bot} \aol^{{1\over 2} -  a} \proj (\cgmt -  \midFun)\|_{\HK} \right).
	\end{align*}
	Introducing with $\cgmt = q_t(\ao) \proj \SX^* \bby,$ 
	\begin{align}\label{eq:3}
		\|\TK^{{1\over 2}-a} (\cgmt - \proj \midFun)\|_{\HK} 
		\leq 
		\DZF^{{1\over 2} - a} \left(\| \fproj \aol^{{1\over 2} -  a}  (q_t(\ao) \proj \SX^* \bby  - \proj \midFun)\|_{\HK} +  \| \fproj^{\bot} \aol^{{1\over 2} -  a} \proj (\cgmt - \midFun)\|_{\HK} \right).
	\end{align}
In what follows, we estimate the last two terms from the above. \\
	{\it Estimating $ \| \fproj^{\bot} \aol^{{1\over 2} -  a} \proj (\cgmt - \midFun)\|_{\HK}$.}
	By a direct calculation, following from the definition of $\ao$ given by \eqref{eq:ao} and $\proj^2 = \proj,$
\begin{align*}
\| \fproj^{\bot} \aol^{{1\over 2} -  a} \proj (\cgmt - \midFun)\|_{\HK} \leq & 
\| \fproj^{\bot} \aol^{1 - a} \ao^{-1}\|   \| \fproj^{\bot}  \aol^{-{1 \over 2}}  \ao \proj (\cgmt - \midFun)\|_{\HK}  \\
\leq &    {\left( u + \lambda\right)^{1-a} \over u}  \| \fproj^{\bot}  \aol^{-{1 \over 2}} (  \ao \cgmt -  \proj \TX \proj \midFun)\|_{\HK}.
	\end{align*}
Adding and subtracting with the same term, and using the triangle inequality,
\begin{align*}
&\| \fproj^{\bot} \aol^{{1\over 2} -  a} \proj (\cgmt - \midFun)\|_{\HK}  \\
\leq & {\left( u + \lambda\right)^{1-a} \over u}  \left( \| \fproj^{\bot}  \aol^{-{1 \over 2}}(\ao \cgmt - \proj \SX^* \bby)\|_{\HK} + \| \fproj^{\bot}  \aol^{-{1 \over 2}}  \proj ( \SX^* \bby  - \TX \proj \midFun)\|_{\HK}  \right) \\
\leq & {\left( u + \lambda\right)^{1-a} \over u} \left( {\| \ao \cgmt - \proj \SX^* \bby\|_{\HK} \over (u + \lambda)^{1\over 2} } +  \|  \aol^{-{1 \over 2}}  \proj \TXL^{1\over 2} \|  \| \TXL^{-{1\over 2}} ( \SX^* \bby  - \TX \proj \midFun)\|_{\HK}  \right).
\end{align*}
Using \eqref{eq:3prodResB}, $\|  \aol^{-{1 \over 2}}  \proj \TXL^{1\over 2} \| \leq \|  \aol^{-{1 \over 2}}  (\proj \TX \proj + \lambda I)^{1\over 2} \| = 1,$ and thus
\begin{align}\label{eq:2}
\| \fproj^{\bot} \aol^{{1\over 2} -  a} \proj (\cgmt - \midFun)\|_{\HK} 
\leq  {\left( u + \lambda\right)^{1-a} \over u} \left( {\| \ao \cgmt - \proj \SX^* \bby\|_{\HK} \over (u + \lambda)^{1\over 2} } +  \| \TXL^{-{1\over 2}} ( \SX^* \bby  - \TX \proj \midFun)\|_{\HK}  \right).
\end{align}
	{\it Estimating  $\| \fproj \aol^{{1\over 2} -  a}  (q_t(\ao) \proj \SX^* \bby  - \midFun)\|_{\HK}$. } Adding and subtracting with the same term, noting that $\proj^2 = \proj $, and using the triangle inequality, we get
\begin{align}
& \| \fproj \aol^{{1\over 2} -  a}  (q_t(\ao) \proj \SX^* \bby  - \proj \midFun)\|_{\HK}  \nonumber\\
\leq & \| \fproj \aol^{{1\over 2} -  a}  q_t(\ao) \proj (\SX^* \bby - \TX \proj \midFun) \|_{\HK }  + \| \fproj \aol^{{1\over 2} -  a}  p_t(\ao) \proj  \midFun \|_{\HK} \nonumber \\
\leq &  \| \fproj \aol^{{1\over 2} -  a}  q_t(\ao) \proj \TXL^{1\over 2} \|   \| \TXL^{-{1\over 2}} ( \SX^* \bby  - \TX \proj \midFun)\|_{\HK}  +  \| \fproj \aol^{{1\over 2} -  a}  p_t(\ao) \proj  \midFun \|_{\HK} . \label{eq:1}
\end{align}
Using \eqref{eq:3prodResB}, 
\begin{align} 
 \| \fproj \aol^{{1\over 2} -  a}  q_t(\ao) \proj \TXL^{1\over 2} \| \leq & \|\fproj \aol^{{1\over 2} -  a}  q_t(\ao) (\proj \TX \proj + \lambda I)^{1\over 2}  \| = \| \fproj \aol^{1 -  a}  q_t(\ao) \| \nonumber \\
 \leq &  \max_{x \in [0, u]} | (x + \lambda)^{1 - a} q_t (x) | \leq \max_{x \in [0, u]} \big( |x q_t(x)|^{1-a} |q_t(x)|^a + \lambda^{1 - a} |q_t (x)| \big) \nonumber\\
 \leq & |p_t' (0)|^a + \lambda^{1-a} |p_t' (0)|, \label{eq: lem1}
\end{align}
where we used Part 2) of Lemma \ref{lem:poly} with $u \in [0,x_{1,t}]$ for the last inequality.  Introducing the above into \eqref{eq:1}, we get
\begin{align*}
 & \| \fproj \aol^{{1\over 2} -  a}  (q_t(\ao) \proj \SX^* \bby  - \proj \midFun)\|_{\HK}  \\
\leq &
(|p_t' (0)|^a + \lambda^{1-a} |p_t' (0)|)   \| \TXL^{-{1\over 2}} ( \SX^* \bby  - \TX \proj \midFun)\|_{\HK}  +  \| \fproj \aol^{{1\over 2} -  a}  p_t(\ao) \proj  \midFun \|_{\HK}.
\end{align*}
Introducing the above and \eqref{eq:2} into \eqref{eq:3}, we get 
	\begin{align}
\|\TK^{{1\over 2}-a} (\cgmt - \proj \midFun)\|_{\HK} 
\leq & 
\DZF^{{1 \over 2} - a} \left(|p_t' (0)|^a + \lambda^{1-a} |p_t' (0)| +  {\left( u + \lambda\right)^{1-a} \over u} \right)  \| \TXL^{-{1\over 2}} ( \SX^* \bby  - \TX \proj \midFun)\|_{\HK} \nonumber \\
& +   \DZF^{{1\over 2} - a} \left(  {\left( u + \lambda\right)^{{1\over 2} - a} \over u}  {\| \ao \cgmt - \proj \SX^* \bby\|_{\HK}}  +  \| \fproj \aol^{{1\over 2} -  a}  p_t(\ao) \proj  \midFun \|_{\HK}  \right). \label{eq:a2}
\end{align}
In what follows, we estimate $\| \fproj \aol^{{1\over 2} -  a}  p_t(\ao) \proj  \midFun \|_{\HK},$ considering two different cases.  \\
If $0< \zeta \leq 1,$ applying Lemma \ref{lem:APmidfun}, 
\begin{align*}
\| \fproj \aol^{{1\over 2} -  a}  p_t(\ao) \proj  \midFun \|_{\HK} 
\leq &    \| \fproj \aol^{{1\over 2} -  a}  p_t(\ao) \aol^{ {1\over 2}} \|  \DZF^{ {1 \over 2}}  R \lambda^{\zeta - 1}  \\
  \leq& \max_{x \in [0, u]} p_t(x) (x + \lambda)^{1 - a} \DZF^{ {1 \over 2}}  R \lambda^{\zeta - 1}  \\
   \leq & ( u + \lambda)^{1 - a} R \DZF^{ {1 \over 2}} \lambda^{\zeta - 1},
\end{align*} 
where we used Part 2) of Lemma \ref{lem:poly} for the last inequality.
Introducing the above and \eqref{eq:samProjErr} into \eqref{eq:a2}, and then \rvo{combining} with \eqref{eq:interm}, one can prove the desired result for $\zeta \leq 1.$\\
{\it If $\zeta \geq 1,$} applying Lemma \ref{lem:APmidfun} with $A = \fproj \aol^{{1\over 2} -  a}  p_t(\ao) ,$ we get
\begin{align}\label{eq:b1}
\| \fproj \aol^{{1\over 2} -  a}  p_t(\ao) \proj   \midFun \|_{\HK} 
\leq    R(  \| A \| C_{\zeta-{1\over 2},\kappa}  \DZT^{(\zeta - {1\over 2}) \wedge 1}  + \|A \ao^{1\over 2}\|  C_{\zeta-1,\kappa}(\DZT + \DZI)^{(\zeta-1) \wedge 1} +  \| A  \ao^{\zeta - {1\over 2}} \|) .
\end{align} 
For any $s \geq 0,$ using Part 2) of Lemma \ref{lem:poly},
$$
\| A \ao^s \| = \max_{x \in [0, u]} (x + \lambda)^{{1\over 2} - a}  p_t(x) x^s \leq   (u + \lambda)^{{1\over 2} - a}   u^s.
$$
Using the above with $s = 0, {1\over 2}, \zeta - {1\over 2}$ into \eqref{eq:b1}, we get 
\begin{align*}
 \| \fproj \aol^{{1\over 2} -  a}  p_t(\ao) \proj   \midFun \|_{\HK}  
\leq &  R (   C_{\zeta-{1\over 2},\kappa}  \DZT^{(\zeta - {1\over 2}) \wedge 1}  +  C_{\zeta-1,\kappa} (\DZT+\DZI)^{(\zeta-1) \wedge 1} u^{ {1\over 2}}  +  u^{\zeta - {1\over 2}} ) (u + \lambda)^{{1\over 2} - a}. 
\end{align*} 
Introducing the above and \eqref{eq:samProjErr} into \eqref{eq:a2}, and then combining with \eqref{eq:interm}, we can prove the desired result for $\zeta \geq 1$.
\end{proof}

From Lemma \ref{lem:step1}, we can see that in order to control the error, we need to estimate the random quantities $\DZF, \DZS,\DZT,\DZN,\DZI$,
$|p_t' (0)|$, and  $\| \ao \cgmt - \proj \SX^* \bby\|_{\HK}.$ 
The random quantities will be estimated in Subsections \ref{subsec:pro} and \ref{subsec:proj}, while $\| \ao \cgmt - \proj \SX^* \bby\|_{\HK}$ can be bounded due to the stopping rule. In order to estimate  $|p_t' (0)|$, we introduce the following two lemmas, from which we can estimate $|p_t' (0)|$ as shown in the coming proof for the main theorem.
\begin{lemma}\label{lemma:interm}
	 The following statements hold. \\
	1) If $\zeta \leq 1$,
	\begin{align*}
		&\| \ao \cgmt - \proj\SX^* \NOutputs\|_{\HK}  \\
		 \leq & \left(   |p_{t}' (0) |^{-{1 \over 2}} + \lambda^{1\over 2} \right)  \DZF^{1\over 2} (\DZS +  (\DZI/\lambda + 1) R \lambda^{\zeta})+  R \DZF^{{1\over 2}}  
		\left( c_{ {3\over 2}} |p_t'(0)|^{- {3\over 2}} \lambda^{\zeta - 1} + 2 \lambda^{\zeta - {1 \over 2}} |p_t'(0)|^{-1}
		\right) .
	\end{align*}
	2) If $\zeta>1,$
	\begin{align}
&	\| \ao \cgmt - \proj\SX^* \NOutputs\|_{\HK} 
	 \leq \left(   |p_{t}' (0) |^{-{1 \over 2}} + \lambda^{1\over 2} \right) \DZF^{1\over 2} (\DZS +  R 
	 \left( \kappa\DZN  + \DZI   \right) \kappa^{2(\zeta-1)}) \nonumber \\
	& \ + R(   2C_{\zeta-{1\over 2},\kappa}  \DZT^{(\zeta - {1\over 2}) \wedge 1}  |p_t'(0)|^{-1} + c_{3 \over 2} C_{\zeta-1,\kappa}(\DZT + \DZI)^{(\zeta-1) \wedge 1}  |p_t'(0)|^{-{3 \over 2}} +   c_{\zeta + {1\over 2}} |p_t'(0)|^{- {(\zeta + { 1 \over 2})}} ).
	\end{align}
		Here, we denote $0^0=1$ and
	$$c_v = (2v)^v,\quad v \geq0 .$$
\end{lemma}

\begin{proof}
		Let $$\phi_t(x) = p_t(x) \left({x_{1,t} \over x_{1,t} - x} \right)^{1\over 2}.$$
	Following from  \cite[(3.8)]{hanke2017conjugate},
		\begin{align*}
	\| \ao \cgmt - \proj\SX^* \NOutputs\|_{\HK} \leq   \|F_{x_{1,t}} \phi_{t}(\ao) \proj\SX^* \NOutputs\|_{\HK}.
	\end{align*}
	Using the triangle inequality, with a basic calculation, we get
	\begin{align}
		& \| \ao \cgmt - \proj\SX^* \NOutputs\|_{\HK} \nonumber \\
		& \ \leq \|F_{x_{1,t}} \phi_{t}(\ao) \proj(\SX^* \NOutputs - \TX \proj \midFun)\|_{\HK} + 
		\|F_{x_{1, t}} \phi_{t}(\ao) \ao \midFun\|_{\HK} \nonumber \\
		& \ \leq  \|F_{x_{1,t}} \phi_{t}(\ao) \proj \TXL^{1\over 2} \|  \| \TXL^{-{1\over 2}} (\SX^* \NOutputs - \TX \proj \midFun)\|_{\HK} + 
		\|F_{x_{1, t}} \phi_{t}(\ao) \ao \midFun\|_{\HK} \nonumber\\ 
		& \ \leq  \|F_{x_{1,t}} \phi_{t}(\ao) \aol^{1\over 2} \|  \| \TXL^{-{1\over 2}} (\SX^* \NOutputs - \TX \proj \midFun)\|_{\HK} + 
		\|F_{x_{1, t}} \phi_{t}(\ao) \ao \midFun\|_{\HK} ,\label{eq:4}
	\end{align}
	where we used \eqref{eq:3prodResB} of Lemma \ref{lem:hansen_ineq} for the last inequality. Note that
	\begin{align*}
		\|F_{x_{1, t}} \phi_{ t}(\ao) \aol^{1\over 2} \|  \leq \sup_{x \in [0,x_{1,t}]} |\phi_{ t}(x) (x + \lambda)^{1\over 2}| \leq  \sup_{x \in [0,x_{1, t}]} |(x^{1\over 2} + \lambda^{1\over 2}) \phi_{ t}(x)|.
	\end{align*}
	Following from \cite[(3.10)]{hanke2017conjugate},
		\be\label{eq:310}
	\sup_{x \in [0,x_{1,t}]} |\phi_t(x) x^{v}| \leq c_v|p'_t(0)|^{-v},\quad v\geq0.
	\ee
	Thus, we get that
	\begin{align*}
		\|F_{x_{1, t}} \phi_{t}(\ao) \aol^{1\over 2}\| \leq   |p_{t}' (0) |^{-{1 \over 2}} + \lambda^{1\over 2}. 
	\end{align*}
	Introducing the above into \eqref{eq:4},  we get that
	\begin{align}
	\| \ao \cgmt - \proj\SX^* \NOutputs\|_{\HK} 
	\ \leq \left(   |p_{t}' (0) |^{-{1 \over 2}} + \lambda^{1\over 2} \right)  \| \TXL^{-{1\over 2}} (\SX^* \NOutputs - \TX \proj \midFun)\|_{\HK} + 
	\|F_{x_{1, t}} \phi_{t}(\ao) \ao \midFun\|_{\HK}. \label{eq:b2}
	\end{align}
	Now, we  consider \rvo{two cases}.\\
	{\it Case I: $\zeta \leq 1$.} \\
	Using Lemma \ref{lem:APmidfun}, with $\ao = \ao \proj,$	
	$$
	\|F_{x_{1,t}} \phi_{t}(\ao) \ao \midFun\|_{\HK} \leq R \|  F_{x_{1, t}} \phi_{t}(\ao) \ao \aol^{ {1\over 2}}\|  \DZF^{1\over 2} \lambda^{\zeta - 1} \leq R \DZF^{{1\over 2}} \lambda^{\zeta - 1}   \max_{x \in [0, x_{1,t}]} |\phi_{t}(x) x (x + \lambda)^{ {1\over 2}}|.
	$$
	Applying \eqref{eq:310}, 
	$$
	\|F_{x_{1,t}} \phi_{t}(\ao) \ao \midFun\|_{\HK} \leq  R \DZF^{{1\over 2}}  
	\left( c_{ {3\over 2}} |p_t'(0)|^{- {3\over 2}} \lambda^{\zeta - 1} + 2 \lambda^{\zeta - {1 \over 2}} |p_t'(0)|^{-1}
	\right).
	$$
	Introducing the above and \eqref{eq:samProjErr} into \eqref{eq:b2}, one can get the desired result.  \\
	{\it Case II: $\zeta >1$.}\\ 
	Applying \eqref{eq:310},  we get that \rvo{for any} $s \geq 0,$
	$$
	\| F_{x_{1, t}} \phi_{t}(\ao) \ao \ao^{s} \|_{\HK} \leq \max_{x \in [0, x_{1,t}]} |\phi_t(x) x^{s+1}| \leq c_{s+1} |p_t'(0)|^{-(s+1)}.
	$$
	Using the above and Lemma \ref{lem:APmidfun}, with $\ao = \ao \proj,$  we get that
	\begin{align*}
&	\| F_{x_{1, t}} \phi_{t}(\ao) \ao \proj \midFun \|_{\HK} \\
	\leq&  R(  2 C_{\zeta-{1\over 2},\kappa}  \DZT^{(\zeta - {1\over 2}) \wedge 1}  |p_t'(0)|^{-1} + c_{3 \over 2} C_{\zeta-1,\kappa}(\DZT + \DZI)^{(\zeta-1) \wedge 1}  |p_t'(0)|^{-{3 \over 2}} +   c_{\zeta + {1\over 2}} |p_t'(0)|^{- {(\zeta + { 1 \over 2})}} ).
	\end{align*}
	Applying the above and \eqref{eq:samProjErr} into \eqref{eq:b2}, we get the desired result.
\end{proof}

\begin{lemma}\label{lem:intermB}
	Let $u \in (0,x_{1,t}].$ Then the following statements hold. \\
	1) If $\zeta \leq 1,$
	\begin{align}
	[p_{t}, p_{t}]_{(0)}^{1\over 2}  \leq 
	(u + \lambda)^{1\over 2}  \DZF^{1\over 2}  \left(  \DZS +  R (\DZI + \lambda) \lambda^{\zeta - 1}  + R u \lambda^{\zeta - 1} \right) +
	u^{-{1\over 2}} [p_{t}^{(2)}, p_{t}^{(2)}]_{(1)}^{1\over 2}.
	\end{align} 
	2) If $\zeta >1,$
	\begin{align}
& [p_{t}, p_{t}]_{(0)}^{1\over 2}  \leq 
(u + \lambda)^{1\over 2}  \DZF^{1\over 2}	 \left( \DZS +  R 
\left( \kappa\DZN  + \DZI   \right) \kappa^{2(\zeta-1)} \right) \nonumber \\ 
& \ + R(   C_{\zeta-{1\over 2},\kappa} u  \DZT^{(\zeta - {1\over 2}) \wedge 1}  +  u^{{3\over 2}}   C_{\zeta-1,\kappa}(\DZT + \DZI)^{(\zeta-1) \wedge 1} + u^{\zeta + {1\over 2}} ) +
u^{-{1\over 2}} [p_{t}^{(2)}, p_{t}^{(2)}]_{(1)}^{1\over 2}.
\end{align} 
\end{lemma}
\begin{proof}
	Since $p_t$ is the minimizer of $[p,p]_{(0)}$ over $\mcP_{t}^{0}$
	and $p_t^{(2)} \in \mcP_{t}^{0}$,
		\begin{align*}
	 [p_{t}, p_{t}]_{(0)}^{1\over 2} 
	 \  \leq  [p_{t}^{(2)}, p_{t}^{(2)}]_{(0)}^{1\over 2}  = \|p_{t}^{(2)}(\ao)\proj\SX^*\NOutputs\|_{\HK} 
	\end{align*}
	Using the triangle inequality, 
	\begin{align*}
	 [p_{t}, p_{t}]_{(0)}^{1\over 2}  
		\leq  &
		\|\fproj p_{t}^{(2)}(\ao) \proj\SX^* \NOutputs\|_{\HK} + \|\fproj^{\perp}p_{t}^{(2)}(\ao) \proj\SX^* \NOutputs\|_{\HK} \\
		 \leq &
		\|\fproj p_{t}^{(2)}(\ao) \proj(\SX^* \NOutputs - \TX \proj \midFun)\|_{\HK} + \TailFunB 
		 +  \|\fproj^{\perp}p_{t}^{(2)}(\ao)\proj \SX^* \NOutputs\|_{\HK} .
	\end{align*}
By a  basic calculation, 
\begin{align*}
[p_{t}, p_{t}]_{(0)}^{1\over 2} \leq & 
\|\fproj p_{t}^{(2)}(\ao) \proj \TXL^{1\over 2}\|  \| \TXL^{-{1\over 2}} (\SX^* \NOutputs - \TX \proj \midFun)\|_{\HK} + \TailFunB   \\
& + \|\fproj^{\perp}\ao^{-{1\over 2}}\| \| \ao^{1\over 2} p_{t}^{(2)}(\ao)\proj \SX^* \NOutputs\|_{\HK}  \\
\leq & 
\|\fproj p_{t}^{(2)}(\ao)  \aol^{1\over 2}\|   \| \TXL^{-{1\over 2}} (\SX^* \NOutputs - \TX \proj \midFun)\|_{\HK}  + \TailFunB +
u^{-{1\over 2}} [p_{t}^{(2)}, p_{t}^{(2)}]_{(1)}^{1\over 2},
\end{align*} 	
where we used \eqref{eq:3prodResB} of Lemma \ref{lem:hansen_ineq} for the last inequality.
	Using Part 2) of Lemma \ref{lem:poly}, we get
	\begin{align*}
	\|\fproj p_{t}^{(2)}(\ao)\aol^{1 \over 2}\| \leq \max_{x \in [0, u]} |p_{t}^{(2)} (x) (x + \lambda)^{1\over 2}|  \leq (u + \lambda)^{1\over 2} ,
	\end{align*}
	and thus 
	\begin{align}\label{eq:c1}
	[p_{t}, p_{t}]_{(0)}^{1\over 2}  \leq 
(u + \lambda)^{1\over 2}    \| \TXL^{-{1\over 2}} (\SX^* \NOutputs - \TX \proj \midFun)\|_{\HK}  + \TailFunB +
	u^{-{1\over 2}} [p_{t}^{(2)}, p_{t}^{(2)}]_{(1)}^{1\over 2}.
	\end{align} 	
{\it Case I: $\zeta \leq 1$.}\\
Using $\proj^2 = \proj$ and Lemma \ref{lem:APmidfun}, 
\begin{align*}
 \TailFunB  = &  \|\fproj p_{t}^{(2)}(\ao) \ao \proj \midFun \| \leq R \|\fproj p_{t}^{(2)}(\ao) \ao  \aol^{ {1\over 2}} \|  \DZF^{ {1 \over 2}}  \lambda^{\zeta - 1} \\
 \leq &   R \max_{x \in [0, u]} |p_{t}^{(2)}(x) x  (x+\lambda)^{ {1\over 2}} |  \DZF^{ {1 \over 2}}  \lambda^{\zeta - 1}.
\end{align*}
Using Part 2) of Lemma \ref{lem:poly}, 
\begin{align*}
\TailFunB  \leq   R u  (u + \lambda)^{ {1\over 2}}   \DZF^{ {1 \over 2}}  \lambda^{\zeta - 1} .
\end{align*}
Introducing the above and \eqref{eq:samProjErr} into  \eqref{eq:c1}, one can get the desired result for $\zeta \leq 1.$ \\
{\it Case II: $\zeta \geq 1$.}\\
Using Part 2) of Lemma \ref{lem:poly}, for any $s \geq 0,$
\begin{align*}
\|\fproj p_{t}^{(2)}(\ao) \ao \ao^{s}\| =  \max_{x \in [0, u]} |p_{t}^{(2)}(x) x^{s+1} |    \leq    u^{s +  1},
\end{align*}
Noting that as $\proj^2 = \proj $, $\TailFunB  =  \| \fproj p_{t}^{(2)}(\ao) \ao \proj \midFun \|$, and combining with Lemma \ref{lem:APmidfun}, we get
\begin{align*}
\TailFunB   \leq  R(   C_{\zeta-{1\over 2},\kappa} u  \DZT^{(\zeta - {1\over 2}) \wedge 1}  +  u^{{3\over 2}}   C_{\zeta-1,\kappa}(\DZT + \DZI)^{(\zeta-1) \wedge 1} + u^{\zeta + {1\over 2}} ). 
\end{align*}
Introducing the above and \eqref{eq:samProjErr} into  \eqref{eq:c1}, one can get the desired result for $\zeta \geq 1.$
\end{proof}

\subsection{Probabilistic Estimates}\label{subsec:pro}

In this subsection, we introduce some probabilistic estimates to bound the random quantities $\DZF$, $\DZS$,
$\DZT$,    and   $\DZN$


\begin{lemma}  \label{lem:operDifRes}
	Under Assumption \ref{as:eigenvalues},
	let $\delta\in(0,1)$, and $\lambda = n^{-\theta}$ with $\theta \in[0,1)$ or $\lambda = [1 \vee \log n^{\gamma}]/n$.  Then with probability at least $1-{\delta},$
	\bea
	\| (\TK+\lambda I)^{1/2}(\TX+\lambda I)^{-1/2}\|^2 \vee 	\| (\TK+\lambda I)^{-1/2}(\TX+\lambda I)^{1/2}\|^2 \leq 3 a(\delta),
	\eea
	where
	$a(\delta) = 8\kappa^2\log{ {4\kappa^2\mathrm{e}(c_{\gamma}+1) }\over \delta \|\TK\|}$
	if $\lambda = [1 \vee \log n^{\gamma}]/n$, or 
	$
	a(\delta)
	=   8\kappa^2\left(\log{ {4\kappa^2(c_{\gamma}+1) }\over \delta \|\TK\|} + 
	{ \theta \gamma \over \mathrm{e}(1-\theta)}\right)
	$ otherwise.
\end{lemma}
The proof of the above result for the case $\lambda = n^{-\theta}$ with $\theta \in [0,1)$ can be found in 
\cite{lin2018optimalconve}. Here, using essentially the same idea, we also provide a similar result considering the case $\lambda = [1\vee \log n^{\gamma}]/n.$ We report the proof in  Appendix \ref{subapp:operDiff}.


\begin{lemma}\label{lem:statEstiOper}
	Let $0<\delta<1/2.$ It holds with probability at least $1-\delta:$
	\bea
	\|\TK-\TX\| \leq \|\TK - \TX\|_{HS} \leq {2 \kappa^2 \log (2/\delta)  \over n} + \sqrt{2 \kappa^4 \log (2/\delta)  \over n} .
	\eea
	Here, $\|\cdot\|_{HS}$ denotes the Hilbert-Schmidt norm.
\end{lemma}
\begin{proof}
	Using Lemma 2 (which is a direct corollary of  the concentration inequality for Hilbert-space valued random variables from \cite{pinelis1986remarks}) from \cite{smale2007learning},  one can prove the desired result.
\end{proof}


\begin{lemma}\label{lem:DZS}
Under Assumptions \ref{as:noiseExp} and \ref{as:eigenvalues}, with probability at least $1 - \delta$, the following holds: 
\begin{align}
 & \|\TKL^{-{1\over 2}}(\TX\midFun - \SX^*\NOutputs) \|_{\HK} \nonumber \\
\leq & 2 \left( {4\kappa(M + \kappa^{1 \vee (2\zeta)} R \lambda^{(\zeta - {1\over 2})_-})  \over n \sqrt{\lambda}} +  \sqrt{ 8(3 R^2 \kappa^2 \lambda^{2\zeta - 1} + (3B^2 + 4 Q^2) c_{\gamma} \lambda^{-\gamma} ) \over n}\right)\log{2 \over \delta} + R \lambda^{\zeta}.
\end{align}
\end{lemma}
The above lemma is proved in \cite{lin2018optimalconve,lin2018optimal}. We provide a proof in Appendix \ref{subapp:variance}.

\begin{lemma}\label{lem:effDifOper}
	Under Assumption \ref{as:eigenvalues},	let $0<\delta<1/2.$ It holds with probability at least $1-\delta:$
	\bea
	\|\TKL^{-{1\over 2}} (\TK - \TX)\|_{HS} \leq 2\kappa \left( {2\kappa \over n\sqrt{\lambda}} + \sqrt{c_{\gamma} \over n \lambda^{\gamma}}
	\right)\log{2\over \delta}.
	\eea
\end{lemma}
The proof for the above lemma can be found in \cite{lin2020convergences}.

\subsection{Projection Errors}\label{subsec:proj}

In this subsection, we estimate projection errors $\|(I - \proj)\TK^{1\over 2}\|^2$, considering different projections.

The first lemma provides upper bounds on projection errors with plain Nystr\"{o}m subsampling.
\begin{lemma}\label{lem:DZI}
	Under Assumption \ref{as:eigenvalues}, let $\proj$ be the projection operator with range $$\HKS = \overline{span\{x_1,\cdots, x_m\}}.$$
	Then with probability at least $1 - \delta,$ ($\delta \in (0,1)$)
	\be
		\|(I - \proj)\TK^{1\over 2}\|^2 \leq \|(I - \proj)\TK_{\eta}^{1\over 2}\|^2  \leq  {1 \vee \log m^{\gamma} \over m} 24 \kappa^2\log{ {4\kappa^2\mathrm{e} (c_{\gamma}+1) }\over \delta \|\TK\|},
	\ee
	where $\eta = {1 \vee \log m^{\gamma} \over m}$.
\end{lemma}
\begin{rem}
	Lemma \ref{lem:DZI} holds for any $$\HKS = \overline{span\{x_1,\cdots, x_m\}}.$$
\end{rem}

The following lemma estimates projection errors with randomized sketches.

\begin{lemma} \label{lem:DZX}
	Under Assumption \ref{as:eigenvalues}, let 
	$\HKS =  \overline{range\{\SX^* \skt^\top\}},$ where  $\skt \in \mR^{m \times n}$ is a random matrix satisfying \eqref{eq:isoPro} and $\proj$ be the projection operator with its range $\HKS$. Then with probability at least $1 - 3\delta$ ($\delta \in (0,1/3)$), we have
\bea
\| (I - \proj) \TK^{1\over 2} \|^2  \leq {1 \over n^{\theta}} \left( 1 \vee {\log n^{\gamma} \over n^{1-\theta}} \right) 7a_{\gamma} \log {4 \over \delta},
\eea
provided that	
	\be\label{eq:skeDim}
	m \geq  \bar{C} n^{\theta \gamma}\log^{\beta} n (1 \vee \log n^{\gamma})^c \log^{3}{4 \over \delta}  , \quad 
	c=
	\begin{cases}
	0, & \mbox{if } \theta <1, \\
	-\gamma, & \mbox{if } \theta =1.
	\end{cases}
	\ee
Here, $a_{\gamma} = 24\kappa^2\log{ {\kappa^2\mathrm{e}^2(c_{\gamma}+1) }\over  \|\TK\|},$ and 
$\bar{C} = 100 c_0' \left( 1+ 10 b_{\gamma} \right) $ with
$$
b_{\gamma} = 24\kappa^2 (4\kappa^2 + 2\kappa\sqrt{c_{\gamma}} + c_{\gamma}) \left(\log {2\kappa^2(c_{\gamma} + 1) \over \|\TK\|}+ 1  + \tilde{c}\right),\quad \tilde{c} = \begin{cases}
{\theta\gamma \over \mathrm{e}(1-\theta)},&  \mbox{if } \theta<1, \\
1, & \mbox{if } \theta =1.
\end{cases}
$$
\end{lemma}

Finally, the next lemma upper bounds projection errors with ALS Nystr\"{o}m  subsampling sketches. 
The ALS Nystr\"{o}m subsampling is defined as follows. 

\paragraph{Approximated Leveraging Scores (ALS) Nystr\"{o}m Subsampling}
In this regime, $\HKS =  \overline{range\{\SX^* \skt^\top\}},$ where each row ${1 \over \sqrt{m}}\ba_j^{\top}$ of $\skt$  is i.i.d. drawn according to
$$
\mP \left(\ba = {1\over \sqrt{q_i}} \mathbf{e}_i \right) = q_i, 
$$
where $q_i > 0$ will be chosen later and $\{ \mathbf{e}_i : i \in[n] \}$ is the standard basis of $\mR^n$.
For every $i \in [n]$ and $\lambda>0,$ the leveraging scores of
$\bK(\bK + \lambda I)$ is the sequence $\{l_i(\lambda)\}_{i=1}^n$ with
$$l_i (\lambda) =  \left( \bK(\bK + \lambda I)^{-1}\right)_{ii}, \quad  \forall i \in [n].$$
In practice,  the leveraging scores of
$\bK(\bK + \lambda I)$ is hard to compute, and we can only compute its approximation $\hat{l}_i (\lambda)$ such that
$$
{1\over L} {l}_{i} (\lambda) \leq \hat{l}_{i}(\lambda) \leq L {l}_{i} (\lambda),
$$ 
for some $L \geq 1$. In the ALS Nystr\"{o}m subsampling, we set 
$$q_i: = q_{i}(\lambda)  =  {\hat{l}_{i}(\lambda)  \over \sum_j \hat{l}_{j}(\lambda) }.$$

\begin{lemma} \label{lem:DZXALS}
	Under Assumption \ref{as:eigenvalues}, let 
	$\HKS =  \overline{range\{\SX^* \skt^\top\}},$ where  $\skt \in \mR^{m \times n}$ is a randomized matrix related to ALS Nystr\"{o}m subsampling, and $\proj$ be the projection operator with its range $\HKS$. Then with probability at least $1 - 3\delta$ ($\delta \in (0,1/3)$), we have
	\bea
	\| (I - \proj) \TK^{1\over 2} \|^2  \leq {1 \over n^{\theta}} \left( 1 \vee {\log n^{\gamma} \over n^{1-\theta}} \right) 4a_{\gamma} \log {4 \over \delta},
	\eea
	provided that	
	\be\label{eq:skeDimALS}
	m \geq  \bar{C}_1 n^{\theta \gamma}(1 \vee \log n^{\gamma})^c \log^{3}{4 \over \delta}  , \quad 
	c=
	\begin{cases}
		1, & \mbox{if } \theta <1, \\
		1-\gamma, & \mbox{if } \theta =1.
	\end{cases}
	\ee
	Here, $\bar{C}_1 = 8 b_{\gamma} L^2 (4 + \log (2b_{\gamma}))$ where $a_{\gamma}$ and $b_{\gamma}$ are given by Lemma \ref{lem:DZX}.
	Here, $\lambda = n^{-\theta}$ if $\theta \in [0,1)$, or $\lambda = {1\vee \log n^{\gamma} \over n}$ if $\theta = 1$.
\end{lemma}

The proofs for the above lemmas can be found in \cite{lin2020convergences}. We provide the proofs in Appendix \ref{subapp:PNy}, \ref{subapp:ske} and \ref{subapp:als}.

\subsection{Deriving Main Results}\label{subsec:der}
We are ready to prove the main theorem and its corollaries.

\begin{proof}[Proof of Theorem \ref{thm:gen}]	
Applying Lemmas \ref{lem:operDifRes}, \ref{lem:statEstiOper}, \ref{lem:DZS} \ref{lem:effDifOper}, and Condition \eqref{eq:projCond}, and noting that $\lambda \in [n^{-1},1],$
 we get that with probability at least $1 - 5 \delta$, the following inequalities hold: 
$$
\DZF \leq C_1 \log {2 \over \delta},
$$
$$
\DZS \leq  C_2 \lambda^{\zeta}  \log {2 \over  \delta},
$$
$$
\DZT \leq \kappa^2(\sqrt{2}+ 2) {1 \over \sqrt{n}} \log {2 \over \delta} \leq  C_3 \lambda^{ (2\zeta + \gamma) \vee 1 \over 2} \log {2 \over \delta}, \quad C_3 = \kappa^2(\sqrt{2}+ 2),
$$
$$
\DZN \leq 2\kappa \left( {2\kappa \over n\sqrt{\lambda}} + \sqrt{c_{\gamma} \over n \lambda^{\gamma}}
\right)\log{2\over \delta} \leq C_4 \lambda^{\zeta} \log {2 \over \delta}, \quad C_4 = 2\kappa(2\kappa + \sqrt{c_{\gamma}}),
$$
$$
\DZI \leq  C_1'  \lambda^{1\vee\zeta - a \over 1 - a} \log {2 \over \delta}, 
$$
where 
$$C_1  = 
\begin{cases}
24 \kappa^2 \left(\log{ {2\kappa^2 \mathrm{e} (c_{\gamma}+1) }\over \|\TK\|} + { \gamma \over 2\zeta + \gamma - 1} \right), & \mbox{if }   2\zeta+\gamma >1,
\\
24 \kappa^2 \log{ {2\kappa^2 \mathrm{e} (c_{\gamma}+1) }\over \|\TK\|} & \mbox{otherwise},
\end{cases}
$$
$$
C_2 = 
2 \left( {4 \kappa (M + \kappa^{1\vee(2\zeta)} R )} +  \sqrt{ 8(3 R^2 \kappa^2  + (3B^2 + 4 Q^2) c_{\gamma}  )}\right) + R.
$$
In what follows, we assume the above estimates hold and we prove the results considering two different cases.

{\it \bf Case I: $ \zeta \leq 1$.} 
We first have
\be\label{eq:biasInt1}
\DZF^{1\over 2} (\DZS +  (\DZI/\lambda + 1) R \lambda^{\zeta}) \leq C_5 \lambda^{\zeta} \log^{3\over 2}{2\over \delta}, 
\ee
where we denote
$$
C_5 =  C_1^{1\over 2} (C_2 + (C_1' + 1 )R ).
$$
By  the above inequality and Lemma \ref{lemma:interm}, we have
\begin{align}
&\| \ao \cgmt - \proj\SX^* \NOutputs\|_{\HK}  \nonumber\\
\leq & \left(   |p_{t}' (0) |^{-{1 \over 2}} + \lambda^{1\over 2} \right)  C_5 \log^{3\over 2} {2 \over \delta} \lambda^{\zeta}  +  R C_1^{{1\over 2}} \log^{1\over 2} {2 \over \delta} 
\left( c_{ {3\over 2}} |p_t'(0)|^{- {3\over 2}} \lambda^{\zeta - 1} + 2 \lambda^{\zeta - {1 \over 2}} |p_t'(0)|^{-1}
\right).\label{eq:c2}
\end{align}
{\it Step 1: }
Now set the stopping rule as 
$$
\| \ao \cgmt - \proj\SX^* \NOutputs\|_{\HK} \leq (\tau + C_5) \log^{3 \over 2} {2 \over  \delta} \lambda^{\zeta+{1\over 2}},
$$
where $\tau>0$ is given later.
From the definition of $\hat{t},$ we have 
\be \label{eq:c4}
\| \ao \omega_{\hat{t} - 1} - \proj\SX^* \NOutputs\|_{\HK} >  (\tau +  C_5) \log^{3 \over 2} {2 \over  \delta} \lambda^{\zeta+{1\over 2}} .
\ee
Combining with \eqref{eq:c2}, noting that $\log {2 \over \delta} \geq 1$, by a simple calculation,
\begin{align*}
&\tau  \log^{3\over 2} {2 \over  \delta}  \lambda^{\zeta + {1\over 2}} \\
\leq &   C_5 |p_{{\hat{t} - 1} }' (0) |^{-{1 \over 2}}  \lambda^{\zeta}  \log^{3\over 2} {2 \over \delta}  +  R C_1^{{1\over 2}}  \log^{1\over 2} {2 \over \delta} 
\left( c_{ {3\over 2}} |p_{\hat{t} - 1}'(0)|^{- {3\over 2}} \lambda^{\zeta - 1} + 2\lambda^{\zeta - {1 \over 2}} |p_{\hat{t} - 1}'(0)|^{-1}
\right)  \\ 
\leq& 3  \log^{3\over 2} {2 \over  \delta} \lambda^{\zeta} \max\left(   C_5   |p_{\hat{t} - 1}' (0) |^{-{1 \over 2}}  ,  R C_1^{{1\over 2}} c_{ {3\over 2}}   |p_{\hat{t} - 1}'(0)|^{- {3\over 2}} \lambda^{ - 1}  ,   2R C_1^{{1\over 2}}    \lambda^{ - {1 \over 2}} |p_{\hat{t} - 1}'(0)|^{-1} 
\right).
\end{align*}
If the maximum is achieved at the first term of the right-hand side from the above, then 
$$
\tau  \log^{3\over 2} {2 \over  \delta}  \lambda^{\zeta + {1\over 2}} \leq 
3   C_5 \log^{3\over 2} {2 \over  \delta}  |p_{\hat{t} - 1}' (0) |^{-{1 \over 2}}  \lambda^{\zeta},
$$
and by a direct calculation, 
$$
|p_{\hat{t} - 1}' (0) | \leq (3  C_5 / \tau)^2 \lambda^{-1}.
$$
If the maximum is achieved  at the second term or the third term, using a similar argument, one can show that at least one of the following two inequalities holds, 
$$
|p_{\hat{t} - 1}' (0) |  \leq (3 R C_1^{1\over 2} c_{3 \over 2} / \tau)^{2 \over 3} \lambda^{-1},
$$
$$
|p_{\hat{t} - 1}' (0) |  \leq (6 C_1^{1\over 2} R /\tau ) \lambda^{-1}.
$$
Now we choose $\tau$ as 
$$
\tau \geq \max \left( 3\sqrt{2}   C_5, 6 \sqrt{2} R C_1^{1\over 2}  c_{3 \over 2}  ,12 C_1^{1\over 2} R  \right). 
$$
Then, following from the above analysis,
\be\label{eq:c3}
|p_{\hat{t}-1}'(0)| \leq {1 \over 2} \lambda^{-1}.
\ee
{\it Step 2:} 
In this step, we choose $u = \lambda$.
Using \eqref{eq:c3} and Part 3) of Lemma \ref{lem:poly},  it is easy to show that 
$$
u \leq |p_{\hat{t} - 1}'(0)|^{-1} \leq x_{1,\hat{t} - 1}. 
$$
Applying Lemma \ref{lem:intermB}, with \eqref{eq:biasInt1},
\begin{align*}
[p_{\hat{t}-1}, p_{\hat{t} - 1}]_{(0)}^{1\over 2}  \leq 
\sqrt{2} \left( C_5  + C_1^{1\over 2}R\right)  \lambda^{\zeta + {1\over 2}}  \log^{3\over 2}{2 \over \delta }   +
u^{-{1\over 2}} [p_{\hat{t} - 1}^{(2)}, p_{\hat{t} - 1}^{(2)}]_{(1)}^{1\over 2}.
\end{align*} 
Combining with \eqref{eq:c4}, 
\begin{align*}
[p_{\hat{t}-1}, p_{\hat{t} - 1}]_{(0)}^{1\over 2}  \leq& 
{\sqrt{2} \left( C_5  + \sqrt{C_1} R\right)  \over C_5 + \tau}  \| \ao \omega_{\hat{t} - 1} - \proj\SX^* \NOutputs\|_{\HK}   +
u^{-{1\over 2}} [p_{\hat{t} - 1}^{(2)}, p_{\hat{t} - 1}^{(2)}]_{(1)}^{1\over 2}  \\
 =& 
{\sqrt{2} \left( C_5  + \sqrt{C_1}R\right)  \over C_5 + \tau}   [p_{\hat{t}-1}, p_{\hat{t} - 1}]_{(0)}^{1\over 2}  +
u^{-{1\over 2}} [p_{\hat{t} - 1}^{(2)}, p_{\hat{t} - 1}^{(2)}]_{(1)}^{1\over 2}  \\
\leq  & 
{1\over 2} [p_{\hat{t}-1}, p_{\hat{t} - 1}]_{(0)}^{1\over 2}  +
u^{-{1\over 2}} [p_{\hat{t} - 1}^{(2)}, p_{\hat{t} - 1}^{(2)}]_{(1)}^{1\over 2},
\end{align*} 
provided that
$$
\tau \geq 2\sqrt{2} (C_5 + \sqrt{C_1} R)  - C_5.
$$
Thus, we get
\begin{align}\label{eq:d1}
{[p_{\hat{t}-1}, p_{\hat{t}-1}]_{(0)}^{1\over 2}  \over  [p_{\hat{t}-1}^{(2)}, p_{\hat{t}-1}^{(2)}]_{(1)}^{1\over 2} }
\leq  2  u^{-{1\over 2}} ,
\end{align}
Combining with Part 4) of Lemma \ref{lem:poly} and \eqref{eq:c3}, we get that 
\be \label{eq:c5}
|p_{\hat{t}}'(0)| \leq  |p_{\hat{t}-1}'(0)| + 4 u^{-1} \leq 5 \lambda^{-1}.
\ee
{\it Step 3.} In this step, we let $u = {1 \over 5} \lambda.$ Then following from \eqref{eq:c5} and Part 3) of Lemma \ref{lem:poly}, we have 
\be\label{eq:d2}
u \leq |p_{\hat{t}}'(0)|^{-1} \leq x_{1,\hat{t}}.
\ee
Using Lemma \ref{lem:step1}, and introducing with \eqref{eq:c5}, \eqref{eq:biasInt1} and the above estimates,
we have 
\begin{align*}
\|\LK^{-a}(\IK \omega_{\hat{t} } - \FH) \|_{\rho} 
\leq  
 C_6 \lambda^{\zeta - a} 
 \log^{2 - a} {2 \over \delta}
+   6C_1^{{1\over 2} - a}    \lambda^{-a - {1\over 2}} 
 {\| \ao \omega_{\hat{t}} - \proj \SX^* \bby\|_{\HK}} \log^{{1\over 2} - a}{2 \over \delta} . 
\end{align*}	
where
$$C_6 =C_1^{{1\over 2}}(5^a  + 11) C_5 + C_1^{1-a}R (6/5)^{1-a} + ((C_1')^{1-a} + 1)R.$$ 
From the definition of the stopping rule,  we get
\begin{align*}
\|\LK^{-a}(\IK \omega_{\hat{t} } - \FH) \|_{\rho} 
\leq  
\left( C_6 + 
  6C_1^{{1\over 2} - a}  (\tau + C_5)  \right) \lambda^{\zeta - a} \log^{2 - a}{2 \over \delta},
\end{align*}	 
which leads to the desired result for $\zeta \leq 1$. 

{\it \bf Case II: $\zeta>1$.} \\
{\it Step 1.}
 Introducing the estimates given in the beginning of the proof and using $\lambda^{\zeta - a \over 1-a} \leq \lambda^{\zeta}$ as $\lambda\leq 1$,
\be\label{eq:biasInt2}
\DZF^{1\over 2}	 \left( \DZS +  R 
\left( \kappa\DZN  + \DZI   \right) \kappa^{2(\zeta-1)} \right) \leq C_7   \lambda^{\zeta}  \log^{3 \over 2}{ 2 \over \delta},  
\ee
where $$C_7 = C_1^{1\over 2} (C_2 + R(\kappa C_4 + C_1') \kappa^{2(\zeta-1)}).$$
Using Lemma \ref{lemma:interm},
	\begin{align}
&	\| \ao \cgmt - \proj\SX^* \NOutputs\|_{\HK} 
\leq \left(  |p_{t}' (0) |^{-{1 \over 2}} + \lambda^{1\over 2} \right) C_7   \lambda^{\zeta} \log^{3 \over 2}{ 2 \over \delta} \nonumber \\
& \ + R(   2C_{\zeta-{1\over 2},\kappa}  \DZT^{(\zeta - {1\over 2}) \wedge 1}  |p_t'(0)|^{-1} + c_{3 \over 2} C_{\zeta-1,\kappa}(\DZT + \DZI)^{(\zeta-1) \wedge 1}  |p_t'(0)|^{-{3 \over 2}} +   c_{\zeta + {1\over 2}} |p_t'(0)|^{- {\zeta + { 1 \over 2}}} ). \nonumber
\end{align}
Notice that by a direct calculation, with $\zeta>1$, $\lambda<1$, $\kappa^2 \geq 1$ and $\log {2 \over \delta} \geq 1,$
\begin{align}
\DZT^{(\zeta - {1\over 2}) \wedge 1}  \leq \left( C_3 \lambda^{\zeta+ \gamma/2} \log {2 \over \delta}  \right)^{(\zeta - {1\over 2}) \wedge 1}  \leq  C_3  \lambda^{\zeta - {1\over 2}} \log {2 \over \delta} , \quad \mbox{ and}   \label{eq:delta3p}
\end{align}
\begin{align}
(\DZT + \DZI)^{(\zeta-1) \wedge 1}  \leq & \left( C_3 \lambda^{\zeta+ \gamma/2} \log {2 \over \delta}  + \lambda^{\zeta - a \over 1 - a} C_1' \log {2 \over \delta} \right)^{(\zeta-1) \wedge 1} \nonumber\\
\leq &  \left(  C_3   +  C_1'  \right) \lambda^{\zeta - 1} \log {2 \over \delta}.\label{eq:delta35p}
\end{align}
Therefore, 
	\begin{align}\label{eq:inter2}
&	\| \ao \cgmt - \proj\SX^* \NOutputs\|_{\HK} 
\leq \left(  |p_{t}' (0) |^{-{1 \over 2}} + \lambda^{1\over 2} \right) C_7   \lambda^{\zeta}  \log^{3 \over 2}{ 2 \over \delta} \nonumber \\
& \ + R(   2 C_{\zeta-{1\over 2},\kappa}  C_3  \lambda^{\zeta - {1\over 2}}   |p_t'(0)|^{-1} + c_{3 \over 2} C_{\zeta-1,\kappa} \left(  C_3 +  C_1'  \right) \lambda^{\zeta - 1} |p_t'(0)|^{-{3 \over 2}} +   c_{\zeta + {1\over 2}} |p_t'(0)|^{- {\zeta + { 1 \over 2}}} )  \log {2 \over\delta}. 
\end{align}
Now set the stopping rule as 
$$
\| \ao \cgmt - \proj\SX^* \NOutputs\|_{\HK} \leq (\tau + C_7) \log^{3 \over 2} {2 \over  \delta} \lambda^{\zeta+{1\over 2}},
$$
where $\tau>0$ is given later. From the definition of $\hat{t},$ we have 
\be \label{eq:c4B}
\| \ao \omega_{\hat{t} - 1} - \proj\SX^* \NOutputs\|_{\HK} >  (\tau +  C_7) \log^{3 \over 2} {2 \over  \delta} \lambda^{\zeta+{1\over 2}}.
\ee
Letting $t = \hat{t} - 1$ in \eqref{eq:inter2} and combining with \eqref{eq:c4B}, by a direct calculation,
\begin{align*}
& \tau  \log^{3 \over 2} {2 \over  \delta}  \lambda^{\zeta + {1\over 2}} \leq   C_7 |p_{{\hat{t} - 1} }' (0) |^{-{1 \over 2}}  \lambda^{\zeta}  \log^{3\over 2} {2 \over \delta} 
\\ & \ \   + R(  2 C_{\zeta-{1\over 2},\kappa}  C_3  \lambda^{\zeta - {1\over 2}}   |p_{\hat{t} - 1}'(0)|^{-1} + c_{3 \over 2} C_{\zeta-1,\kappa} \left(  C_3 +  C_1'  \right) \lambda^{\zeta - 1} |p_{\hat{t} - 1}'(0)|^{-{3 \over 2}} +   c_{\zeta + {1\over 2}} |p_{\hat{t} - 1}'(0)|^{- {(\zeta + { 1 \over 2})}} )  \log {2 \over \delta} \\ 
& \leq 4  \log^{3 \over 2} {2 \over  \delta} \lambda^{\zeta} \max \left(   C_7   |p_{\hat{t} - 1}' (0) |^{-{1 \over 2}}  ,  R c_{3 \over 2} C_{\zeta-1,\kappa} \left(  C_3 +  C_1'  \right)  |p_{\hat{t} - 1}'(0)|^{- {3\over 2}} \lambda^{ - 1} , \right. \\
&\ \  \left.   2R C_{\zeta-{1\over 2},\kappa}  C_3     \lambda^{ - {1 \over 2}} |p_{\hat{t} - 1}'(0)|^{-1} 
, c_{\zeta + {1\over 2}} R  \lambda^{-\zeta}|p_{\hat{t}-1}'(0)|^{- ({\zeta + { 1 \over 2}})} \right).
\end{align*}
Therefore, if 
$$
\tau \geq \max \left( 4\sqrt{2} C_7, 8\sqrt{2} R c_{3 \over 2} C_{\zeta-1,\kappa} \left(  C_3 +  C_1'  \right) , 16 R C_{\zeta-{1\over 2},\kappa}  C_3 ,  2^{\zeta + {5 \over 2}} c_{\zeta+ {1\over 2}}R \right),
$$
then \eqref{eq:c3} holds, using a similar basic argument\\
{\it Step 2.} In this step, we let $u  = \lambda.$  Using \eqref{eq:c3} and Part 3) of Lemma \ref{lem:poly},  it is easy to show that $ u \leq |p_{\hat{t} - 1}'(0)|^{-1} \leq x_{1,\hat{t} - 1}. $
Applying Lemma \ref{lem:intermB}, introducing with \eqref{eq:biasInt2}, \eqref{eq:delta3p} and \eqref{eq:delta35p}, and by a direct calculation,
	\begin{align}
& [p_{\hat{t}-1}, p_{\hat{t}-1}]_{(0)}^{1\over 2}  \leq 
 C_8  \lambda^{\zeta+ {1\over 2}} \log^{3 \over 2}{2\over \delta} +
\lambda^{-{1\over 2}} [p_{\hat{t}-1}^{(2)}, p_{\hat{t}-1}^{(2)}]_{(1)}^{1\over 2}, \nonumber
\end{align} 
where
$$
C_8  = \sqrt{2}C_7 + R\left(   C_{\zeta-{1\over 2},\kappa} C_3 +  C_{\zeta-1,\kappa}(C_3 +C_1') + 1\right) .
$$
Combining with \eqref{eq:c4B}, we get  that 
$$
[p_{\hat{t}-1}, p_{\hat{t}-1}]_{(0)}^{1\over 2} \leq  {C_8 \over \tau + C_7}  \| \ao \omega_{\hat{t} - 1} - \proj\SX^* \NOutputs\|_{\HK}  +  \lambda^{-{1\over 2}} [p_{\hat{t}-1}^{(2)}, p_{\hat{t}-1}^{(2)}]_{(1)}^{1\over 2} \leq {1\over 2} [p_{\hat{t}-1}, p_{\hat{t}-1}]_{(0)}^{1\over 2} +  \lambda^{-{1\over 2}} [p_{\hat{t}-1}^{(2)}, p_{\hat{t}-1}^{(2)}]_{(1)}^{1\over 2},
$$
provided that
$$
\tau \geq  2 C_8 - C_7.
$$
This leads to \eqref{eq:d1}.  Combining with Part 4) of Lemma \ref{lem:poly} and \eqref{eq:c3}, we get that \eqref{eq:c5} holds. \\
{\it Step 3.} In this step, we let $u = {1 \over 5} \lambda$.  Then following from \eqref{eq:c5} and Part 3) of Lemma \ref{lem:poly}, we have \eqref{eq:d2}.  The rest of the proof parallelizes as that for the case $\zeta \leq 1.$ We thus include the sketch only. Applying Lemma \ref{lem:step1},  introducing with \eqref{eq:c5}, \eqref{eq:biasInt2}, \eqref{eq:delta3p} and \eqref{eq:delta35p},
\begin{align*}
\|\LK^{-a}(\IK \omega_{\hat{t}} - \FH) \|_{\rho} 
\leq C_9 
\lambda^{\zeta -a}\log^{2-a} {2 \over \delta}   +    C_1^{{1\over 2} - a} 6{\| \ao \omega_{\hat{t}} - \proj \SX^* \bby\|_{\HK}}   \lambda^{-{1 \over 2}-a} \log^{{1\over 2} -a } {2 \over \delta} ,
\end{align*}
where $$C_9 = C_1^{{1\over 2}-a} \left( C_7(5^a + 11) + R C_{\zeta - {1\over 2}, \kappa}(6/5)^{{1\over 2} - a} (C_3 + (C_3 + C_1')/\sqrt{5} + 5^{{1\over 2} - \zeta}) \right) + R (\kappa^{2(\zeta-1)} (C_1')^{1-a} + 1). $$
Following from the definition of the stopping rule, one can get the desired result for the case $\zeta \geq 1.$

The proof for \eqref{eq:resHK} with $\zeta \geq 1/2$ is the same as  we can replace
$\|\LK^{-a}(\IK \cgmt - \FH) \|_{\HK}$ by $\|\TK^{{1 \over 2}-a}( \cgmt - \omega_{\HK}) \|_{\HK}$  in the whole proof for the convergence with respect to $\LR$-norm.      
\end{proof}

\begin{proof}
	[Proof of Corollary \ref{cor:RanSke}]
	We use Theorem \ref{thm:gen} and Lemma \ref{lem:DZX} to prove the result. 
	We only need to verify \eqref{eq:projCond} is satisfied.  In Lemma \ref{lem:DZX}, we let
	$$
	\theta = \begin{cases}
	{\zeta -a \over (1-a)(2\zeta+\gamma)}, & \mbox{if } \zeta > 1, \\
	{1 \over 2\zeta+\gamma}, & \mbox{othwewise} , \\
	1, & \mbox{if } 2\zeta+\gamma\leq 1.
	\end{cases}
	$$
	Clearly, $\theta\leq 1.$
For $\theta<1,$ we have ${\log n^{\gamma} \over n^{1-\theta}} = {\gamma \log n^{1-\theta} \over (1 -\theta) n^{1-\theta}} \leq  {\gamma \over 1-\theta}.$ Therefore, following from Lemma \ref{lem:DZX} and Condition \eqref{eq:mnum}, we have that with probability at least $1 - \delta,$
with probability at least $1 - 3\delta$ ($\delta \in (0,1/3)$), we have
\bea
\| (I - \proj) \TK^{1\over 2} \|^2  \leq C'' \lambda^{\zeta \vee 1 - a \over 1 -a } \log{4 \over \delta},
\eea
with $C'' = {7 a_{\gamma}}$ if $\lambda = [1 \vee \log n^{\gamma}] /n$ or $C'' = {7 a_{\gamma} \over 1 - \theta}$ otherwise.
The proof is complete.
\end{proof}
\begin{proof}[Proof of Corollary \ref{thm:NySub}]
The proof for Corollary \ref{thm:NySub} can be done by using Theorem \ref{thm:gen} and  Lemma \ref{lem:DZI}. 
\end{proof}

Combining Theorem \ref{thm:gen} with Lemma \ref{lem:DZXALS}, we get the \rvo{following} result for KCGM with ALS Nystr\"{o}m sketches.

\begin{corollary}\label{thm:ALSNy}
	Under Assumptions \ref{as:noiseExp},  \ref{as:regularity} and \ref{as:eigenvalues}, let $\delta\in (0,1)$, $a \in [0, {\zeta} \wedge {1\over 2}]$, and  $\HKS = \overline{span\{\tilde{x}_1,\cdots,\tilde{x}_m\}}$ with $\tilde{x}_j$ i.i.d drawn according to the ALS Nystr\"{o}m subsampling regime in Lemma \ref{lem:DZXALS} (with an appropriate $\lambda$).  Assume that
		\be\label{eq:mnumALS}
	m \geq \tilde{C}_5 L^2 \log^3{3\over \delta}  \begin{cases}
		n^{\gamma}[1\vee \log n^{\gamma}]^{1-\gamma},  & \mbox{if } 2\zeta+\gamma \leq 1,\\
		n^{\gamma(\zeta-a) \over (1-a)(2\zeta+\gamma)} [1\vee \log n^{\gamma}],  & \mbox{if } \zeta \geq 1, \\
		n^{\gamma \over 2\zeta+\gamma} [1\vee \log n^{\gamma}]   & \mbox{otherwise},
	\end{cases}
	\ee
	for some $\tilde{C}_5>0$ (which depends only on
	$\zeta,\gamma,c_{\gamma}, \|\TK\|,\kappa^2, M, Q,B, R).$ Then the conclusions in Theorem \ref{thm:gen} are true.
\end{corollary}

%
%

 
 \section*{Acknowledgements}
 This manuscript version is made available under the CC-BY-NC-ND 4.0 license. 
 The authors would like to thank the editors and the  anonymous referees for their valuable suggestions. 
This work was sponsored by the Department of the Navy, Office of Naval Research (ONR) under a grant number N62909-17-1-2111.  It has also received funding from Hasler Foundation Program: Cyber Human Systems (project number 16066),  the European Research Council (ERC) under the European Union's Horizon 2020 research and innovation program (grant agreement n 725594-time-data), the NSF of China under grant numbers
11971427, 11901518, and “the Fundamental Research Funds for the Central Universities”+2020XZZX002-03.
 

\appendix

\section{Learning with Kernel Methods}\label{sec:learningKernel} 
Let $\Xi$ be a closed subset of Euclidean space $\mR^d$. Let $\mu$ be an unknown but fixed Borel probability measure on $\Xi \times Y$. Assume that $\mathbf \{(\xi_i, y_i)\}_{i=1}^n$ are i.i.d. from the distribution  $\mu$. A reproducing kernel $K$ is a symmetric function $K: \Xi
\times \Xi \to \mR$ such that $(K(u_i, u_j))_{i, j=1}^\ell$ is
positive semidefinite for any finite set of points
$\{u_i\}_{i=1}^\ell$ in $\Xi$. The kernel $K$ defines a reproducing
kernel Hilbert space (RKHS) $(\mathcal{H}_K, \|\cdot\|_K)$ as the
completion of the linear span of the set $\{K_{\xi}(\cdot):=K(\xi,\cdot):
\xi\in \Xi\}$ with respect to the inner product $\la K_{\xi},
K_u\ra_{K}:=K(\xi,u).$ For any $f \in \mathcal{H}_K$, the reproducing property holds: $f(\xi) = \la K_{\xi}, f\ra_K.$
In learning with kernel methods, one considers the following minimization problem
$$ \inf_{f\in \mathcal{H}_K} \int_{\Xi \times \mR} (f(\xi) - y)^2 d\mu(\xi,y).$$
Since $f(\xi) = \la K_{\xi},f\ra_{K} $ by the reproducing property, the above can be rewritten as
$$ \inf_{f\in \mathcal{H}_K} \int_{\Xi \times \mR} (\la f, K_{\xi} \ra_{K} - y)^2 d\mu(\xi,y).$$
Defining another probability measure
$\rho(K_{\xi},y) = \mu(\xi,y)$, the above reduces to \eqref{expectedRisk}.

\section{Proof of Lemmas}

\subsection{Proof of Lemma \ref{lem:hansen_ineq}}\label{appen:hansen_ineq}	Note that $X^* X \preceq I$ since $\|X\| \leq 1$. In fact, 
	$$
	\la X^* X \omega, \omega \ra_\HK = \| X \omega \|^2_\HK \leq  \|\omega\|^2_\HK = \la \omega, \omega \ra_\HK.
	$$
	Following from  \cite{hansen1980operator}, the fact that the function $u^s$ is operator monotone, one can prove \eqref{eq:3prod}:
	$$
	X^* (A+\lambda I )^s X   \preceq ( X^* A X + \lambda X^* X )^s \preceq ( X^* A X + \lambda I )^s .
	$$
	The proof for \eqref{eq:3prodRes} can be done by applying \eqref{eq:3prod}:
	\begin{align*}
	\|(A+ \lambda I )^{s \over 2} X \omega\|^2_\HK =   \la X^* (A+ \lambda I )^{s} X \omega, \omega \ra_\HK
	\leq  \la ( X^*A  X + \lambda I  )^{s}  \omega, \omega \ra_\HK = \| (X^*A X + \lambda I )^{s\over 2} \omega\|^2_\HK.
	\end{align*}
	The proof for \eqref{eq:3prodResB} can be done by applying \eqref{eq:3prod}:
	$$
	\| F X^* (A + \lambda I )^{s \over 2} \|^2  = 	\| F X^* (A + \lambda I )^{s} X F^* \|  \leq  \| F (X^* A X + \lambda I )^{s} F^* \| =   
	\| F (X^* A X + \lambda I )^{s \over 2} \|^2.
	$$

\subsection{Proof of Lemma \ref{lem:samProjErr}}\label{subapp:48}
	Adding and subtracting with the same term, and using the triangle inequality, 
	\begin{align}
	\| \TXL^{-{1\over 2}} ( \SX^* \bby  - \TX \proj \midFun)\|_{\HK}  \leq  &	\| \TXL^{-{1\over 2}} ( \SX^* \bby  - \TX \midFun)\|_{\HK} + \| \TXL^{-{1\over 2}} \TX (I - \proj) \midFun)\|_{\HK} \nonumber \\
	\leq  &  	\DZF^{1\over 2}  \DZS + \| \TXL^{-{1\over 2}} \TX (I - \proj) \midFun)\|_{\HK}. \label{eq:a1}
	\end{align}
	In what follows, we estimate $\| \TKL^{-{1\over 2}} \TX (I - \proj) \midFun)\|_{\HK}  $, considering two different cases. \\
	{\it Case I: $\zeta \leq 1$.}\\ 
	We have
	\begin{align*}
	\|  \TXL^{-{1\over 2}}\TX( I - \st) \midFun \|_{\HK} \leq \| \TXL^{-{1\over 2}} \TX\TXL^{-1\over 2}\| \|\TXL^{1\over 2} \TKL^{-{1\over 2}}\| \|\TKL^{1\over 2} ( I - \st) \midFun\|_{\HK} \leq \DZF^{1\over 2} \|\TKL^{1\over 2} ( I - \st) \midFun\|_{\HK} .
	\end{align*}
	Since $\st$ is a projection operator, 
	$(I-\st)^2 = I -\st$, and we thus have
	\begin{align*}
	\|  \TXL^{-{1\over 2}}\TX( I - \st) \midFun  \|_{\HK} \leq \DZF^{1\over 2} \|\TKL^{1\over 2} ( I - \st)\| \|(I-\st)\TK^{1\over 2}\|  \|\TK^{-{1\over 2}} \midFun \|_{\HK} \leq  
	\DZF^{1\over 2} \|\TKL^{1\over 2} ( I - \st)\| \DZI^{1\over 2}  R\lambda^{\zeta-1},
	\end{align*}
	where for the last inequality, we used Part 2) of Lemma \ref{lemma:midFun}.
	Note that for any $\omega \in \HK$ with $\|\omega\|_{\HK}=1,$
	\[
	\begin{split}
	\|\TKL^{1\over 2} ( I - \st)\omega\|_{\HK}^2 = &\la \TKL(I - \st) \omega, (I - \st)\omega \ra_{\HK}  
	=   \|\TK^{1\over 2}(I - \st) \omega\|_{\HK}^2 + \lambda \|(I-\st)\omega\|_{\HK}^2 \\
	\leq & \|\TK^{1\over 2} (I - \st)\|^2 + \lambda \leq \DZI + \lambda.
	\end{split}
\]
	It thus follows that 
	\be\label{eq:cc}
	\|\TKL^{1\over 2} ( I - \st)\| \leq (\DZI + \lambda)^{1\over2},\ee
	and thus
	\begin{align*}
	\|  \TXL^{-{1\over 2}}\TX( I - \st) \midFun  \|_{\HK}  \leq  
	\DZF^{1\over 2} (\DZI + \lambda) R\lambda^{\zeta-1}.
	\end{align*}
	Introducing the above into \eqref{eq:a1},  one can get the desired result for the case $\zeta \leq 1.$\\
	{\it Case II: \rvo{$\zeta> 1$}.}\\
	We first have
	\begin{align*}
	\|\TXL^{-{1\over 2}}\TX( I - \st) \midFun \|_{\HK}  
	\leq&  \DZF^{1\over 2}  \| \TKL^{-{1\over 2}}\TX( I - \st) \midFun) \|_{\HK} \\ 
	\leq&  \DZF^{1\over 2} \left(  \| \TKL^{-{1\over 2}}(\TX-\TK)( I - \st) \midFun \|_{\HK} + \| \TKL^{-{1\over 2}}\TK ( I - \st)\midFun\|_{\HK} \right)\\
	\leq&  \DZF^{1\over 2}  \left(  \DZN \|( I - \st) \midFun  \|_{\HK} + \| \TK^{1\over 2} ( I - \st) \midFun \|_{\HK} \right).
	\end{align*}	
	Since $\st$ is a projection operator, 
	$(I-\st)^2 = I -\st$,  we thus have
	\begin{align*}
	\|\TXL^{-{1\over 2}}\TX( I - \st) \midFun \|_{\HK}
	\leq &  \DZF^{1\over 2} \left( \DZN \|I - \st\| \|\TK^{1\over 2}\| \|\TK^{-{1\over 2}} \midFun  \|_{\HK} + \| \TK^{1\over 2} ( I - \st)\| \| ( I - \st) \TK^{1\over 2}\| \|\TK^{-{1\over 2}}\midFun \|_{\HK}  \right) \\
	\leq & 
	\DZF^{1\over 2} \left( \kappa\DZN  + \DZI   \right) \|\TK^{-{1\over 2}} \midFun  \|_{\HK} ,
	\end{align*}
	where we used \eqref{eq:TKBound} for the last inequality. 
	Applying Part 2) of Lemma \ref{lemma:midFun},  we get
	\begin{align*}
	\|\TXL^{-{1\over 2}}\TX( I - \st) \midFun \|_{\HK}
	\leq & 
	\DZF^{1\over 2} \left( \kappa\DZN  + \DZI   \right) \kappa^{2(\zeta-1)}R.
	\end{align*}
	Introducing the above into \eqref{eq:a1}, we get the desired result for \rvo{$\zeta> 1.$}

\subsection{Proof of Lemma  \ref{lem:operDifRes}}\label{subapp:operDiff}
We need the following lemma to prove the result.

\begin{lemma}
	\label{lem:concentrSelfAdjoint}
	Let $\mcX_1, \cdots, \mcX_m$ be a sequence of independently and identically distributed self-adjoint Hilbert-Schmidt operators on a separable Hilbert space.
	Assume that $\mE [\mcX_1] = 0,$ and $\|\mcX_1\| \leq B$ almost surely for some $B>0$. Let $\mathcal{V}$ be a positive trace-class operator such that $\mE[\mcX_1^2] \preccurlyeq \mathcal{V}.$
	Then with probability at least $1-\delta,$ ($\delta \in \rvo{(0,1)}$), there holds
	\bea
	\left\| {1 \over m} \sum_{i=1}^m \mcX_i \right\| \leq {2B \beta \over 3m} + \sqrt{2\|\mathcal{V}\|\beta \over m }, \qquad \beta = \log {4 \tr \mathcal{V} \over \|\mathcal{V}\|\delta}.
	\eea
\end{lemma}
The proof for the above result is based on 
the  lemma  in \cite[Theorem 7.7.1]{tropp2015introduction} 
for the matrix case, using the same argument for extending the result from the matrix case to the general  operator case in \cite{minsker2011some}. Refer to \cite{rudi2015less} for details.

Using the above lemma, we can prove the following result. Refer to \cite{rudi2015less,lin2018optimalconve} for proof details.
\begin{lemma}\label{lem:operDifEff}
	Let $0<\delta <1$ and $\lambda>0$. With probability at least $1-\delta,$ the following holds:
	\bea
	\left\| (\TK+\lambda)^{-1/2}(\TK - \TX)(\TK+\lambda)^{-1/2}  \right\| \leq {4\kappa^2 \beta  \over 3 {|\bx| }\lambda} + \sqrt{2\kappa^2 \beta  \over {|\bx|}\lambda}, \quad \beta = \log {4\kappa^2( \mcN(\lambda)+1) \over \delta \|\TK\|}.
	\eea
\end{lemma}

We are now ready to prove Lemma \ref{lem:operDifRes}.

\begin{proof}[Proof of Lemma  \ref{lem:operDifRes}]
  By a simple calculation, we have if $ 0\leq u  \leq {1/2},$ then
	$2u^2/3 + u\leq 2/3.$
	Letting $\sqrt{2\kappa^2\beta  \over |\bx|\lambda'}=u,$ and combining with Lemma \ref{lem:operDifEff},
	we know that if
	\bea
	\sqrt{2\kappa^2\beta \over |\bx|\lambda'}   \leq {1\over 2},
	\eea
	which is equivalent to
	\be\label{eq:num1}
	|\bx| \geq {8\kappa^2 \beta \over \lambda'}, \quad \beta  = \log{ 4\kappa^2( 1 + \mcN(\lambda')) \over \delta \|\TK\|},
	\ee
	then
	with probability at least $1-\delta,$
	\be\label{eq:A1}
	\left\| \TK_{\lambda'}^{-1/2}(\TK - \TX)\TK_{\lambda'}^{-1/2}  \right\| \leq 2/3.
	\ee
	Note that \eref{eq:A1} implies
	\be\label{eq:2b}
 \| \TK_{\lambda'}^{1/2}\TK_{\bx\lambda'}^{-1/2}\|^2 \vee \| \TK_{\bx\lambda'}^{1/2}\TK_{\lambda'}^{-1/2}\|^2 \leq 3.
	\ee
	Indeed,
	\bea
	\| \TK_{\lambda'}^{1/2}\TK_{\bx\lambda'}^{-1/2}\|^2 = \| \TK_{\lambda'}^{1/2}\TK_{\bx\lambda'}^{-1}\TK_{\lambda'}^{1/2}\|
	=\|(I - \TK_{\lambda'}^{-1/2}(\TK - \TX)\TK_{\lambda'}^{-1/2})^{-1} \| \leq 3,
	\eea
	and 
	$$\| \TK_{\bx\lambda'}^{1/2}\TK_{\lambda'}^{-1/2}\|^2 =  \| \TK_{\lambda'}^{-1/2} \TK_{\bx\lambda'}\TK_{\lambda'}^{-1/2}\| = 
	\| \TK_{\lambda'}^{-1/2} (\TX - \TK)\TK_{\lambda'}^{-1/2} + I \| \leq 3.
	$$
	From the above analysis, we know that for any fixed $\lambda'>0$ such that \eref{eq:num1}, then with probability at least $1-\delta,$
	\eref{eq:2b} holds.
	
Let $\lambda' = a \lambda $, where for notational simplicity,
	we denote  $a(\delta)$ by $a$. We will prove that the choice on $\lambda'$ ensures the condition \eref{eq:num1} is satisfied, and thus with probability at least $1-\delta,$
	\eref{eq:2b} holds. Obviously, one can easily prove that $a\geq 1.$ Therefore, $\lambda' \geq \lambda,$ and
	\bea
	\| \TK_{\lambda}^{1/2}\TK_{\bx\lambda}^{-1/2}\| \leq \| \TK_{\lambda}^{1/2} \TK_{\lambda'}^{-1/2}\|
	\|\TK_{\lambda'}^{1/2} \TK_{\bx\lambda'}^{-1/2}\| \| \TK_{\bx \lambda'}^{1/2} \TK_{\bx\lambda}^{-1/2}\| \leq \|\TK_{\lambda'}^{1/2} \TK_{\bx\lambda'}^{-1/2}\|\sqrt{\lambda'/\lambda},
	\eea
	where for the last inequality, we used $\| \TK_{\bx \lambda'}^{1/2} \TK_{\bx\lambda}^{-1/2}\|^2 \leq \sup_{u\geq 0} {u + \lambda' \over u+\lambda} \leq \lambda'/\lambda.$
	Similarly, 
	$$
	\| \TK_{\lambda}^{-1/2}\TK_{\bx\lambda}^{1/2}\|  \leq \|\TK_{\lambda'}^{-1/2} \TK_{\bx\lambda'}^{1/2}\|\sqrt{\lambda'/\lambda}.
	$$
	Combining with \eref{eq:2b}, and
	by a simple calculation, one can prove the desired bounds.
	What remains is to prove that the condition \eref{eq:num1} is satisfied.
	By Assumption \ref{as:eigenvalues} and $a\geq 1,$ for $\lambda = |\bx|^{-\theta}$ with $\theta \in[0,1)$,
	\bea
	\beta \leq \log{ 4\kappa^2( 1 + c_{\gamma} a^{-\gamma} |\bx|^{\theta\gamma}) \over \delta \|\TK\|} \leq  \log{ 4\kappa^2( 1 + c_{\gamma}) |\bx|^{\theta\gamma} \over \delta \|\TK\|} = \log{ 4\kappa^2( 1 + c_{\gamma})  \over \delta \|\TK\|} + \log |\bx|^{\theta\gamma},
	\eea
	while for $\lambda = (1 \vee \log |\bx|^{\gamma})/ |\bx|,$
	\bea
	\beta \leq \log{ 4\kappa^2( 1 + c_{\gamma} a^{-\gamma} \lambda^{-\gamma}) \over \delta \|\TK\|} \leq  \log{ 4\kappa^2( 1 + c_{\gamma}) |\bx|^{\gamma} \over \delta \|\TK\|} = \log{ 4\kappa^2( 1 + c_{\gamma})  \over \delta \|\TK\|} + \log |\bx|^{\gamma},
	\eea
	If  $\lambda = |\bx|^{-\theta}$ with $\theta\in [0,1)$ and $\theta\gamma=0$,  or $\lambda = (1 \vee \log |\bx|^{\gamma})/ |\bx|,$
	then the condition \eref{eq:num1}  follows trivially.
	Now consider the case $\lambda = |\bx|^{-\theta}$ with $\theta\in (0,1)$ and $\theta\gamma \neq 0$. 
	The maximum of the function $g(u) = \mathrm{e}^{-cu}u^{\alpha}$ (with $c>0$) over $ \mR_+ $ is achieved at $u_{\max}= \alpha/c,$ and thus
	\be\label{exppoly1}
	\sup_{u \geq 0} \mathrm{e}^{-cu} u^{\alpha} =  \left({\alpha \over \mathrm{e}c} \right)^{\alpha}.
	\ee
	 We
	apply the above with $u = |\bx|^{\theta\gamma \zeta'}$, $\alpha = 1/\zeta'$, we know that for any $c',\zeta'>0$
	\bea
	\beta \leq \log{ 4\kappa^2(1+c_{\gamma}) \over \delta \|\TK\|} + c' |\bx|^{\theta \gamma\zeta'} + {1 \over \zeta'} \log {1 \over \zeta' \mathrm{e}c'}.
	\eea
	Selecting $\zeta' = {1-\theta\over \theta\gamma}$ and  $c'={\theta \gamma \over \mathrm{e}(1-\theta)}$,
	we know that a sufficient condition for \eref{eq:num1} is
	\bea
	{|\bx|^{1-\theta}a \over 8 \kappa^2} \geq  \log{ 4\kappa^2(1+c_{\gamma}) \over \delta \|\TK\|} + {\theta\gamma \over \mathrm{e}(1-\theta)} |\bx|^{1-\theta}.
	\eea
	From the definition of $a$, and by a direct calculation, one can prove that the condition \eref{eq:num1} is satisfied.
	\end{proof}

\subsection{Proof of Lemma \ref{lem:DZS}}\label{subapp:variance}

To prove the result, we need the following concentration inequality.
\begin{lemma}
	\label{lem:Bernstein}
	Let $w_1,\cdots,w_m$ be i.i.d random variables in a separable Hilbert space with norm $\|\cdot\|$. Suppose that
	there are two positive constants $L$ and $\sigma^2$ such that
	\be\label{bernsteinCondition}
	\mE [\|w_1 - \mE[w_1]\|^l] \leq {1 \over 2} l! L^{l-2} \sigma^2, \quad \forall l \geq 2.
	\ee
	Then for any $0< \delta <1/2$, the following holds with probability at least $1-\delta$,
	$$ \left\| {1 \over m} \sum_{k=1}^m w_m - \mE[w_1] \right\| \leq 2\left( {L \over m} + {\sigma \over \sqrt{ m }} \right) \log {2 \over \delta} .$$
	In particular, \eref{bernsteinCondition} holds if
	\be\label{bernsteinConditionB}
	\|w_1\| \leq L/2 \ \mbox{ a.s.}, \quad \mbox{and } \quad \mE [\|w_1\|^2] \leq \sigma^2.
	\ee
\end{lemma}
The above lemma is a reformulation of the concentration inequality for sums of Hilbert-space-valued random variables from \cite{pinelis1986remarks}.
We refer to  \cite{caponnetto2007optimal} for the detailed proof.

\begin{proof}[Proof of Lemma \ref{lem:DZS}]
	Using the triangle inequality, we  have 
	\begin{align*}
	\|\TKL^{-{1\over 2}}(\TX\midFun - \SX^*\NOutputs) \|_{\HK}  
	\leq  \|\TKL^{-{1\over 2}}(\TX\midFun - \SX^*\NOutputs - \TK \midFun + \IK^* \FH) \|_{\HK} + \|\TKL^{-{1\over 2}}( \TK \midFun -  \IK^* \FH) \|_{\HK}. 
	\end{align*}
	Note that $\TK = \IK^* \IK$, $\|\TKL^{-{1\over 2}} \IK^*\|  = 
	\|\TKL^{-{1\over 2}} \IK^* \IK \TKL^{-{1\over 2}}\|^{1 \over 2} 
	\leq 1$, and 
	$$
	\|\TKL^{-{1\over 2}}( \TK \midFun -  \IK^* \FH) \|_{\HK} = \|\TKL^{-{1\over 2}}\IK^*( \IK \midFun -  \FH) \|_{\HK} \leq 
	\|\TKL^{-{1\over 2}}\IK^* \| \| \IK \midFun -  \FH \|_{\rho} \leq R \lambda^{\zeta}, 
	$$ 
	where we used Lemma \ref{lemma:midFun} for the last inequality.
	Therefore, 
	\begin{align}\label{eq:nonsen1}
	\|\TKL^{-{1\over 2}}(\TX\midFun - \SX^*\NOutputs) \|_{\HK}  
	\leq  \|\TKL^{-{1\over 2}}(\TX\midFun - \SX^*\NOutputs - \TK \midFun + \IK^* \FH) \|_{\HK} + R \lambda^{\zeta}. 
	\end{align}
	In what follows, we use Lemma \ref{lem:Bernstein} to estimate the first term of the right-hand side from the above. We  let $\xi_i = \TKL^{-{1\over 2}} (\la \midFun, x_i\ra_{\HK} - y_i) x_i$ for all $i \in [n].$
	 It is easy to see that $\xi_i$ is a random variable depending on $(x_i, y_i).$ 
	From the definition of the regression function $\FR$ in \eref{regressionfunc} and \eqref{frFH}, a simple calculation shows that
	\be\label{eq:interm1}
	\mE[\xi] = \mE[\TKL^{-{1\over 2}} (\la \midFun, x\ra_{\HK} - \FR(x)) x] = \TKL^{-{1\over 2}}(\TK \midFun - \IK^* \FR) =  \TKL^{-{1\over 2}}(\TK \midFun - \IK^* \FH).
	\ee
	Combining with the definition of $\TX$ and $\SX^*$, we have 
	$$
	\|\TKL^{-{1\over 2}}(\TX\midFun - \SX^*\NOutputs - \TK \midFun + \IK^* \FH) \|_{\HK}  = \left\| {1 \over n} \sum_{i=1}^n (\xi_i - \mE[\xi]) \right\|_{\HK}
	$$
	In order to apply Lemma \ref{lem:Bernstein}, we need to estimate $\mE[\|\xi - \mE[\xi]\|_{\HK}^l]$ for any $l \in \mN$ with $l\geq 2.$
	In fact, using H\"{o}lder's inequality twice,
	\begin{align}\label{eq:interm6}
	\mE\|\xi - \mE[\xi]\|_{\HK}^l  \leq  \mE\left(\|\xi\|_{\HK}+ \mE\|\xi\|_{\HK}\right)^l \leq 2^{l-1} (\mE \|\xi\|_{\HK}^l + (\mE\|\xi \|_{\HK})^l) \leq 2^{l} \mE \|\xi\|_{\HK}^l. 
	\end{align}
	We now estimate $\mE \|\xi\|_{\HK}^l.$ By H\"{o}lder's inequality,
	\begin{align*}
	\mE \|\xi\|_{\HK}^l = \mE [\|\TKL^{-{1\over 2}}x\|_{\HK}^l (y- \la\midFun,x\ra_{\HK})^l]  \leq 2^{l-1}\mE [\|\TKL^{-{1\over 2}}x\|_{\HK}^l (|y|^l +  |\la \midFun,x\ra_{\HK}|^l)].
	\end{align*}
	According to \eqref{boundedKernel}, one has 
	\be\label{eq:interm2o}
	\|\TKL^{-{1\over 2}} x\|_{\HK} \leq \|\TKL^{-{1\over 2}}\| \|x\|_{\HK} \leq {1\over \sqrt{\lambda}} \kappa.\ee Moreover, by Cauchy-Schwarz inequality and \eqref{boundedKernel}, $|\la \midFun, x \ra_{\HK} |\leq \|\midFun\|_{\HK} \|x\|_{\HK} \leq \kappa \|\midFun\|_{\HK} .$ Thus, we get
	\begin{align}\label{eq:interm5}
	\mE \|\xi\|_{\HK}^l  \leq 2^{l-1} \left({\kappa \over \sqrt{\lambda}} \right)^{l-2}\mE [\|\TKL^{-{1\over 2}}x\|_{\HK}^2 (|y|^l + (\kappa \|\midFun\|_{\HK})^{l-2} |\la\midFun,x\ra_{\HK}|^2)].
	\end{align}
	Note that by \eqref{noiseExp},
	\begin{align*}
	\mE [\|\TKL^{-{1\over 2}}x\|_{\HK}^2 |y|^l ] =& \int_{\HK} \|\TKL^{-{1\over 2}}x\|_{\HK}^2 \int_{\mR} |y|^l d\rho(y|x) d\rho_X(x) \\
	\leq & {1\over 2} l! M^{l-2} Q^2 \int_{\HK} \|\TKL^{-{1\over 2}}x\|_{\HK}^2 d\rho_X(x). 
	\end{align*}
	Using $\|w\|_{\HK}^2 = \tr(w\otimes w)$ which implies 
	\be\label{eq:interm3o}
	\int_{\HK}  \|\TKL^{-{1\over 2}}x\|_{\HK}^2 d\rho_X(x) = \int_{\HK}  \tr(\TKL^{-{1\over 2}}x\otimes x \TKL^{-{1\over 2}}) d\rho_X(x) = \tr(\TKL^{-{1\over 2}}\TK \TKL^{-{1\over 2}})   = \mcN(\lambda),
	\ee
	we get
	\begin{align}\label{eq:interm4o}
	\mE [\|\TKL^{-{1\over 2}}x\|_{\HK}^2 |y|^l ] 
	\leq  {1\over 2} l! M^{l-2} Q^2 \mcN(\lambda). 
	\end{align}
	Besides, by Cauchy-Schwarz inequality, $$
	\mE [\|\TKL^{-{1\over 2}}x\|_{\HK}^2 |\la\midFun,x\ra_{\HK}|^2] \leq 3	\mE [\|\TKL^{-{1\over 2}}x\|_{\HK}^2(|\la\midFun,x\ra_{\HK} - \FH(x)|^2 + |\FH(x) - \FR(x)|^2 +|\FR(x)|^2 )].$$ 
	By \eqref{eq:interm2o} and \eqref{eq:trueBias},
	$$ 
	\mE [\|\TKL^{-{1\over 2}}x\|_{\HK}^2(|\la\midFun,x\ra_{\HK} - \FH(x)|^2] \leq {\kappa^2 \over \lambda} \mE[|\la\midFun,x\ra_{\HK} - \FH(x)|^2] = {\kappa^2 \over \lambda} \|\IK \midFun - \FH\|_{\rho}^2 \leq R^2\kappa^2 {\lambda^{2\zeta - 1}},
	$$
	and by \eqref{eq:bounRegFunc} and \eqref{eq:interm3o}, 
	$$\mE [\|\TKL^{-{1\over 2}}x\|_{\HK}^2 |\FR(x)|^2] \leq Q^2 \mE [\|\TKL^{-{1\over 2}}x\|_{\HK}^2] = Q^2 \mcN(\lambda).
	$$
	Therefore, 
	$$
	\mE [\|\TKL^{-{1\over 2}}x\|_{\HK}^2 |\la\midFun,x\ra_{\HK}|^2] \leq 3\left( R^2\kappa^2 \lambda^{2 \zeta - 1} + 
	\mE [\|\TKL^{-{1\over 2}}x\|_{\HK}^2 |\FH(x) - \FR(x)|^2] +Q^2 \mcN(\lambda) \right).$$ 
	Using $\|w\|_{\HK}^2 = \tr(w\otimes w)$  and \eqref{eq:FHFR}, we have
	\begin{align*}
	\mE [\|\TKL^{-{1\over 2}}x\|_{\HK}^2 |\FH(x) - \FR(x)|^2]  = & \mE [ |\FH(x) - \FR(x)|^2\tr(\TKL^{-{1\over 2}}x\otimes x \TKL^{-{1\over 2}})] \\
	=&  \tr(\TKL^{-1}\mE [(\FH(x) - \FR(x))^2 x\otimes x] ) \\
	\leq& B^2\tr(\TKL^{-1} \TK) = B^2\mcN({\lambda}),
	\end{align*}
	and therefore,
	$$
	\mE [\|\TKL^{-{1\over 2}}x\|_{\HK}^2 |\la\midFun,x\ra_{\HK}|^2] \leq 3\left( \kappa^2 R^2 \lambda^{ 2\zeta - 1} + 
	(B^2+Q^2) \mcN(\lambda) \right).$$ 
	Introducing the above estimate and \eqref{eq:interm4o} into \eqref{eq:interm5}, we derive
	\begin{align*}
	\mE\|\xi\|_{\HK}^l \leq& 2^{l-1} \left( {\kappa \over \sqrt{\lambda}} \right)^{l-2}\left( {1\over 2} l! M^{l-2} Q^2 \mcN(\lambda) + 3(\kappa \|\midFun\|_{\HK})^{l-2}( R^2 \kappa^2 \lambda^{2\zeta - 1} + (B^2+ Q^2)\mcN(\lambda)) \right) \\
	\leq & 2^{l-1} \left( {\kappa  M + \kappa^2\|\midFun\|_{\HK}\over \sqrt{\lambda}} \right)^{l-2} {1\over 2} l! \left(  Q^2 \mcN(\lambda) + 3( R^2 \kappa^2  \lambda^{ 2\zeta - 1} + (B^2+ Q^2)\mcN(\lambda)) \right),\\
	\leq & 2^{l-1} \left( { \kappa  M + \kappa^2\|\midFun\|_{\HK}\over \sqrt{\lambda}} \right)^{l-2} {1\over 2} l! \left(   3 R^2 \kappa^2 \lambda^{2\zeta - 1} + (3B^2+ 4Q^2)c_{\gamma} \lambda^{-\gamma} \right),
	\end{align*}
	where for the last inequality, we used Assumption \ref{as:eigenvalues}.
	Introducing the above estimate into \eref{eq:interm6}, and then substituting with \eref{eq:popSeqNorm},
	we get
	$$
	\mE[\|\xi - \mE[\xi]\|_{\HK}^l] \leq {1\over 2} l! \left( {4\kappa(M + \kappa^{1\vee (2\zeta)} R \lambda^{(\zeta - {1\over 2})_-}) \over \sqrt{\lambda} } \right)^{l-2} 8\left(  3 R^2 \kappa^2 \lambda^{ 2\zeta - 1} + (3B^2+ 4Q^2)c_{\gamma} \lambda^{-\gamma} \right).
	$$
	Applying Lemma \ref{lem:Bernstein}, we get that with probability at least $1 - \delta,$ 
	\begin{align*}
	& \|\TKL^{-{1\over 2}}(\TX\midFun - \SX^*\NOutputs - \TK \midFun + \IK^* \FH) \|_{\HK} \\
	& \leq 2 \left( {4\kappa(M +  \kappa^{1\vee (2\zeta)} R \lambda^{(\zeta - {1\over 2})_-}) \over n \sqrt{\lambda}} +  \sqrt{ 8(3 R^2 \kappa^2 \lambda^{2\zeta - 1} + (3B^2 + 4 Q^2) c_{\gamma} \lambda^{-\gamma} ) \over n}\right)  \log{2 \over \delta} .
	\end{align*}
	Introducing the above into \eqref{eq:nonsen1}, one can prove the desired result.\\
\end{proof}

\subsection{Estimating Projection Errors with Plain Nystr\"{o}m Sketches}\label{subapp:PNy}

\begin{proof}[Proof of Lemma \ref{lem:DZI}]
	As $\proj$ is the projection operator onto $\overline{range\{\SXS^*\}}$ with $\tilde{\bx} = \{x_1,\cdots, x_m\}$,
	$$
	\proj = \SXS^* (\SXS \SXS^*)^{\dagger} \SXS \succeq  \SXS^* (\SXS \SXS^* + \eta I)^{-1} \SXS = \SXS^*\SXS  ( \SXS^* \SXS + \eta I)^{-1}  = \TXS (\TXS + \eta I)^{-1},
	$$
	where for the last second equality, we used Lemma \ref{lem:sss}. Thus,
	$$
	I - \proj \preceq I - \TXS (\TXS + \eta I)^{-1}  = \eta (\TXS + \eta I)^{-1}.
	$$
	It thus follows that
	$$
	\TK_{\eta}^{1\over 2}(I - \proj) ^{1\over 2} \TK_{\eta}^{1\over 2} \preceq \eta\TK_{\eta}^{1\over 2} (\TXS + \eta I)^{-1} \TK_{\eta}^{1\over 2}. 
	$$ 
	Using $\|A^*A\| = \|A\|^2$ and the above,
	\begin{align}
	\|(I - \proj) \TK_{\eta}^{1\over 2} \|^2 =   \|\TK_{\eta}^{1\over 2}(I - \proj) \TK_{\eta}^{1\over 2} \|
	\leq \eta \|\TK_{\eta}^{1\over 2} (\TXS + \eta I)^{-1} \TK_{\eta}^{1\over 2}\| 
	=  \eta\| ( \TXS + \eta I)^{-1/2} \TK_{\eta}^{1\over 2}\|^2. \label{eq:t5}
	\end{align}
	Thus,
	$$
	\|(I - \proj)\TK^{1\over 2}\|^2 \leq  \|(I - \proj)\TK_{\eta}^{1\over 2}\|^2 \leq  \eta \|(\TXS+\eta I)^{-1/2}(\TK+\eta I)^{1/2}\|^2.
	$$
	Using Lemma \ref{lem:operDifRes} with $\eta = {1 \vee \log m^{\gamma} \over m}$, one can prove the desired result.
\end{proof}

\subsection{Estimating Projections Errors with Randomized Sketches}\label{subapp:ske}
In this subsection, we prove Lemma \ref{lem:DZX}.
The basic idea of the proof is to approximate
$\|(I - \proj) \TK^{1\over 2}\|^2$ in terms of its ``empirical version", $\|(I - \proj)\TX^{1\over 2}\|^2$. The term   $\|(I - \proj)\TX^{1\over 2}\|^2$ can be estimated using the following lemma.

\begin{lemma}\label{lem:OperDiffProd}
	Let $0<\delta <1$ and $\theta \in[0,1].$
	Given a fixed input set $\bx \subseteq \HK^n$, assume that for $\lambda \in [0,1]$,
	\be\label{eq:empEffDim}
	\tr ((\TX+\lambda)^{-1}\TX) \leq b_{\gamma} \lambda^{-\gamma}
	\ee
	holds for some $b_{\gamma}>0$, $\gamma \in[0,1]$.
	Then there exists a subset $U_{\bx} $ of $\mR^{m\times n}$ with measure at least $1-\delta$, such that for all $\skt \in U_{\bx}$, 
	$$
	\|(I - \proj)\TX^{1\over 2}\|^2 \leq  6\lambda,
	$$
	provided that
	\be\label{eq:subsamLev}
	m \geq 100c_0' \log^{\beta}n \lambda^{-\gamma}\log{3 \over \delta} \left( 1 + 10 b_{\gamma}   \right) .
	\ee	
\end{lemma}
The proof for the above lemma can be found in \cite{lin2020convergences}. We provide a  proof here.
\begin{proof}
	Let $\SX = U \Sigma V^* $ be the singular value decomposition of
	$\SX$, where $V:  \mR^{r} \to \HK,$ $U \in \mR^{n \times r}$ and $\Sigma = \mbox{diag}(\sigma_1,\sigma_2,\cdots, \sigma_r)$ with $V^* V = I_{r}$, $U^*U = I_r$ and $\sigma_1 \geq \sigma_2,\cdots, \sigma_r>0.$
	In fact, we can write $V = [v_1, \cdots, v_r]$ with
	$$
	V \ba = \sum_{i=1}^r \ba(i)v_i, \quad \forall \ba \in \mR^r,
	$$
	with $v_i \in \HK$ such that $\la v_i, v_j \ra_{\HK} = 0$ if $i \neq j$ and $\la v_i, v_i \ra_{\HK} =1 $. Similarly, we write $U = [u_1,\cdots, u_r]$, and 
	$$
	\SX = \sum_{i=1}^r \sigma_i \la v_i, \cdot \ra_{\HK}  u_i = \sum_{i=1}^r \sigma_i u_i \otimes v_i.
	$$
For any $\mu\geq 0$, we decompose $\SX$ as $\mcS_{1,\mu} + \mcS_{2,\mu}$ with 
	$$
	\mcS_{1,\mu} = \sum_{\sigma_i >\mu }\sigma_i u_i \otimes v_i, \quad  \mcS_{2,\mu} = \sum_{\sigma_i \leq \mu  }\sigma_i u_i \otimes v_i,
	$$
	and we will drop $\mu$ to
	write $\mcS_{j,\mu}$ as $\mcS_j$ when it is clear in the text.
	Denote $d$ the cardinality of $\{\sigma_i: \sigma_i>\mu\}$.
	Correspondingly,
	\be\label{eq:S1}
	\mcS_1 = U_1 \Sigma_1 V_1^*, \quad \mcS_2 = U_2 \Sigma_2 V_2^*, 
	\ee
	where $V_1 = [v_1, \cdots, v_{d}]$, $V_2 = [v_{d+1} ,\cdots,v_r],$ $U_1 = [u_1,\cdots, u_d]$,
	$
	U_2 = [u_{d+1},\cdots,u_r],
	$ 
	$
	\Sigma_1 = \mbox{diag}(\sigma_1,\cdots, \sigma_d),
	$
	and $\Sigma_2 = \mbox{diag}(\sigma_{d+1},\cdots,d_r).$
	As the range of $\proj$ is $range(\SX^* \skt^*)$, we can let 
	$$
	\proj = \proj_1 + \proj_2,
	$$
	where $\proj_1$ and $\proj_2$ are projection operators on $range(\mcS_1^* \skt^*)$ and  $range(\mcS_2^* \skt^*)$, respectively.

	As $$\TX = \SX^* \SX =  (U \Sigma V^*)^* U \Sigma V^* = V \Sigma^2 V^* ,$$ we have
	$$
	\|(I - \proj) \TX^{1\over 2}\| =  \|(I - \proj) V \Sigma V^*\| = \|(I - \proj_1 - \proj_2) \sum_{i=1}^2V_i \Sigma_i V_i^*\|.
	$$ 
	As $\proj_1$  is a projection operator on $range(\mcS_1^* \skt^*) (\subseteq range(V_1))$ and  $range(\mcS_1^* \skt^*) (\subseteq range(V_2))$, and $V_1^* V_2 = \bf{0},$ we know that
	$\proj_i V_j = 0$ when $i\neq j$.
	Thus, it follows that
	\begin{align*}
	\|(I - \proj) \TX^{1\over 2}\| = & \|\sum_{i=1}^2(I - \proj_i) (V_i \Sigma_i V_i^*)\| \\
	\leq& \sum_{i=1}^2\|(I - \proj_i) (V_i \Sigma_i V_i^*)\| \\
	\leq& \|(I - \proj_1) (V_1 \Sigma_1 V_1^*)\| + \|I - \proj_2\| \|V_2\|  \|\Sigma_2\|  \|V_2^*\|.
	\end{align*}
	As $\Sigma_2 = diag(\sigma_{d+1},\cdots, \sigma_r)$ with $\sigma_r\leq ,\cdots, \sigma_{d+1} \leq \mu,$ we get
	\be\label{eq:s5}
	\|(I - \proj) \TX^{1\over 2}\|  \leq  \|(I - \proj_1) (V_1 \Sigma_1 V_1^*)\| + \mu.
	\ee
	As $\proj_1$ is the projection operator on $range(\mcS_1^* \skt^*)$, letting $W = \skt \mcS_1$ and for any $\lambda>0$,
	$$\proj_1 = W^* (W W^*)^{\dagger} W \succeq  W^* (W W^* + \lambda I)^{-1} W =  W^* W (W^*W + \lambda I)^{-1},
	$$
	and thus
	$$
	I - \proj_1 \preceq I -   W^* W (W^*W + \lambda I)^{-1} = \lambda (W^*W + \lambda I)^{-1}.
	$$
	It thus follows that
	$$
	T_1^{1\over 2} (I - \proj_1) T_1^{1\over 2} \preceq \lambda T_1^{1\over 2} (W^*W + \lambda I)^{-1} T_1^{1\over 2},
	$$
where for notational simplicity, we write 
	\be \label{eq:T1}T_1 =  (V_1 \Sigma_1 V_1^*)^2.\ee
	Combining with 
	$$\| (I - \proj) T_1^{1\over 2}\|^2 = \| T_1^{1\over 2}(I - \proj)^2 T_1^{1\over 2}\| =
	\| T_1^{1\over 2} (I - \proj) T_1^{1\over 2}\| ,
	$$
	we know that
$$
	\| (I - \proj) T_1^{1\over 2}\|^2  \preceq \lambda \| T_1^{1\over 2} (W^*W + \lambda I)^{-1}T_1^{1\over 2} \| \leq  \lambda \| T_{1\lambda}^{1\over 2} (W^*W + \lambda I)^{-1} T_{1\lambda}^{1\over 2}\|.
	$$
	As 
	$$
	T_{1\lambda}^{1\over 2} (W^*W + \lambda I)^{-1}T_{1\lambda}^{1\over 2} = \left( T_{1\lambda}^{-{1\over 2}} (W^*W + \lambda I)T_{1\lambda}^{- {1\over 2}} \right)^{-1} =
	\left(I -  T_{1\lambda}^{-{1\over 2}}(T_1 - W^*W )T_{1\lambda}^{-{1\over 2}} \right)^{-1}, 
	$$
	and if 
	\be\label{eq:s4}
	\|T_{1\lambda}^{-{1\over 2}}(T_1 - W^*W )T_{1\lambda}^{-{1\over 2}}\|  \leq c<1,
	\ee
	then according to Neumann series,
		\be\label{eq:s8}
	\| (I - \proj) T_1^{1\over 2}\|^2  \preceq \lambda \| T_{1\lambda}^{-{1\over 2}} (W^*W + \lambda I)^{-1}T_{1\lambda}^{-{1\over 2}} \| \leq (1 - c)^{-1}\lambda.
	\ee
	If we	choose $\mu = \sqrt{\lambda},$ and introduce the above with $c={1\over 2}$  into \eqref{eq:s5}, one can get 
	\be
	\|(I - \proj) \TX^{1\over 2}\|^2  \leq  (\sqrt{2} + 1)^2\lambda \leq 6\lambda,
	\ee
	which leads to the desired bound.
	
	In what follows, we show that	\eqref{eq:s4}  with $c={1\over 2}$  holds with high probability under the constraint \eqref{eq:subsamLev}. Recall \eqref{eq:T1} and that $W = \skt \mcS_1$ with $\mcS_1$ given by \eqref{eq:S1}.
	Thus, 
	$
	T_1 =V_1 \Sigma_1 V_1^* V_1 \Sigma_1 V_1^* = V_1 \Sigma_1^2 V_1^*,
	$
	and 
	$$
	W^* W =  \mcS_1^* \skt^* \skt \mcS_1 = V_1 \Sigma_1 U_1^* \skt^* \skt U_1 \Sigma_1 V_1^*.
	$$
	Therefore, with $V_1^* V_1 = I,$
	\begin{align}
	T_{1\lambda}^{-{1\over 2}}(T_1 - W^*W )T_{1\lambda}^{-{1\over 2}} = &
	V_1 (\Sigma_{1}^2 + \lambda I)^{-1/2} V_1^*
	V_1 \Sigma_1 ( I - U_1^* \skt^* \skt U_1 )\Sigma_1 V_1^*
	V_1 (\Sigma_{1}^2 + \lambda I)^{-1/2} V_1^* \nonumber \\
	=& V_1 (\Sigma_{1}^2 + \lambda I)^{-1/2} \Sigma_1 ( I - U_1^* \skt^* \skt U_1 )\Sigma_1 (\Sigma_{1}^2 + \lambda I)^{-1/2} V_1^*. \label{eq:s7}
	\end{align}
	It follows that
	$$
	\| T_{1\lambda}^{-{1\over 2}}(T_1 - W^*W )T_{1\lambda}^{-{1\over 2}} \| \leq 
	\| V_1\|  \|(\Sigma_{1}^2 + \lambda I)^{-1/2} \Sigma_1\|^2 \|  I - U_1^* \skt^* \skt U_1  \| \| V_1^*\| 
	\leq  \|  I - U_1^* \skt^* \skt U_1  \|.
	$$
	Using $U_1^* U_1 = I$,
	\begin{align*}
	\|  I - U_1^* \skt^* \skt U_1  \| =  &  \|   U_1^*(I - \skt^* \skt) U_1  \|  \\
	= & \max_{\ba \in \mR^d, \|\ba\|_2=1} | \la U_1^*(I - \skt^* \skt) U_1 \ba, \ba  \ra_2 | \\
	= & \max_{\ba \in \mR^d, \|\ba\|_2=1} | \|U_1 \ba\|_2^2  - \| \skt U_1\ba \|_2^2 |.
	\end{align*}
	Based on a standard argument as that for \cite[Lemma 5.1]{baraniuk2008simple}, we know that
	$$\max_{\ba \in \mR^d, \|\ba\|_2=1}| \|U_1 \ba\|_2^2  - \| \skt U_1\ba \|_2^2 |\leq {1 \over 2} $$ with probability at least 
	$$
	1 - 2 (60)^d \exp\left(- {m \over 100 c_0' \log^{\beta}n} \right) \geq 1-\delta, 
	$$
	provided that
	\be\label{eq:s6}
	m \geq 100 c_0' \log^{\beta} n\left( \log {2 \over \delta}+ 5d\right).
	\ee
	Note that by \eqref{eq:empEffDim}
	$$
	b_{\gamma}\lambda^{-\gamma} \geq \tr(\TX \TXL^{-1}) = \sum_{i} {\sigma_i^2 \over \sigma_i^2 + \lambda} \geq \sum_{\sigma_i^2 >\lambda} {\sigma_i^2 \over \sigma_i^2 + \lambda}  \geq {d \over 2}.
	$$
	Thus, a stronger condition for \eqref{eq:s6} is
	\eqref{eq:subsamLev}.
	The proof is complete.
\end{proof}
In order to use Lemma \ref{lem:OperDiffProd} to prove the result, we need to estimate the ``empirical" effective dimension $\tr(\TXL^{-1}\TX)$, which can be done by using the following lemma. 
\begin{lemma}\label{lem:empEffdim}
	Under Assumption \ref{as:eigenvalues}, let $0<\delta <1$. 
	For any fixed $\lambda = n^{-\theta}$ with $\theta\in[0,1)$, or  $\lambda = {1\vee \log n^{\gamma } \over n},$ with probability at least $1 -\delta$, the following holds:
	\be\label{eq:empEffdim}
	\begin{split}
		&\tr ((\TX+\lambda I )^{-1}\TX)  \leq b_{\gamma} \log^2 {4 \over \delta}  \lambda^{-\gamma}.
	\end{split}
	\ee
Here, $b_{\gamma}$ is a positive constant given by
$$
b_{\gamma} = 24\kappa^2 (4\kappa^2 + 2\kappa\sqrt{c_{\gamma}} + c_{\gamma}) \left(\log {2\kappa^2(c_{\gamma} + 1) \over \|\TK\|}+ 1  + \tilde{c}\right),\quad \tilde{c} = \begin{cases}
1, & 
\mbox{if }  \lambda = {1 \vee \log n^{\gamma} \over n} , \\
{\theta\gamma \over \mathrm{e}(1-\theta)},&  \mbox{otherwise}.
\end{cases}
$$
\end{lemma}
\begin{proof}
	The proof can be found in \cite{lin2020convergences}.
			We first use Lemma \ref{lem:Bernstein} to estimate $\tr(\TKL^{-{1\over 2}} (\TX - \TK) \TKL^{-{1\over 2}}).$ Note that
		\bea
		\tr(\TKL^{-{1\over2}}\TX \TKL^{-{1\over 2}}) = {1\over n} \sum_{j=1}^n \|\TKL^{-{1\over 2}} x_j\|_{\HK}^2 = {1\over n} \sum_{j=1}^n \xi_j,
		\eea
		where we
		let $\xi_j = \|\TKL^{-{1\over 2}} x_j\|_{\HK}^2$ for all $j \in [n].$ Besides, it is easy to see that 
		$$
		\tr(\TKL^{-{1\over 2}} (\TX - \TK) \TKL^{-{1\over 2}}) = {1\over n}\sum_{j=1}^n (\xi_j - \mE[\xi_j]).
		$$
		Using Assumption \eqref{boundedKernel},
		$$
		\xi_1 \leq {1\over \lambda} \|x_1\|_{\HK}^2 \leq {\kappa^2 \over \lambda},
		$$
		and
		$$
		\mE[\|\xi_1\|^2] \leq {\kappa^2 \over \lambda} \mE\|\TKL^{-{1\over 2}} x_1\|_{\HK}^2\leq {\kappa^2 \mcN(\lambda) \over \lambda}.
		$$
		Applying Lemma \ref{lem:Bernstein}, we get that there exists a subset $\Omega_1$ of $Z^n$ with measure at least $1-\delta$, such that for all $\bz \in \Omega_1$,
		$$
		\tr(\TKL^{-{1\over2}}(\TX - \TK) \TKL^{-{1\over 2}}) \leq 2\left( {2\kappa^2 \over n\lambda} + \sqrt{\kappa^2 \mcN(\lambda) \over  n\lambda} \right) \log {2 \over \delta} .
		$$
		Combining with Lemma \ref{lem:operDifRes}, taking the union bounds, rescaling $\delta$, and noting that
		\begin{align*}
		\tr(\TXL^{-1}\TX) = & \tr(\TXL^{-{1\over2 }}\TKL^{1\over 2}\TKL^{-{1\over 2}} \TX \TKL^{-{1\over 2}}\TKL^{{1\over 2}}\TXL^{-{1\over 2}}) \\ 
		\leq& \|\TKL^{{1\over 2}}\TXL^{-{1\over 2}}\|^2
		\tr(\TKL^{-{1\over 2}} \TX \TKL^{-{1\over 2}}) \\
		= & \|\TKL^{{1\over 2}}\TXL^{-{1\over 2}}\|^2 \left( \tr(\TKL^{-{1\over 2}} (\TX - \TK) \TKL^{-{1\over 2}} ) + \mcN(\lambda) \right).
		\end{align*}
		we get that there exists a subset $\Omega$ of $Z^n$ with measure at least $1-\delta$, such that for all $\bz \in \Omega$, 
		\bea
		\tr (\TXL^{-1}\TX) \leq 3 a({\delta / 2}) \left(2\left( {2\kappa^2 \over n\lambda} + \sqrt{\kappa^2 \mcN(\lambda) \over  n\lambda} \right) \log {4 \over \delta}   +\mcN(\lambda) \right),
		\eea
		which leads to the desired result using $\lambda \leq 1$, $n\lambda\geq 1$ and Assumption \ref{as:eigenvalues}.
	\end{proof}

Now, we are ready to prove Lemma \ref{lem:DZX}.
\begin{proof}[Proof of Lemma \ref{lem:DZX}]
	Let $\eta = {1\vee \log n^{\gamma} \over n}$, and $\lambda = n^{-\theta}$ with $\theta \in[0,1)$ or $\lambda = {1 \vee \log n^{\gamma} \over n}.$
	 By a simple calculation,
	$$
	\| (I - \proj) \TK^{1\over 2} \|^2  \leq \| (I - \proj) \TK_{\bx \eta }^{1 \over 2}\|^2 \|\TK_{\bx \eta}^{-{1 \over 2}} \TK_{\eta}^{1\over 2}\|^2 .
	$$
	Using
	$$
	\| (I - \proj) \TK_{\bx \eta}^{1 \over 2}\|^2  =  \| (I - \proj) \TK_{\bx \eta}(I - \proj)\| \leq  \| (I - \proj) \TX(I - \proj)\|  + \eta \|(I - \proj)^2\| \leq  \| (I - \proj) \TX^{1\over 2}\|^2 + \eta, 
	$$
	we get
	\be\label{eq:s3}
	\| (I - \proj) \TK^{1\over 2} \|^2  \leq \left(\| (I - \proj) \TX^{1\over 2}\|^2 + \eta \right)\| \TK_{\bx \eta}^{-{1 \over 2}} \TK_{\eta}^{1\over 2}\|^2 .
	\ee

 Following from Lemma \ref{lem:empEffdim} and Lemma \ref{lem:operDifRes}, we know that there exists a subset $\Omega_1$ of $\HK^n$ with measure at least $1- 2 \delta$ such that for every $\bx \in \Omega_1$,
	$$
	\tr(\TXL^{-1} \TX) \leq b_{\gamma,\delta}\lambda^{-\gamma}  ,       
	$$
	and 
	\be
	\|\TK_{\bx \eta}^{-{1\over 2}} \TK^{1\over 2}_{\eta}\|^2 \leq a_{\gamma} \log{4 \over \delta}, \label{eq:extreDiff}
	\ee
	where $b_{\gamma,\delta} = b_{\gamma} \log^2{4\over \delta}.$ 
	For every $\bx \in \Omega_1$, according to Lemma \ref{lem:OperDiffProd}, we know that there exists a subset $U_{\bx}$ of $\mR^{m\times n}$ with measure at least $1-\delta,$ such that for all $\skt \in U_{\bx},$
	\be\label{eq:s2}
	\|(I - \proj)\TX^{1\over 2}\|^2 \leq  6 \lambda,    
	\ee
	provided that,
	\be\label{eq:s1}
	m \geq 100c_0' \log^{\beta}n \lambda^{-\gamma}\log^3{3 \over \delta} \left( 1 + 10 b_{\gamma}   \right),
	\ee
	which is satisfied under the constraint \eqref{eq:skeDim}.
	From the above analysis, we can conclude that if \eqref{eq:skeDim} holds, then with probability at least 
	$1 - 3\delta,$ \eqref{eq:s2} and \eqref{eq:extreDiff} hold.  Introducing \eqref{eq:s2} and \eqref{eq:extreDiff} into \eqref{eq:s3}, one gets that with probability at least $1 -3\delta,$
	\bea
	\| (I - \proj) \TK^{1\over 2} \|^2  \leq \left(6\lambda + \eta \right)a_{\gamma} \log {4 \over \delta},
	\eea
	which leads to the desired result.
\end{proof}

\subsection{Estimating Projection Errors with ALS Nystr\"{o}m Subsampling}\label{subapp:als}
To prove Lemma \ref{lem:DZXALS},  we first introduce the following lemma, which estimates the empirical version of the projection error. 
\begin{lemma}\label{lem:OperDiffProdALS}
	Let $0<\delta <1$ and $\theta \in[0,1].$
	Given a fixed input subset $\bx \subseteq \HK^n$, assume that for $\lambda \in [0,1]$,
	\eqref{eq:empEffDim}
	holds for some $b_{\gamma}>0$, $\gamma \in[0,1]$.
	Then there exists a subset $U_{\bx} $ of $\mR^{m\times n}$ with measure at least $1-\delta$, such that for all $\skt \in U_{\bx}$, 
	\be\label{eq:o3}
	\|(I - \proj)\TX^{1\over 2}\|^2 \leq  3\lambda,
	\ee
	provided that
	\be\label{eq:subsamLevLS}
	m \geq 8 b_{\gamma} \lambda^{-\gamma} L^2 \log {8 b_{\gamma} \lambda^{-\gamma} \over  \delta}.
	\ee	
\end{lemma}

\begin{proof}
	If we choose $\mu = 0$ in the proof of Lemma \ref{lem:OperDiffProd}, then $\SX = \mcS_1$ and $\mcS_2 = 0$. Similarly, $\TX = T_1$.
	In this case, \eqref{eq:s7} reads as
	\begin{align*}
	\TXL^{-{1\over 2}}(\TX - W^*W )\TXL^{-{1\over 2}} 
	=& V (\Sigma^2 + \lambda I)^{-1/2} \Sigma  ( I - U^* \skt^* \skt U) \Sigma (\Sigma^2 + \lambda I)^{-1/2} V^*. 
	\end{align*}
	Thus, using $V^* V = I$, $U^*U = I$ and $U$ is of full column rank,
	\begin{align*}
	\|\TXL^{-{1\over 2}}(\TX - W^*W )\TXL^{-{1\over 2}} \|
	\leq & \|V\| \|U^*U (\Sigma^2 + \lambda I)^{-1/2} \Sigma U^* ( I -  \skt^* \skt)U\Sigma (\Sigma^2 + \lambda I)^{-1/2} U^*U \|  \|V^*\|\\
	\leq & \| U (\Sigma^2 + \lambda I)^{-1/2} \Sigma U^* ( I -  \skt^* \skt)U\Sigma (\Sigma^2 + \lambda I)^{-1/2} U^*\|. 
	\end{align*}
Using $\bK: = \bK_{ \bx \bx} = \SX \SX^* = U \Sigma^2 U^*,$ we get
	\begin{align*}
	\|\TXL^{-{1\over 2}}(\TX - W^*W )\TXL^{-{1\over 2}} \|
	\leq & \| \left( \bK(\bK + \lambda I)^{-1}\right)^{1/2}( I -  \skt^* \skt)\left( \bK(\bK + \lambda I)^{-1}\right)^{1/2}\|. 
	\end{align*}
	Letting $\mcX_i = \left( \bK(\bK + \lambda I)^{-1}\right)^{1/2} \ba_i \ba_i^{*}  \left( \bK(\bK + \lambda I)^{-1}\right)^{1/2}$, it is easy to prove that $\mE [\ba_i \ba_i^{*}] = I,$ according to the definition of ALS Nystr\"{o}m subsampling. Then
	 the above inequality can be written as
	\begin{align*}
	\|\TXL^{-{1\over 2}}(\TX - W^*W )\TXL^{-{1\over 2}} \|
	\leq & \| {1\over m} \sum_{i=1}^m (\mE[\mcX_i] - \mcX_i )\|. 
	\end{align*}
	A simple calculation shows that 
	\begin{align*}
		\| \mcX_i\| =&  \ba_i^{*}  \left( \bK(\bK + \lambda I)^{-1}\right)\ba_i  \leq \max_{j \in [n]} {\left( \bK(\bK + \lambda I)^{-1}\right)_{jj} \over q_{j} }\\
	=&
	\max_{j \in [n]}  {l_j(\lambda) \over  q_j}  =  \max_{j \in [n]}	{l_j(\lambda) \sum_k \hat{l}_{k}(\lambda)   \over  \hat{l}_{j}(\lambda) }  \leq L^2 \sum_j {l}_{j}(\lambda) = L^2 \tr(\bK \bK_\lambda^{-1}),
	\end{align*}
	and 
	$$
	\mE[\mcX_i^2] = \mE[\ba_i^{*}  \left( \bK(\bK + \lambda I)^{-1}\right)\ba_i \mcX_i] \leq L^2 \tr(\bK \bK_\lambda^{-1}) \mE[\mcX_i] = L^2 \tr(\bK \bK_\lambda^{-1}) \bK\bK_\lambda^{-1}.
	$$
	Thus, 
	$$
	\| \mE[\mcX_i] - \mcX_i \|   \leq   \mE\| \mcX_i\| +  \|\mcX_i \| \leq 2 L^2 \tr(\bK\bK_\lambda^{-1}), 
	$$
	and 
	$$
	\mE\Big[\big(\mcX_i - \mE[\mcX_i]\big)^2\Big]  \preceq \mE[\mcX_i^2]  \preceq  L^2 \tr(\bK \bK_\lambda^{-1}) \bK\bK_\lambda^{-1}. 
	$$
	Letting $\mcV = L^2 \tr(\bK \bK_\lambda^{-1}) \bK\bK_\lambda^{-1} ,$ we have 
	$$\|\mcV\| \leq L^2 \tr(\bK \bK_\lambda^{-1}) ,
	$$
	and 
	 $$
	{\tr(\mcV) \over \|\mcV\|} =   {\tr(\bK \bK_\lambda^{-1}) \over \| \bK \bK_\lambda^{-1}\|} = {\tr(\bK \bK_\lambda^{-1}) \left( 1 + {\lambda \over \|\bK\|}\right)}.
	$$
	Applying Lemma \ref{lem:concentrSelfAdjoint}, noting that $\tr(\bK \bK_\lambda^{-1}) = \tr(\TX \TXL^{-1})$ and $\|\bK\| = \|\TX\|$ as $\TX = \SX^*\SX$, we get that there exists a subset $U_\bx$ of  $\mR^{m \times n}$ with measure at least $1 -\delta$ such that for all $\skt \in U_\bx,$
		\begin{align*}
	\|\TXL^{-{1\over 2}}(\TX - W^*W )\TXL^{-{1\over 2}} \|
	\leq {4 L^2 \tr(\TX \TXL^{-1}) \beta \over 3 m} + \sqrt{2 L^2 \tr(\TX \TXL^{-1}) \beta \over m }, \quad  \beta = \log {4 \tr(\TX \TXL^{-1})(1 + \lambda /\|\TX\|) \over \delta}.
	\end{align*}
	If $\lambda \leq \|\TX\|,$ 	 using Condition \eqref{eq:empEffDim},  we have
	$$\beta \leq \log {4 b_{\gamma} \lambda^{-\gamma} (1+ \lambda/\|\TX\|) \over  \delta} \leq \log {8 b_{\gamma} \lambda^{-\gamma} \over  \delta},
	$$
	and, combining with \eqref{eq:subsamLevLS},
	$$
{4 L^2 \tr(\TX \TXL^{-1}) \beta \over 3 m} + \sqrt{2 L^2 \tr(\TX \TXL^{-1}) \beta \over m } \leq {2\over 3}.
	$$
	Thus, 
	\bea
	\left\| \TXL^{-1/2}(\TK - M)\TXL^{-1/2}  \right\| \leq {2\over 3}, \quad \forall \skt \in U_{\bx}.
	\eea
	Following from \eqref{eq:s4} and \eqref{eq:s8}, one can prove \eqref{eq:o3} for the case $\lambda\leq \|\TX\|$. The proof for the case $\lambda \geq \|\TX\|$ is trivial:
	$$
	\|(I - \proj )\TX^{1\over2}\|^2 \leq \|I - \proj\|^2 \|\TX^{1\over2}\|^2 \leq \|\TX\| \leq \lambda.
	$$
	The proof is complete.
	\end{proof}

With the above lemma, and using a similar argument as that for Lemma \ref{lem:DZX}, we can prove Lemma \ref{lem:DZXALS}.
We thus skip it.

\end{document}